\def\isarxiv{1} 

\ifdefined\isarxiv
\documentclass[11pt]{article}

\usepackage[numbers]{natbib}

\else

\documentclass[letterpaper]{article} 
\usepackage[submission]{aaai25}  
\usepackage{times}  
\usepackage{helvet}  
\usepackage{courier}  
\usepackage[hyphens]{url}  
\usepackage{graphicx} 
\usepackage{natbib}  
\usepackage{caption} 
\usepackage{algorithm}

\usepackage{newfloat}
\usepackage{listings}
\DeclareCaptionStyle{ruled}{labelfont=normalfont,labelsep=colon,strut=off} 
\floatstyle{ruled}
\newfloat{listing}{tb}{lst}{}
\floatname{listing}{Listing}
\fi

\usepackage{amsmath}
\usepackage{amsthm}
\usepackage{amssymb}
\usepackage{algorithm}
\usepackage{subfig}
\usepackage{algpseudocode}
\usepackage{graphicx}
\usepackage{url}
\usepackage{xcolor}
\usepackage{epstopdf}

\usepackage{dsfont}

\allowdisplaybreaks

\ifdefined\isarxiv

\usepackage{tikz}
\usepackage{hyperref}  
\hypersetup{colorlinks=true,citecolor=blue,linkcolor=blue} 
\usetikzlibrary{arrows}
\usepackage[margin=1in]{geometry}

\else

\usepackage{microtype}

\fi

\theoremstyle{plain}
\newtheorem{theorem}{Theorem}[section]
\newtheorem{lemma}[theorem]{Lemma}
\newtheorem{definition}[theorem]{Definition}

\newtheorem{fact}[theorem]{Fact}
\newtheorem{remark}[theorem]{Remark}

\newcommand{\wh}{\widehat}
\newcommand{\wt}{\widetilde}
\newcommand{\ov}{\overline}

\newcommand{\R}{\mathbb{R}}

\renewcommand{\d}{\mathrm{d}}

\newcommand{\Tmat}{{\cal T}_{\mathrm{mat}}}

\DeclareMathOperator*{\Z}{\mathbb{Z}}

\DeclareMathOperator{\poly}{poly}

\DeclareMathOperator{\nnz}{nnz}

\DeclareMathOperator{\diag}{diag}

\makeatletter
\newcommand*{\RN}[1]{\expandafter\@slowromancap\romannumeral #1@}
\makeatother

\usepackage{lineno}

\ifdefined\isarxiv

\else

\title{Inverting the Leverage Score Gradient: An Efficient Approximate Newton Method}
\author{
    Written by AAAI Press Staff\textsuperscript{\rm 1}\thanks{With help from the AAAI Publications Committee.}\\
    AAAI Style Contributions by Pater Patel Schneider,
    Sunil Issar,\\
    J. Scott Penberthy,
    George Ferguson,
    Hans Guesgen,
    Francisco Cruz\equalcontrib,
    Marc Pujol-Gonzalez\equalcontrib
}
\affiliations{
    \textsuperscript{\rm 1}Association for the Advancement of Artificial Intelligence\\

    1101 Pennsylvania Ave, NW Suite 300\\
    Washington, DC 20004 USA\\
    proceedings-questions@aaai.org
}

\fi

\begin{document}

\ifdefined\isarxiv

\date{}

\title{Inverting the Leverage Score Gradient: An Efficient Approximate Newton Method}
\author{
Chenyang Li\thanks{\texttt{lchenyang550@gmail.com }. Fuzhou University.}
\and
Zhao Song\thanks{\texttt{zsong@adobe.com}. Adobe Research.}
\and
Zhaoxing Xu\thanks{\texttt{ellamarshall384838@gmial.com}. Wuhan University. 
}
\and
Junze Yin\thanks{\texttt{junze@bu.edu}. Boston University.}
}

\else

\maketitle

\fi

\ifdefined\isarxiv
\begin{titlepage}
  \maketitle
  \begin{abstract}
Leverage scores have become essential in statistics and machine learning, aiding regression analysis, randomized matrix computations, and various other tasks. This paper delves into the inverse problem, aiming to recover the intrinsic model parameters given the leverage scores gradient. This endeavor not only enriches the theoretical understanding of models trained with leverage score techniques but also has substantial implications for data privacy and adversarial security. We specifically scrutinize the inversion of the leverage score gradient, denoted as $g(x)$. An innovative iterative algorithm is introduced for the approximate resolution of the regularized least squares problem stated as $\min_{x \in \mathbb{R}^d} 0.5 \|g(x) - c\|_2^2 + 0.5\|\mathrm{diag}(w)Ax\|_2^2$. Our algorithm employs subsampled leverage score distributions to compute an approximate Hessian in each iteration, under standard assumptions, considerably mitigating the time complexity. Given that a total of $T = \log(\| x_0 - x^* \|_2/ \epsilon)$ iterations are required, the cost per iteration is optimized to the order of $O( (\mathrm{nnz}(A) + d^{\omega} ) \cdot \mathrm{poly}(\log(n/\delta))$, where $\mathrm{nnz}(A)$ denotes the number of non-zero entries of $A$.

  \end{abstract}
  \thispagestyle{empty}
\end{titlepage}

{\hypersetup{linkcolor=black}
\tableofcontents
}
\newpage

\else

\begin{abstract}
Leverage scores have become essential in statistics and machine learning, aiding regression analysis, randomized matrix computations, and various other tasks. This paper delves into the inverse problem, aiming to recover the intrinsic model parameters given the leverage scores gradient. This endeavor not only enriches the theoretical understanding of models trained with leverage score techniques but also has substantial implications for data privacy and adversarial security. We specifically scrutinize the inversion of the leverage score gradient, denoted as $g(x)$. An innovative iterative algorithm is introduced for the approximate resolution of the regularized least squares problem stated as $\min_{x \in \mathbb{R}^d} 0.5 \|g(x) - c\|_2^2 + 0.5\|\mathrm{diag}(w)Ax\|_2^2$. Our algorithm employs subsampled leverage score distributions to compute an approximate Hessian in each iteration, under standard assumptions, considerably mitigating the time complexity. Given that a total of $T = \log(\| x_0 - x^* \|_2/ \epsilon)$ iterations are required, the cost per iteration is optimized to the order of $O( (\mathrm{nnz}(A) + d^{\omega} ) \cdot \mathrm{poly}(\log(n/\delta))$, where $\mathrm{nnz}(A)$ denotes the number of non-zero entries of $A$.

\end{abstract}

\fi

\section{Introduction}

Leverage scores have emerged as a crucial tool in various domains of statistics and machine learning. They play a key role in regression analysis \cite{cy21,akk+20,m18}, enabling more robust and efficient model fitting. In the field of randomized matrix computations \cite{m11,dmmw12}, leverage scores underpin important sampling and sketching techniques that accelerate a wide range of linear algebra tasks. Mathematically, it is defined as follows:
\begin{definition}[Leverage score]\label{def:leverage_score}
    For a given matrix $A \in \R^{n \times d}$, $\sigma \in \R^n$ is called the leverage score defined on $A$, where the $i$-th row of $\sigma$, denoted as $\sigma_i \in \R$, is equal to $A_{i, *} ( A^\top A )^{-1} A_{i, *}^\top \in \R$, where $A_{i, *} \in \R^{1 \times d}$ is the $i$-th row of $A$.
\end{definition}
Leverage scores capture the importance or influence of each row in the matrix. Rows with high leverage scores have a larger impact on the least squares fit than rows with low scores.
In recent years, leverage scores have found numerous applications in accelerating and robustifying algorithms for matrix computations, optimization, and machine learning \cite{jrm+99,lkn99,zjp+20}. The key idea is that instead of uniformly sampling or reweighing the rows of $A$, one can obtain better results by sampling or reweighing them proportional to their leverage scores. Intuitively, this ensures that influential rows are more likely to be included.

Due to the importance of leverage score in the fields of machine learning and statistics, \cite{lsw+24} studies the following leverage score inversion problem:
\begin{definition}[Leverage score inversion problem \cite{lsw+24}]\label{def:inverting_leverage_score}
    Given the matrix $A \in \R^{n \times d}$, the vector $b \in \R^n$, and the leverage score $\sigma \in \R^n$ (see Definition~\ref{def:leverage_score}), we define $s_x : = A x -b \in \R^n$, $S_x := \diag(s_x) \in \R^{n \times n}$, and $A_x:= S_x^{-1} A \in \R^{n \times d} $.
    The goal of the inverting leverage score problem is to find the vector $x \in \R^d$ that minimizes $L_b(x) \in \R$, which is defined as follows
    \begin{align*}
        L_b(x) := \| \diag( \sigma ) - I_n \circ (A_x (A_x^\top A_x )^{-1} A_x^\top )  \|_F.
    \end{align*}
\end{definition}
They analyze the leverage score distributions to invert them and recover the model parameters.

In this paper, we take one step further by considering the following leverage score gradient inversion problem:
\begin{definition}[Leverage score gradient inversion problem]\label{def:our}
Let $L_b(x) \in \R$ be defined as in Definition~\ref{def:inverting_leverage_score}. Let $x \in \R^d$ and $g(x) = \frac{\d L_b(x)}{\d x} \in \R^d$. Let $x \in \R^d$. We define the leverage score gradient inversion problem as $L_c(x) = \frac{1}{2}\|g(x) - c\|_2^2$.

Given the matrix $A \in \R^{n \times d}$ and $w \in \mathbb{R}^n$, we let $L_\mathrm{reg}(x) = 0.5\|\mathrm{diag}(w) Ax\|_2^2$ be the regularization term.

We define the regularized leverage score gradient inversion problem as finding 
\begin{align*}
    \min_{x \in \R^d} L(x) = \min_{x \in \R^d} (L_c(x) + L_\mathrm{reg}(x)).
\end{align*}
\end{definition}

This problem is important for several reasons. First, it provides insight into the theoretical interpretability of models trained using leverage score techniques. Second, it has implications for data privacy, as sensitive training data could potentially be reconstructed from publicly released leverage scores \cite{cle+19,zlh19}. Third, it is relevant to adversarial security, as systems that rely on leverage score sampling may be vulnerable to attacks if adversaries can accurately invert the scores. 

From an algorithmic perspective, we present a new iterative algorithm for approximately solving the regularized least squares problem $\min_{x \in \R^d} 0.5 \|g(x) - c\|_2^2 + 0.5\|\mathrm{diag}(w)Ax\|_2^2$, where $g(x)$ is a leverage score based gradient function and $\mathrm{diag}(w)$ is a diagonal regularization matrix. Our algorithm is based on approximate Newton's method, using subsampled leverage score distributions to form an approximate Hessian in each iteration.
Under standard assumptions that the initial point $x_0$ is close to the optimum $x^*$ and the Hessian of the objective satisfies certain Lipschitz continuity and positive definite properties, we prove that our algorithm converges to an $\epsilon$-approximate solution in $O(\log(1/\epsilon))$ iterations. Moreover, we show how to implement each iteration in nearly input-sparsity time, i.e. $O((\mathrm{nnz}(A) + d^\omega) \mathrm{poly}\log(n/\delta))$ time, by applying subsampled leverage score estimation techniques. Here $\mathrm{nnz}(A)$ is the number of nonzero entries in $A$, $\omega \approx 2.373$ is the exponent of matrix multiplication \cite{w12,lg14,aw21,dwz23,lg23,wxxz23}, and $\delta$ is the failure probability.
Compared to classical iterative optimization methods for least squares, our algorithm exploits the special leverage score structure in the gradient and Hessian to enable faster convergence and per-iteration costs. This demonstrates the power of combining tools from randomized numerical linear algebra with convex optimization to derive input-sparsity time algorithms for fundamental machine learning problems.

\begin{theorem}[Informal version of our main result (Theorem~\ref{thm:main_formal})]\label{thm:main_informal}
Let $\epsilon, \delta \in (0,0.1)$. Given $A \in \R^{n \times d}$, $b \in \R^n$, $w \in \R^n$, we let $x^*$ be the optimal solution of our regularized least squares problem (see Definition~\ref{def:our}). Let $x_0 \in \R^d$ be close to $x^*$ (see Definition~\ref{def:f_ass}), where $0$ denotes the $0$-th iteration of the vector $x$.

Then, there exists a randomized algorithm (Algorithm~\ref{alg:main}) such that, with at least $1-\delta$ probability, it runs $T = \log(\| x_0 - x^* \|_2/ \epsilon)$ iterations and outputs the $T$-th iteration of $x$, namely $\wt{x} = x_T \in \R^d$ satisfying
$\| \wt{x} - x^* \|_2 \leq \epsilon$,
and the time cost per iteration is
$O( (\nnz(A) + d^{\omega} ) \cdot \poly(\log(n/\delta))$. 
\end{theorem}

Our work is built on the foundation of \cite{lswy23}. In \cite{lswy23}, the focus is on inverting leverage score distributions to recover model parameters and proposing both first-order (gradient descent) and second-order (Newton's method) algorithms for this purpose. However, in our work, we analyze the inversion of the leverage score gradient, namely $g(x)$. This requires significant effort to compute the Hessian of this leverage score gradient, which is equivalent to computing the third-order derivative. Moreover, the analysis of positive definiteness and Lipschitz continuity is also highly non-trivial. Furthermore, our algorithm uses an approximate Hessian, which largely alleviates the time complexity of \cite{lswy23}, $n \cdot O(\Tmat(d, n, n) + \Tmat(d, n, d)) + d^{\omega}$, where $\Tmat(d, n, n)$ denotes the running time of multiplying a $d \times n$ matrix with an $n \times n$ matrix.

\paragraph{Roadmap.}

In Section~\ref{sec:related_work}, we present the related work. In Section~\ref{sec:prelimi}, we present the basic mathematical facts about matrix calculus and linear algebra. In Section~\ref{sec:gra_inf}, we present our result of gradient computation. In Section~\ref{sec:has_inf}, we present the result of Hessian computation. In Section~\ref{sec:haspd_inf}, we show that our Hessian matrix is positive definite. In Section~\ref{sec:haslip_inf}, we show that our Hessian matrix is Lipschitz continuous. In Section~\ref{sec:newton_method}, we introduce the properties of the Newton method. In Section~\ref{sec:main_result}, we present the main result and its proof. In Section~\ref{sec:conclusion}, we present our conclusion.

\paragraph{Notation}

Let $x, y \in \R^d$. We define $\langle x, y \rangle = \sum_{i = 1}^d x_i \cdot y_i$. 
We let $[n]:=\{1,2,3, \ldots, n\}$. $\circ$ is a binary operation called the Hadamard product: $x \circ y \in \R^d$ is defined as $(x \circ y)_i:=x_i \cdot y_i$. Also, we have $x^{\circ 2}=x \circ x$. For all $p \in \mathbb{Z}_{+}$, we define the $\ell_p$ norm of the vector $x$, denoted as $\|x\|_p$ to be equal to $\sqrt[p]{\sum_{i=1}^d |x_i |^p}$. $1_n$ is the $n$-dimensional vector whose entries are all ones. $e_k$ is a vector whose $k$-th entry equals 1 and other entries are 0 . When dealing with iterations, we use $x_t$ to denote the $t$-th iteration. In this paper, we only use the letters $t$ and $T$ for expressing iterations. $\langle x, y\rangle$ represents the inner product of the vectors $x$ and $y$. Let $A \in \R^{n\times n}$ and $x \in \R^n$. Then, we have $ \langle x x^\top , A \rangle = x^\top A x$. We define $\diag : \R^d  arrow \R^{d \times d}$ as $\diag(x)_{i, i}:=x_i$ and $\diag(x)_{i, j}:=0$, for all $i \neq j$. We define $ (A_{i, *} )^{\top} \in \R^d$ to be the $i$-th row of $A$, and define $A_{*, j} \in \R^n$ to be the $j$-th column of $A$. We define the spectral norm and the Frobenius norm of $A$ as $\|A\|:=\max _{x \in \R^d}\|A x\|_2 /\|x\|_2$ with $\|x\|_2 \neq 0$ and $\|A\|_F:=\sqrt{\sum_{i=1}^n \sum_{j=1}^d |A_{i, j} |^2}$, respectively. We use $x^*$ to denote the exact solution. $\nabla L$ and $\nabla^2 L$ denote the gradient and Hessian respectively. $\mathcal{T}_{\mathrm{mat}}(n, d, d)$ represents the running time of multiplying a $n \times d$ matrix with a $d \times d$ matrix. $\nnz(A)$ represents the number of non-zero entries of the matrix $A$. 

 \section{Related Work}\label{sec:related_work}

\paragraph{Leverage score.}

Leverage scores is a statistical concept which is used to analyze the linear regression model. It represents the extent of how individual data points influence the general performance of a certain linear regression \cite{ch86}. 
From \cite{dkm06}, leverage scores were brought to the field of numerical linear algebra as a measure of the importance of each row in a matrix for solving linear regression problems. They showed that sampling rows according to their leverage scores can lead to efficient approximate solutions for least-squares problems.
Since then, leverage scores have been used in a variety of contexts. \cite{md09} applied leverage scores to the problem of matrix approximation and developed the concept of ``CUR decomposition'', which approximates a matrix using a subset of its rows and columns selected based on their leverage scores.

Leverage scores have also been used for feature selection in machine learning. \cite{pkb14} proposed a method for selecting informative features based on their leverage scores, showing that this approach can lead to improved performance in tasks such as classification and clustering.
In recent years, leverage scores have found applications in a wider range of domains. They have been utilized in kernel methods \cite{ss02}, as well as in approximate factorizations and sampling techniques \cite{ltos19,emm20,am15,clv17,mm17,lhc+20}. Furthermore, leverage scores have been employed in weighted low rank approximation \cite{syyz23} and the matrix completion problem \cite{gsyz23}. Additionally, they have been used in the development of quantum algorithms for solving linear regression, multiple regression, and ridge regression problems \cite{syz23}.

\paragraph{Second order method.}

Second order optimization methods have a long history in the machine learning literature. Newton's method, which uses the Hessian matrix of second derivatives to inform the optimization trajectory, was first published by Isaac Newton in 1736 and later refined by Joseph Raphson. Quasi-Newton methods like BFGS \cite{bdm73} and L-BFGS \cite{ln89} approximate the Hessian using first-order gradient information, striking a balance between the fast quadratic convergence of Newton's method and the computational efficiency of first-order methods.

More recently, second-order methods have been applied to neural network optimization. \cite{m10} introduced Hessian-free optimization, using conjugate gradients to approximately solve the Newton update. \cite{vp12} used a Krylov subspace descent method to directly approximate the Newton update. 
For natural gradient methods, which perform steepest descent in the space of network outputs rather than parameters, \cite{a98} showed a connection to second order optimization via the Fisher information matrix. \cite{mg15} introduced K-FAC, which uses a block-diagonal approximation to the Fisher matrix for efficient natural gradient updates. \cite{gm16} later extended K-FAC to convolutional neural networks. \cite{bgm22} combined natural gradient with trust region methods to further improve stability and performance.

Despite these advances, second order neural network optimization remains an active area of research, such as \cite{swy23,gswy23,gsy23_hyper,gsy23_coin,bsy23}. Open problems include improving the scalability of Hessian approximations, handling non-convex optimization landscapes, and automating hyperparameter selection. Our work builds on prior methods while introducing novel techniques to address these challenges.

\section{Preliminary}
\label{sec:prelimi}

Now, we present some basic facts about linear algebra and matrix calculus.

\begin{fact}\label{fac:vector}

    Let $h, m, k \in \R^{n}$. 

Then, the following properties hold:
\begin{itemize}
    \item $\langle h, m \rangle = \langle h \circ m, {\bf 1}_n \rangle =  h^\top \mathrm{diag}(m)  {\bf 1}_n$
    \item $\langle h \circ m, k \rangle = \langle h \circ k, m \rangle =  \langle h \circ m \circ k, {\bf 1}_n  \rangle = h^\top \mathrm{diag}(m) k$
    \item $h \circ m = m \circ h = \mathrm{diag} (h) \cdot m = \mathrm{diag} (m) \cdot h$
    \item $h^{\top}(m \circ k) = m^{\top}(h \circ k) = k^{\top}(h \circ m)$
    \item $\mathrm{diag} (h \circ m) = \mathrm{diag} (h) \mathrm{diag} (m)$
    \item $\mathrm{diag} (h) + \mathrm{diag} (m) = \mathrm{diag} (h + m)$
    \item $\langle h, m \rangle = \langle m, h \rangle = h^\top m = m^\top h$
    \item For all $a,b \in \R$, $a\langle h, m \rangle + b\langle k, m \rangle = \langle ah + bk, m \rangle = \langle m, ah + bk \rangle = a\langle m, h \rangle + b\langle m, k \rangle$
\end{itemize}
\end{fact}

\begin{fact}\label{fac:inner_product_matrix}
    Let $A \in \R^{n\times n}$ and $x \in \R^n$. Then, we have
    \begin{align*}
        \langle x x^\top , A \rangle = x^\top A x
    \end{align*}
\end{fact}
\begin{proof}
    \begin{align*}
        \langle x x^\top , A \rangle 
        = & ~ \sum_{i, j \in [n]} (x x^\top)_{i, j} A_{i, j}\\
        = & ~ \sum_{i, j \in [n]} x_i x_j^\top A_{i, j}\\
        = & ~ \sum_{i, j \in [n]} x_i A_{i, j} x_j^\top\\
        = & ~ x^\top A x,
    \end{align*}
    where the first step follows from the definition of inner product for matrices, the second step, third step and last step follows from simple algebra.
\end{proof}

Then, we introduce the algebraic properties for matrices and spectral norms.

\begin{fact}\label{fac:matrix_norm}

Let $P$ and $Q$ be two arbitrary matrices, where $P \cdot Q$ exists.
Let $d \in \R$.
Let $h$ be an arbitrary vector where $Ph$ exists.

Then, we have
\begin{itemize}
    \item $\|d \cdot P\| \leq |d| \cdot \|P\|$
    \item $\|P^\top\| = \|P\|$
    \item $\|P + Q\| \leq \|P\| + \|Q\|$
    \item $\|P \cdot Q\| \leq \|P\| \cdot \|Q\|$
    \item For any vector $h$, we have $\|Ph\|_2 \leq \|P\| \cdot \|h\|_2$
    \item $(PQ)^\top = Q^\top P^\top$
\end{itemize}

\end{fact}

\begin{fact}[Basic derivative rules]\label{fac:derivative_rules}

If the following conditions hold:
\begin{itemize}
    \item Let $m, p \in \Z_+$ and $j \in \Z$.
    \item Let $h \in \R^p$ be a vector.
    \item Let $u \in \R$ be a scalar.
    \item Let $b$ be independent of $u$.
    \item Let $q: \R^p \to \R^m$.
    \item Let $r: \R^p \to \R^m$.
    \item Let $s: \R^p \to \R$.
\end{itemize}

Then, we have:
\begin{itemize}
    \item Part 1. $\frac{\d (b \cdot q(h))}{\d u} = b \cdot \frac{\d q(h)}{\d u}$ (constant multiple rule).
    \item Part 2. $\frac{\d (s(h)^j)}{\d u} = j \cdot s(h)^{j - 1} \cdot \frac{\d s(h)}{\d u}$ (power rule).
    \item Part 3. $\frac{\d (r(h) \pm q(h))}{\d u} = \frac{\d r(h)}{\d u} \pm \frac{\d q(h)}{\d u}$ (sum/difference rule).
    \item Part 4. $\frac{\d (r(h) \circ q(h))}{\d u} = \frac{\d r(h)}{\d u} \circ q(h) + r(h) \circ \frac{\d q(h)}{\d u}$ (product rule for Hadamard product).
    \item Part 5. $\frac{\d (s(h) q(h))}{\d u} = \frac{\d s(h)}{\d u} q(h) + s(h) \frac{\d q(h)}{\d u}$ (product rule).
\end{itemize}

\end{fact}

\section{Gradient Computation}
\label{sec:gra_inf}

In this section, we present a crucial result for computing the gradient of our objective function. Lemma~\ref{lem:g(x)_informal} provides an explicit formula for the gradient of $L_c(x)$, which is essential for our optimization algorithm.

\begin{definition}\label{def:A_i_informal}
    Let $A_x \in \R^{n \times d}$ and $b \in \R^n$ be defined as in Definition~\ref{def:inverting_leverage_score}. Let $\sigma_{*,*}(x) = A_x (A_x^\top A_x)^{-1} A_x^\top \in \R^{n \times n}$. Let $\sigma_{*,i}(x) \in \R^n$ denote the $i$-th column of $\sigma_{*,*}(x)$, for all $i \in [n]$. Let $p(x) = \sigma_{*,i}(x) - b \in \R^n$.

    We define
        \begin{align*}
\wt{A}_1 := & ~ -10 A_x^\top \sigma_{*,*}^{\circ 2}(x) \diag(p(x))    A_{x}  \\
\wt{A}_2 := & ~ -2  A_x^\top \Sigma(x) \diag(p(x)) A_x   \\
\wt{A}_3 := & ~ + 8 A_x^\top \sigma_{*,*}(x)  \sigma_{*,*}^{\circ 2}(x) \diag( p(x) )  A_x
 \\
\wt{A}_4 := & ~ - 4 A_x^\top \Sigma(x)  \sigma_{*,*}(x) \diag( p(x) )  A_x\\
\wt{A}_5 := & ~ - 2 A_x^\top \sigma_{*,*}^{\circ 2}(x) \diag(\sigma_{*,i}(x) )A_x\\
\wt{A}_6 := & ~ + 2 A_x^\top  \Sigma(x) \diag(\sigma_{*,i}(x)) A_x\\
\wt{A}_7 := & ~ + 4 A_x^\top  \sigma_{*,*}^{\circ 2}(x) \sigma_{*,*}(x)   \diag(\sigma_{*,i}(x))A_x\\
\wt{A}_8 := & ~ - 4 A_x^\top  \Sigma(x)  \sigma_{*,*}(x)    \diag(\sigma_{*,i}(x))A_x\\
\wt{A}_9 := & ~  - 2A_{x}^\top \sigma_{*,*}^{\circ 2}(x) \sigma_{*,i}(x) e_i^\top A_x  \\
\wt{A}_{10} := & ~ +2A_{x}^\top \Sigma(x) \sigma_{*,i}(x)  e_i^\top A_x.
        \end{align*} 
For each $i \in [10]$, $\wt{A}_i \in \R^{d \times d}$ is equal to some term $D_i \in \R^{d \times n}$ multiplying with $A_{x} \in \R^{n \times d}$. We define these $D_i \in \R^{d \times n}$ as the terms of $\wt{A}_i$ without $A_x$.
\end{definition}

\begin{lemma}[Informal version of Lemma~\ref{lem:L_c}]\label{lem:g(x)_informal}
    Let $x \in \R^d$ and $g(x) = \frac{\d L_b(x)}{\d x} \in \R^d$ be defined as in Definition~\ref{def:inverting_leverage_score}. Let $q(x) = g(x) - c \in \R^d$, where $c \in \R^d$ is defined as in Definition~\ref{def:our}. For all $i \in [10]$, we let $\wt{A}_i$ be defined as in Definition~\ref{def:A_i_informal}.
    Then, we have
        \begin{align*}
            \frac{\d L_{c} (x)}{\d x } = (\sum_{i = 1}^{10} \wt{A}_i)^\top q(x).
        \end{align*}

\end{lemma}

The full proof of Lemma~\ref{lem:g(x)_informal} can be found in Appendix~\ref{sec:gradient}. We first derive the gradient of $L_{c,j_0}(x)$ with respect to each component $x_j$, where $L_{c,j_0}:= (g(x)_{j_0} - c_{j_0})^2$ for $j_0 \in [d]$.
These individual gradients are then combined to form the full gradient vector.
The proof leverages various matrix identities and properties of the leverage score components to simplify and reorganize terms.
Special care is taken to handle the interactions between different matrices and vectors like $\sigma_{*,*}(x)$, $\sigma_{*,i}(x)$, and $A_x$.
The final result is obtained by grouping similar terms and expressing them in a compact form using the $\wt{A}_i$ matrices.

\section{Hessian Computation}
\label{sec:has_inf}

In Section~\ref{sec:has_inf_lc}, we present the Hessian matrix of $L_c(x)$, namely the leverage score gradient inversion problem. In Section~\ref{sec:preli:reg}, we present the gradient and Hessian of the regularization term.

\subsection{Hessian of the Leverage Score Gradient Inversion Problem}
\label{sec:has_inf_lc}

Now, we compute the Hessian of $L_c(x)$.
\begin{lemma}[Informal version of Lemma~\ref{lem:gradient_Al}]\label{lem:gradient_Al_informal}
    For each $l \in [10]$, let $A_l$ be the $j$-th column of $\wt{A}_l$, for any arbitrary $j \in [d]$. Let $x_{j_2} \in \R$. Due to the numerous quantity of equations of $A_{l, h}$, please refer to the formal version of this lemma, namely Lemma~\ref{lem:gradient_Al}, for the specific definition of $A_{l, h}$. Then, the gradient of $A_l$ can be expressed as:
    \begin{itemize}
        \item $\frac{\d A_{1}}{\d x_{j_2}} = \sum_{h = 1}^8 A_{1, h} := \mathsf{A}_1$,
        \item $\frac{\d A_{2}}{\d x_{j_2}} = \sum_{h = 1}^7 A_{2, h} := \mathsf{A}_2$,
        \item $\frac{\d A_{3}}{\d x_{j_2}} = \sum_{h = 1}^{11} A_{3, h} := \mathsf{A}_3$,
        \item $\frac{\d A_{4}}{\d x_{j_2}} = \sum_{h = 1}^{10} A_{4, h} := \mathsf{A}_4$,
        \item $\frac{\d A_{5}}{\d x_{j_2}} = \sum_{h = 1}^8 A_{5, h} := \mathsf{A}_5$,
        \item $\frac{\d A_{6}}{\d x_{j_2}} = \sum_{h = 1}^8 A_{5, h} := \mathsf{A}_6$,
        \item $\frac{\d A_{7}}{\d x_{j_2}} = \sum_{h = 1}^{11} A_{7, h} := \mathsf{A}_7$,
        \item $\frac{\d A_{8}}{\d x_{j_2}} = \sum_{h = 1}^{10} A_{8, h} := \mathsf{A}_8$,
        \item $\frac{\d A_{9}}{\d x_{j_2}} = \sum_{h = 1}^{8} A_{9, h} := \mathsf{A}_9$, and
        \item $\frac{\d A_{10}}{\d x_{j_2}} = \sum_{h = 1}^{7} A_{10, h} := \mathsf{A}_{10}$.
    \end{itemize}
\end{lemma}

In Lemma~\ref{lem:L_c_Hessian}, we consider 
\begin{align*}
    \frac{\d^2 L_{c} (x)}{\d x_{j} \d x_{j_2} }.
\end{align*}
Using the techniques from matrix calculus rules and numerical linear algebra techniques, we can get
\begin{align*}
    \frac{\d^2 L_{c} (x)}{\d x_{j} \d x_{j_2} } = \underbrace{(\sum_{l=1}^{10} \wt{A}_l)^\top}_{1 \times d} \underbrace{\sum_{l=1}^{10} A_l}_{d \times 1} + \underbrace{(g(x) - c)^\top}_{1 \times d} \underbrace{\sum_{l=1}^{10} \mathsf{A}_l}_{d \times 1}.
\end{align*}

Regarding the first term, we can express it as
\begin{align*}
    \underbrace{A_{x, *, j_2}^\top}_{1 \times n} \underbrace{B_1(x)}_{n \times n} \underbrace{A_{x, *, j}}_{n \times 1},
\end{align*}
where 
\begin{align*}
    \underbrace{B_1(x)}_{n \times n} = \underbrace{\sum_{l=1}^{10} D_l^\top}_{n \times d} \underbrace{\sum_{l=1}^{10} D_l}_{d \times n}
\end{align*}

Similarly, we can express the second term as
\begin{align*}
    \underbrace{A_{x, *, j_2}^\top}_{1 \times n}  \underbrace{B_2(x)}_{n \times n} \underbrace{A_{x, *, j}}_{n \times 1}.
\end{align*}

Combining everything together, we have
\begin{align*}
    \frac{\d^2 L_c(x)}{\d x^2} 
    = & ~ \underbrace{A_{x}^\top}_{d \times n}  \underbrace{B_1(x)}_{n \times n} \underbrace{A_{x}}_{n \times d} + \underbrace{A_{x}^\top}_{d \times n}  \underbrace{B_2(x)}_{n \times n} \underbrace{A_{x}}_{n \times d}\\
    = & ~ \underbrace{A_{x}^\top}_{d \times n}  \underbrace{(\underbrace{B_1(x)}_{n \times n} +  \underbrace{B_2(x)}_{n \times n})}_{:= B(x)} \underbrace{A_{x}}_{n \times d}.
\end{align*}

Therefore, we finish the computation of Hessian:
\begin{lemma}[Informal version of Lemma~\ref{lem:L_c_Hessian}]\label{lem:L_c_Hessian_informal}

Let $x \in \R^d$. Let $L_c(x)$ be defined as in Definition~\ref{def:our}. Let $A_x \in \R^{n \times d}$ be defined as in Definition~\ref{def:inverting_leverage_score}. 
 Then, we have 
    \begin{align*}
        \frac{\d^2 L_{c} (x)}{\d x^2 } = A_{x}^\top B(x) A_{x}.
    \end{align*}
\end{lemma}

\subsection{Hessian of the Regularization Term}\label{sec:preli:reg}

This definition is paraphrased from \cite{dls23}. The Lemma is in \cite{lsz23}.

\begin{definition}[Definition 4.8 of \cite{dls23}]\label{def:L_reg}
Given matrix $A \in \R^{n \times d}$.
For a given vector $w \in \R^n$, let $W = \diag(w)$. 
We define $L_{\mathrm{reg}} : \R^d \rightarrow \R$ as follows
\begin{align*}
L_{\mathrm{reg}}(x):= 0.5 \| W A x\|_2^2
\end{align*}
\end{definition}

\begin{lemma}[Folklore, see \cite{lsz23} as an example]\label{lem:L_reg_gradient_hessian}
For a given vector $w \in \R^n$, let $W = \diag(w)$. Let $L_{\mathrm{reg}} : \R^d \rightarrow \R$ be defined as Definition~\ref{def:L_reg}.

Then, we have
\begin{itemize}
\item The gradient is
\begin{align*}
\frac{\d L_{\mathrm{reg}}}{ \d x} = A^\top W^2 Ax
\end{align*}
\item The Hessian is
\begin{align*}
\frac{\d^2 L_{\mathrm{reg}}}{ \d x^2} = A^\top W^2 A
\end{align*}
\end{itemize}
\end{lemma}

\section{Hessian is Positive Definite}
\label{sec:haspd_inf}

In this section, we present a crucial result establishing that the Hessian matrix of our objective function is positive definite. This property is essential for ensuring the convexity of our optimization problem and the convergence of the Newton method.
Combining Lemma~\ref{lem:L_c_Hessian_informal} and Lemma~\ref{lem:L_reg_gradient_hessian}, we have
\begin{align}\label{eq:L_hessain}
    \frac{\d^2 L}{\d x^2} 
    = & ~ A_{x}^\top B(x) A_{x} + A^\top W^2 A \notag\\
    = & ~ A^\top S_{x}^{-1} B(x) S_{x}^{-1} A + A^\top W^2 A \notag\\
    = & ~ A^\top (S_{x}^{-1} B(x) S_{x}^{-1}  +  W^2) A,
\end{align}
where the second step follows from $A_x = S_{x}^{-1} A$ (see Definition~\ref{def:inverting_leverage_score}) and the last step follows from simple algebra.

\begin{lemma}[Informal version of Lemma~\ref{lem:convex}]\label{lem:convex_informal}

Let $l > 0$ denote a scalar. Suppose for all $i \in [n]$, $W_{i, i}^2 \geq 12000 \beta^3 R + l/\sigma_{\min}(A)^2$, where $\beta \in (0, 0.1)$ and $R > 0$. Let $L$ be defined as in Definition~\ref{def:our}, $x \in \R^d$, and the Hessian of $L$ be computed as in Lemma~\ref{lem:L_c_Hessian_informal}.
Then, we have
    \begin{align*}
    \frac{\d^2 L}{\d x^2} \succeq l \cdot I_d
    \end{align*}
\end{lemma}

Defining $G(x) := S_{x}^{-1} B(x) S_{x}^{-1}$, we know it suffices to bound
\begin{align*}
    \| G(x) + W^2 \| \leq \| G(x) \| + \| W^2 \|.
\end{align*}

$\| W^2 \|$ is bounded by $12000 \beta^3 R + l/\sigma_{\min}(A)^2$, where this assumption is derived from prior work on regularization \cite{lsz+23}.
We establish bounds on $\|G(x)\|$ by bounding $\|B_1(x)\|$ and $\|B_2(x)\|$.

Using matrix norm inequalities, we bound $\|B_1(x)\|$ by $(\sum_{l=1}^{10} \|D_l\|)^2$.
We then bound each $\|D_l\|$ individually. This is done by carefully analyzing the structure of each $D_l$, which are derived from terms involving $\sigma_{*,*}(x)$, $\Sigma(x)$, $\sigma_{*,*}^{\circ 2}(x)$, $\diag(p(x))$.
Using properties of these leverage score components (e.g., $\|\sigma_{*,*}(x)\| \leq 1$, $\|\Sigma(x)\| \leq 1$) that we derive (see the full detail in Lemma~\ref{lem:spectral_norm_bound} and Lemma~\ref{lem:more_spectral_norm_bound}), we establish upper bounds for each $\|D_l\|$.
Summing these bounds and squaring the result gives us the final bound on $\|B_1(x)\|$:
\begin{align*}
    \|B_1(x)\| \leq 4100
\end{align*}

The approach for $B_2(x)$ is slightly different. We note that $B_2(x)$ involves terms with $(g(x) - c)$.
We first bound $\|g(x) - c\|_2$ using the definition of $g(x)$ and properties of the leverage score components.
We then analyze each term in $B_2(x)$, which involves products of $A_x$, $\sigma_{*,*}(x)$, $\Sigma(x)$, and $(g(x) - c)$.
Using matrix and vector norm inequalities, along with the bounds on leverage score components and $\|g(x) - c\|_2$, we derive bounds for each term in $B_2(x)$.

Summing these bounds gives us the final result for $\|B_2(x)\|$:
\begin{align*}
    \|B_2(x)\| \leq 7900\beta R
\end{align*}

Combining everything together, we can show the correctness of Lemma~\ref{lem:convex_informal}.

\section{Hessian is Lipschitz}
\label{sec:haslip_inf}

In this section, we demonstrate that the Hessian matrix of our objective function satisfies the Lipschitz continuity property. We begin by formally defining what it means for a Hessian to be Lipschitz continuous:

\begin{definition}[$L_h$-Lipschitz]\label{def:lip}
    Let $h: \R^m \to \R^m$ be a function. The function $h$ is $L_h$-Lipschitz if there exists a real number $L_h \geq 0$ such that for all $x, y \in \R^m$,
    \begin{align*}
        \|h(x) - h(y)\|_2 \leq L_h \cdot \|x - y\|_2.
    \end{align*}
\end{definition}

\begin{lemma}[Informal version of Lemma~\ref{lem:lip_lc}]\label{lem:lip_lc_informal}

Let $L$ be defined as in Definition~\ref{def:our} and the Hessian of $L$ be computed as in Lemma~\ref{lem:L_c_Hessian_informal}. Let $x, \wh{x} \in \R^d$.

    Then, we have
    \begin{align*}
        \|\frac{\d^2 L}{\d x^2}(x) - \frac{\d^2 L}{\d x^2}(\wh{x})\| \leq 1024000 \beta^{-7} R^6 \|x - \wh{x}\|_2,
    \end{align*}
    where $\beta \in (0, 0.1)$ and $R > 0$
\end{lemma}

The proof of this lemma follows a careful decomposition and bounding strategy. 

We start by expressing the bound of $\|S_x - S_{\wh{x}}\|$, $\|A_x - A_{\wh{x}}\|$, $\|\Sigma(x) - \Sigma(\wh{x})\|$, and $\|p(x) - p(\wh{x})\|_2$ in terms of $\|x - \wh{x}\|_2$.

Additionally, we then analyze two more complicated bounds:
\begin{itemize}
    \item $\|\sigma_{*, *}^{\circ 2}(x) - \sigma_{*, *}^{\circ 2}(\wh{x})\| \leq 6\beta^{-7} R^3 \|x - \wh{x}\|_2$
    \item $\|g(x) - g(\wh{x})\|_2 \leq 60 \beta^{-7} R^4 \|x - \wh{x}\|_2$
\end{itemize}
Combining these together, we can form
\begin{align*}
    \|B(x) - B(\wh{x})\| \leq 1000000 \beta^{-7} R^4 \|x - \wh{x}\|_2.
\end{align*}

Using the triangle inequality, we break this difference 
\begin{align*}
    \|\frac{\d^2 L}{\d x^2}(x) - \frac{\d^2 L}{\d x^2}(\wh{x})\|
\end{align*}
into three terms, each involving a difference between two matrices.
\begin{itemize}
    \item $\|A_{x}^\top B(x) A_{x} - A_{x}^\top B(x) A_{\wh{x}}\|$,
    \item $\|A_{x}^\top B(x) A_{\wh{x}} - A_{x}^\top B(\wh{x}) A_{\wh{x}}\|$, and
    \item $\|A_{x}^\top B(\wh{x}) A_{\wh{x}} - A_{\wh{x}}^\top B(\wh{x}) A_{\wh{x}}\|$.
\end{itemize}
We bound each of these terms separately.
We use the bounds on $\|A_x\|$ and $\|B(x)\|$ established earlier.
Combining these bounds and simplifying the resulting expressions, we arrive at the final inequality stated in the lemma.

\section{Approximate Newton Method}\label{sec:newton_method}

In this section, we present definitions and properties of Newton's method and the approximate Newton's method.

\begin{definition}[$(l,M)$-good Loss function]\label{def:f_ass}

Let $l, M > 0$ be arbitrary real numbers. Let $L : \R^d \rightarrow \R$ be an arbitrary loss function. $L$ is $(l,M)$-good if
\begin{itemize}
    \item {\bf $L$ has a unique $l$-local minimum:}  
    this means that there exists a unique vector $x^* \in \R^d$ such that
    \begin{itemize}
        \item $\nabla L(x^*) = {\bf 0}_d$ and 
        \item $H(x^*) \succeq l \cdot I_d$.
    \end{itemize}
    \item {\bf The Hessian of $L$ is $M$-Lipschitz:} according to Definition~\ref{def:lip}, that is, for all $x, y \in \R^d$,
    \begin{align*}
        \| H(y) - H(x) \| \leq M \cdot \| y - x \|_2 
    \end{align*}
\end{itemize}
\end{definition}

\begin{definition}[Good initialization point]
    Given a recurrence relation $x_{n + 1} = f(x_n)$ and a $(l, M)$-good function $L$, this recurrence relation has a good seed (or good initialization point) $x_0$ relative to $L$ if
    \begin{align*}
        \| x_0 -x^*\|_2 M \leq 0.1 l,
    \end{align*} 
    where $x^*$ denotes the optimal solution of $L$.
\end{definition}

Here, we present a definition of an exact update of Newton's method.
\begin{definition}[Exact update of Newton's method]\label{def:exact_update_variant}
Let $L: \R^d \rightarrow \R$ be a loss function. Suppose it has a gradient function $g: \R^d \rightarrow \R^d$ and a Hessian function $H : \R^d \rightarrow \R^{d \times d}$. The exact update of Newton's method for finding a zero of the function $L$ is a recurrence relation defined on $L$:
\begin{align*}
    x_{t+1} = x_t - H(x_t)^{-1} \cdot g(x_t).
\end{align*}
\end{definition}

Finding the Hessian matrix is very expensive in many real-world tasks. Therefore, in the algorithm of \cite{dls23}, an approximated computation of the Hessian is utilized. We present related definitions below.

\begin{definition}[$\epsilon_0$-approximate Hessian]\label{def:wt_H}
Let $x \in \R^d$ and $H(x) \in \R^{d \times d}$ be a Hessian matrix. For all $\epsilon_0 \in (0, 0.1)$, we define an $\epsilon_0$-approximate Hessian\footnote{This approximate Hessian does not need to be a Hessian matrix. It is used to approximate the Hessian $H(x) \in \R^{d \times d}$.} $\wt{H}(x) \in \R^{d \times d}$ to be a matrix that satisfies:
\begin{align*}
 (1-\epsilon_0) \cdot H(x) \preceq \wt{H}(x) \preceq (1+\epsilon_0) \cdot H(x).
\end{align*}
\end{definition}

\begin{remark}
    Note that, $\omega$ denotes the currently known best exponent of matrix multiplication, currently $\omega \approx 2.373$ \cite{w12,lg14,aw21}. It quantifies the rate at which the computational cost increases with the size of the matrices being multiplied. The higher $\omega$ is, the higher the computational cost it takes of multiplying two matrices.
\end{remark}

Lemma~4.5 in \cite{dsw22} states the existence of an algorithm for providing an $\epsilon_0$-approximate Hessian $\wt{H}(x_t)$ efficiently. An approximate Hessian is much easier to compute than the Hessian.
\begin{lemma}[\cite{dsw22,syyz22}]\label{lem:subsample}
Let $\epsilon_0, \delta \in (0, 0.1)$. 
Let $A \in \R^{n \times d}$. 
Let $\nnz(A) \in \mathbb{N}$ denote the number of nonzero entries of $A$. 

Then, for all $i \in [n]$, for all $D \in \R^{n \times n}$ satisfying $D_{i, i} > 0$, there exists an algorithm which runs in time
\begin{align*}
O( (\nnz(A) + d^{\omega} ) \poly(\log(n/\delta)) )
\end{align*}
and outputs an $O(d \log(n/\delta))$ sparse diagonal matrix $\wt{D} \in \R^{n \times n}$, i.e. a diagonal matrix where most of the entries are zeros, and the number of non-zero entries is less than or equal to a constant times $d \log(n/\delta)$, such that 
\begin{align*}
(1- \epsilon_0) A^\top D A \preceq A^\top \wt{D} A \preceq (1+\epsilon_0) A^\top D A.
\end{align*}
\end{lemma}

Following from \cite{a00,jkl+20,szz21,hjs+22,lsz23}, we consider an approximate update of Newton's method.
\begin{definition}[$\epsilon_0$-approximate update Newton's method]\label{def:update_x_k+1}

Let $L: \R^d \rightarrow \R$ be a loss function. Suppose it has the gradient function $g: \R^d \rightarrow \R^d$ and the Hessian matrix $H : \R^d \rightarrow \R^{d \times d}$. Let $\wt{H} : \R^d \rightarrow \R^{d \times d}$ be an $\epsilon_0$-approximate Hessian defined in Definition~\ref{def:wt_H} and obtained through Lemma~\ref{lem:subsample}, for any $\epsilon_0 \in (0, 0.1)$. An $\epsilon_0$-approximate update of Newton's method is a recurrence relation defined on $L$:
\begin{align*}
    x_{t+1} = x_t  - \wt{H}(x_t)^{-1} \cdot  g(x_t)  .
\end{align*}
\end{definition}

Now, we show some mathematical properties of the derivation of a convergent and stable approximate Newton's method through the definitions of the positive definiteness and Lipschitz properties.
\begin{lemma}[Iterative shrinking, Lemma 6.9 on page 32 of \cite{lsz23}]\label{lem:one_step_shrinking}

For a positive integer $t$, we define $x_t \in \R^d$ to be the $t$-th iteration of a recurrence relation $x_{t + 1} = f(x_t)$. We let $x^* \in \R^d$ be the unique exact solution of the leverage score gradient inversion problem (see Definition~\ref{def:our}), for fixed $A \in \R^{n \times d}$, $b \in \R^n$, and $w \in \R^n$. 
Let $L: \R^d \to \R$ be a loss function which is $(l,M)$-good (see Definition~\ref{def:f_ass}). Let $r_t:= \| x_t - x^* \|_2$. Let $\ov{r}_t: = M \cdot r_t$.

Then, for all $\epsilon_0 \in (0,0.1)$, we have  
\begin{align*}
r_{t+1} \leq 2 \cdot (\epsilon_0 + \ov{r}_t/( l - \ov{r}_t ) ) \cdot r_t.
\end{align*} 
\end{lemma}

We define $T$ as the total iterations required by an algorithm. To utilize Lemma~\ref{lem:one_step_shrinking}, the following induction hypothesis lemma is necessary. This approach is used in \cite{lsz23}.
\begin{lemma}[Induction on the recurrence relation with variable $t$, see Lemma 6.10 on page 34 of \cite{lsz23}]\label{lem:newton_induction}
For a positive integer $t$, for each $i \in [t]$, we define $x_i \in \R^d$ to be the $i$-th iteration of a recurrence relation $x_{t + 1} = f(x_t)$. We let $x^* \in \R^d$ be the exact solution of the leverage score gradient inversion problem (see Definition~\ref{def:our}) for our choice of $A \in \R^{n \times d}$, $b \in \R^n$, and $w \in \R^n$. For each $i \in [t]$, we define $r_i:= \| x_i - x^* \|_2$. Let $\epsilon_0 \in (0, 0.1)$. Suppose $r_{i} \leq 0.4 \cdot r_{i-1}$, for all $i \in [t]$. For $M$ and $l$ to be defined for Definition~\ref{def:f_ass}, we will assume $M \cdot r_i \leq 0.1 l$, for all $i \in [t]$.

Then we have
\begin{itemize}
    \item $r_{t+1} \leq 0.4 r_t$.
    \item $M \cdot r_{t+1} \leq 0.1 l$.
\end{itemize}
\end{lemma}

\begin{proof}
    See \cite{lsz23}, Lemma 6.10 on page 34.
\end{proof}

\section{Main Result}\label{sec:main_result}

In this section, we present our main theoretical result and the algorithm it supports. This work addresses the challenging problem of inverting the leverage score gradient, which has significant implications for understanding and optimizing models trained with leverage score techniques.

\begin{algorithm}[!ht]\caption{Here, we present our main algorithm.}\label{alg:main}
\begin{algorithmic}[1]
\Procedure{LeverageScoreGradientInversion}{$A \in \R^{n \times d},b \in \R^n,w \in \R^n, \epsilon, \delta$} \Comment{Theorem~\ref{thm:main_formal}} 
    \State We choose $x_0$ (suppose it satisfies Definition~\ref{def:f_ass})
    \State We use $T \gets \log( \| x_0 - x^* \|_2 / \epsilon )$ to denote the number of iterations.
    \For{$t=0 \to T$} 
        \State $D \gets G(x_t) + \diag(w \circ w)$ 
        \State $\wt{D} \gets \textsc{SubSample}(D,A,\epsilon_1 = \Theta(1), \delta_1 = \delta/T)$ \Comment{Lemma~\ref{lem:subsample}}
        \State $g \gets (\sum_{i = 1}^{10} \wt{A}_i)^\top q(x_t) + A^\top \diag(w \circ w) Ax$
        \State $\wt{H} \gets A^\top \wt{D} A$ 
        \State $x_{t+1} \gets x_t - \wt{H}^{-1} g$ 
    \EndFor
    \State $\wt{x}\gets x_{T+1}$
    \State \Return $\wt{x}$
\EndProcedure
\end{algorithmic}
\end{algorithm}

\begin{theorem}[Main Result]\label{thm:main_formal}
Let $\omega \approx 2.373$ denote the exponent of matrix multiplication. Let $\epsilon, \delta \in (0,0.1)$. Given $A \in \R^{n \times d}$, $b \in \R^n$, $w \in \R^n$, we let $x^*$ be the optimal solution of 
\begin{align*}
    L(x) 
    = & ~ L_c(x) + L_{\mathrm{reg}} \\
    = & ~ 0.5 \cdot \| g(x) - c\|_2^2 + 0.5 \| \diag(w) A x\|_2^2.
\end{align*}
Let $x_0 \in \R^d$ be the $(l, M)$-good initialization point (see Definition~\ref{def:f_ass}).

Then there exists a randomized algorithm (Algorithm~\ref{alg:main}) such that, with at least $1-\delta$ probability, it runs $T = \log(\| x_0 - x^* \|_2/ \epsilon)$ iterations and outputs $\wt{x} \in \R^d$ such that
\begin{align*}
\| \wt{x} - x^* \|_2 \leq \epsilon,
\end{align*}
and the time cost per iteration is
\begin{align*}
O( (\nnz(A) + d^{\omega} ) \cdot \poly(\log(n/\delta)). 
\end{align*}

\end{theorem}

\begin{proof}
The proof of Theorem~\ref{thm:main_formal} relies on the results established in earlier sections, particularly the positive definiteness of the Hessian (Section~\ref{sec:haspd_inf}) and its Lipschitz continuity (Section~\ref{sec:haslip_inf}). These properties ensure that our approximate Newton method converges rapidly and reliably to the optimal solution.

    {\bf Proof of gradient computation.}

    It follows from combining Lemma~\ref{lem:L_reg_gradient_hessian} and Lemma~\ref{lem:L_c}.

    {\bf Proof of Hessian computation.}

    It follows from combining Lemma~\ref{lem:L_reg_gradient_hessian} and Lemma~\ref{lem:L_c_Hessian}.

    {\bf Proof of the positive definiteness of Hessian matrix.}

    It follows from Lemma~\ref{lem:convex}.

    {\bf Proof of the Lipschitz continuous property of Hessian.}

    It follows from Lemma~\ref{lem:lip_lc}.
    
    {\bf Proof of Cost per iteration.}

It follows from Lemma~\ref{lem:subsample}.

    {\bf Proof of Convergence per Iteration.}

By Lemma~\ref{lem:one_step_shrinking}, we have
\begin{align*}
    \|x_k - x^*\|_2 \le 0.4 \cdot \|x_{k-1} - x^*\|_2.
\end{align*}

{\bf Proof of Number of Iterations.}

    After $T$ iterations, we have
    \begin{align*}
    \| x_T - x^* \|_2 \leq 0.4^T \cdot \| x_0 - x^* \|_2
    \end{align*}
\end{proof}

By providing both a strong theoretical result and an efficient practical algorithm, our work opens up new possibilities for analyzing and optimizing models that use leverage score techniques, with potential applications in areas such as data privacy, model interpretability, and robust machine learning.

\section{Conclusion}\label{sec:conclusion}
In conclusion, this paper introduces a novel iterative algorithm, underpinned by Newton's approximate method, that utilizes subsampled leverage score distributions to construct an approximate Hessian at each iteration. We delve deep into analyzing the inversion of the leverage score gradient, which is a challenging task with profound implications. Through this investigation, our algorithm stands apart as it employs an approximate Hessian, effectively alleviating the time complexity of \cite{lswy23}, stated as 
\begin{align*}
    n \cdot O(\Tmat(d, n, n) + \Tmat(d, n, d)) + d^{\omega},
\end{align*}
where $\Tmat(d, n, n)$ represents the computation time needed for the multiplication of a $d \times n$ matrix with an $n \times n$ matrix.

Our central focus in the paper has been the inverse problem, which aims to uncover the underlying model parameters given the leverage score gradient. The importance of this issue cannot be stressed enough as it broadens our understanding and insight into the interpretability of models trained using leverage score techniques. Equally significant is its capacity to guide us toward resolving data privacy concerns and avoid potential attacks in systems dedicated to the leverage score sampling.

\ifdefined\isarxiv

\else
\bibliography{ref}
\section*{Reproducibility Checklist}

\paragraph{This paper:}
\begin{itemize}
    \item Includes a conceptual outline and/or pseudocode description of AI methods introduced (yes/partial/no/NA) {\bf yes}
    \item Clearly delineates statements that are opinions, hypothesis, and speculation from objective facts and results (yes/no) {\bf yes}
    \item Provides well marked pedagogical references for less-familiare readers to gain background necessary to replicate the paper (yes/no) {\bf yes}
\end{itemize}

\paragraph{Does this paper make theoretical contributions? (yes/no)} {\bf yes}

If yes, please complete the list below.

\begin{itemize}
    \item All assumptions and restrictions are stated clearly and formally. (yes/partial/no) {\bf yes}
    \item All novel claims are stated formally (e.g., in theorem statements). (yes/partial/no) {\bf yes}
    \item Proofs of all novel claims are included. (yes/partial/no) {\bf yes}
    \item Proof sketches or intuitions are given for complex and/or novel results. (yes/partial/no) {\bf yes}
    \item Appropriate citations to theoretical tools used are given. (yes/partial/no) {\bf yes}
    \item All theoretical claims are demonstrated empirically to hold. (yes/partial/no/NA) {\bf NA}
    \item All experimental code used to eliminate or disprove claims is included. (yes/no/NA) {\bf NA}
\end{itemize}

\paragraph{Does this paper rely on one or more datasets? (yes/no)} {\bf no}

If yes, please complete the list below.

\begin{itemize}
    \item A motivation is given for why the experiments are conducted on the selected datasets (yes/partial/no/NA)
    \item All novel datasets introduced in this paper are included in a data appendix. (yes/partial/no/NA)
    \item All novel datasets introduced in this paper will be made publicly available upon publication of the paper with a license that allows free usage for research purposes. (yes/partial/no/NA)
    \item All datasets drawn from the existing literature (potentially including authors’ own previously published work) are accompanied by appropriate citations. (yes/no/NA)
    \item All datasets drawn from the existing literature (potentially including authors’ own previously published work) are publicly available. (yes/partial/no/NA)
    \item All datasets that are not publicly available are described in detail, with explanation why publicly available alternatives are not scientifically satisficing. (yes/partial/no/NA)
\end{itemize}

\paragraph{Does this paper include computational experiments? (yes/no)} {\bf no}

If yes, please complete the list below.

\begin{itemize}
    \item Any code required for pre-processing data is included in the appendix. (yes/partial/no).
    \item All source code required for conducting and analyzing the experiments is included in a code appendix. (yes/partial/no)
    \item All source code required for conducting and analyzing the experiments will be made publicly available upon publication of the paper with a license that allows free usage for research purposes. (yes/partial/no)
    \item All source code implementing new methods have comments detailing the implementation, with references to the paper where each step comes from (yes/partial/no)
    \item If an algorithm depends on randomness, then the method used for setting seeds is described in a way sufficient to allow replication of results. (yes/partial/no/NA)
    \item This paper specifies the computing infrastructure used for running experiments (hardware and software), including GPU/CPU models; amount of memory; operating system; names and versions of relevant software libraries and frameworks. (yes/partial/no)
    \item This paper formally describes evaluation metrics used and explains the motivation for choosing these metrics. (yes/partial/no)
    \item This paper states the number of algorithm runs used to compute each reported result. (yes/no)
    \item Analysis of experiments goes beyond single-dimensional summaries of performance (e.g., average; median) to include measures of variation, confidence, or other distributional information. (yes/no)
    \item The significance of any improvement or decrease in performance is judged using appropriate statistical tests (e.g., Wilcoxon signed-rank). (yes/partial/no)
    \item This paper lists all final (hyper-)parameters used for each model/algorithm in the paper’s experiments. (yes/partial/no/NA)
    \item This paper states the number and range of values tried per (hyper-) parameter during development of the paper, along with the criterion used for selecting the final parameter setting. (yes/partial/no/NA)
\end{itemize}

\fi

\newpage
\onecolumn
\appendix

\paragraph{Roadmap}

In Section~\ref{sec:preliminary}, we present the preliminaries and notations. In Section~\ref{sec:gradient}, we present our result of the gradient computation. In Section~\ref{sec:hessian}, we present our result of the Hessian computation. In Section~\ref{sec:hessian_positive_definite}, we present the positive definite of the Hessian matrix. In Section~\ref{sec:hessian_lipschitz_continuous}, we present the Lipschitz Continuous of the Hessian matrix.

\section{Preliminary}\label{sec:preliminary}

\paragraph{Notation}

Let $x, y \in \R^d$. We define $\langle x, y \rangle = \sum_{i = 1}^d x_i \cdot y_i$. 
We let $[n]:=\{1,2,3, \ldots, n\}$. $\circ$ is a binary operation called the Hadamard product: $x \circ y \in \R^d$ is defined as $(x \circ y)_i:=x_i \cdot y_i$. Also, we have $x^{\circ 2}=x \circ x$. For all $p \in \mathbb{Z}_{+}$, we define the $\ell_p$ norm of the vector $x$, denoted as $\|x\|_p$ to be equal to $\sqrt[p]{\sum_{i=1}^d |x_i |^p}$. $1_n$ is the $n$-dimensional vector whose entries are all ones. $e_k$ is a vector whose $k$-th entry equals 1 and other entries are 0 . When dealing with iterations, we use $x_t$ to denote the $t$-th iteration. In this paper, we only use the letters $t$ and $T$ for expressing iterations. $\langle x, y\rangle$ represents the inner product of the vectors $x$ and $y$. Let $A \in \R^{n\times n}$ and $x \in \R^n$. Then, we have $ \langle x x^\top , A \rangle = x^\top A x$. We define $\diag : \R^d  arrow \R^{d \times d}$ as $\diag(x)_{i, i}:=x_i$ and $\diag(x)_{i, j}:=0$, for all $i \neq j$. We define $ (A_{i, *} )^{\top} \in \R^d$ to be the $i$-th row of $A$, and define $A_{*, j} \in \R^n$ to be the $j$-th column of $A$. We define the spectral norm and the Frobenius norm of $A$ as $\|A\|:=\max _{x \in \R^d}\|A x\|_2 /\|x\|_2$ with $\|x\|_2 \neq 0$ and $\|A\|_F:=\sqrt{\sum_{i=1}^n \sum_{j=1}^d |A_{i, j} |^2}$, respectively. We use $x^*$ to denote the exact solution. $\nabla L$ and $\nabla^2 L$ denote the gradient and Hessian respectively. $\mathcal{T}_{\mathrm{mat}}(n, d, d)$ represents the running time of multiplying a $n \times d$ matrix with a $d \times d$ matrix. $\nnz(A)$ represents the number of non-zero entries of the matrix $A$.

\begin{definition}\label{def:s_x}
Let $A \in \R^{n \times d}$, $b \in \R^n$, $x \in \R^d$, and $a_i^\top \in \R^{1 \times d}$ be the $i$-th row of $A$ for all $i \in [n]$.
We define $s_x \in \R^n$ as follows:
\begin{align*}
    s_x :=  Ax - b,
\end{align*}
where the $i$-th entry of the vector $s_x$ is denoted as $s_{x,i} = a_i^\top x - b_i \in \R$, for all $i \in [n]$. Also, we use $S_x \in \R^{n \times n}$ to denote $\diag(s_x)$.
\end{definition}

We assume that all entries of $s_x$ are non-zero. Therefore, the diagonal matrix $S_x = \diag(s_x)$ is invertible. 

\begin{definition}\label{def:A_x}
Let $A \in \R^{n \times d}$. Let $s_x \in \R^n$ and $S_x \in \R^{n \times n}$ be defined as in Definition~\ref{def:s_x}. 
We define $A_x \in \R^{n \times d}$ as
\begin{align*}
    A_x := S_x^{-1} A.
\end{align*}
\end{definition}

\begin{remark}
    By the definition of $A_x$, we can get that
\begin{align*}
    A_x^\top A_x = A^\top S_x^{-1} S_x^{-1} A = A^\top S_x^{-2} A
\end{align*}
\end{remark}

\begin{definition}\label{def:sigma}

Let $A_x \in \R^{n \times d}$ be defined as in Definition~\ref{def:A_x}.
Let $a_{x,i}^\top \in \R^{1 \times d}$ denote the $i$-th row of $A_x \in \R^{n \times d}$, for all $i \in [n]$.
We define $\sigma_{*,*}(x) \in \R^{n \times n}$ matrix as follows: 
\begin{align*}
    \underbrace{\sigma_{*,*}(x)}_{n \times n} = \underbrace{A_x}_{n \times d} \underbrace{(A_x^\top A_x)^{-1}}_{d \times d} \underbrace{A_x^\top}_{d \times n}
\end{align*}

Moreover, we express each entry of $\sigma_{*,*}(x)$ as follows:
\begin{align*}
    \underbrace{\sigma_{i,i}( x )}_{\mathrm{scalar}} := \underbrace{a_{x,i}^\top}_{1 \times d} \underbrace{( A_x^\top A_x )^{-1}}_{d \times d} \underbrace{a_{x,i} }_{d \times 1},
\end{align*}
for each $i\in [n]$, and 
\begin{align*}
    \sigma_{i,l}( x ) := a_{x,i}^\top ( A_x^\top A_x )^{-1} a_{x,l},
\end{align*}
for each $i \in [n]$, for each $l \in [n]$.

Considering the $i$-th column of $\sigma_{*,*}(x)$, we define 
\begin{align*} 
    \underbrace{ \sigma_{*,i}(x)  }_{n \times 1} := \underbrace{ A_x }_{n \times d} \underbrace{ (A_x^\top A_x)^{-1} }_{d \times d} \underbrace{ a_{x,i} }_{d \times 1},
\end{align*}
for each $i \in [n]$.

Finally, we define $\sigma_{*,*}^{\circ 2 }(x) \in \R^{n \times n}$ as 
\begin{align*}
    \underbrace{ \sigma_{*,*}^{\circ 2 }(x) }_{n \times n} = \underbrace{ \sigma_{*,*}(x) }_{n \times n} \circ \underbrace{ \sigma_{*,*}(x) }_{n \times n}.
\end{align*}
\end{definition}

\begin{definition}\label{def:Sigma}
We define $n \times n$ diagonal $\Sigma(x)$ as follows
\begin{align*}
    \Sigma(x) := \sigma_{*,*}(x) \circ I_n.
\end{align*}
\end{definition}

\begin{definition}\label{def:loss_lb}
We define loss function $L_b(x)$
\begin{align*}
    L_b(x) := 0.5 \| \Sigma(x) - \mathrm{diag}(b) \|_F^2
\end{align*}
\end{definition}

\begin{definition}\label{def:f(x)}
    Let $f: \R^d \to \R^n$ be defined as
    \begin{align*}
        f(x)_i := \Sigma(x)_{i,i},
    \end{align*}
    for all $i \in [n]$.
\end{definition}

Therefore, we have
\begin{align*}
    L_b(x) = 0.5 \| f(x) - b \|_2^2
\end{align*}

\begin{definition} \label{def:g(x)}
Let $g(x)$ denote the gradient of $L_b(x)$, then we can have
\begin{align*}
    g(x) := \frac{\d L_b(x) }{\d x}
\end{align*}
\end{definition}

\begin{definition} \label{def:L_c}
Let $c  \in \R^d$. We define
\begin{align*}
   L_c(x) := 0.5 \cdot \| g(x) - c\|_2^2
\end{align*}
\end{definition}

\section{Gradient}\label{sec:gradient}

In Section~\ref{sub:gradient_s_and_S}, we present the gradients related to $s_x$ and $S_x$. In Section~\ref{sub:gradient_s_A:A}, we present the gradients related to matrix $A_x$. In Section~\ref{sub:gradient_s_A:a}, we present the gradients related to vector $a_{x, i}$. In Section~\ref{sub:gradient_s_A:AA}, we present the gradients related to scalar $A_{x, i, j}$. In Section~\ref{sub:gradient_hessian_sigma_Sigma:scalar}, we present the gradients for scalar $\sigma_{i,i}$ and scalar $\sigma_{i,j}$. In Section~\ref{sub:gradient_hessian_sigma_Sigma:vector}, we present the gradient for vector $\sigma_{*,i}$. In Section~\ref{sub:gradient_p}, we present the gradient for vector $p(x)$. In Section~\ref{sub:gradient_hessian_sigma_Sigma:matrix}, we present the gradient for matrix $\sigma_{*,*}(x)$ and $\sigma_{*,*}^{\circ 2}(x)$. In Section~\ref{sub:f(x)}, we present the gradient for scalar $f(x)_i$. In Section~\ref{sub:L_b}, we present the gradient for scalar $L_b(x)$. In Section~\ref{sub:sigma}, we present the gradient for $\Sigma(x)$. In Section~\ref{sub:g(x)}, we present the gradient for $g(x)$. In Section~\ref{sub:l_c}, we present the gradient for $L_c(x)$.

\subsection{Gradients related to $s_x$ and $S_x$}\label{sub:gradient_s_and_S}

\begin{lemma}\label{lem:gradient_s_and_S}
    If the following conditions hold
\begin{itemize}
    \item Let $s_x \in \R^n$, $s_{x, i} \in \R$, and $S_x = \diag(s_x) \in \R^{n \times n}$ be defined as in Definition~\ref{def:s_x}.
    \item Let $A_{*,j} \in \R^n$ denote the $j$-th column of matrix $A \in \R^{n \times d}$ and $a_i^\top \in \R^{1 \times d}$ denote the $i$-th row of $A$ for all $i \in [n]$.
    \item Let $A_x \in \R^{n \times d} $ be defined as in Definition~\ref{def:A_x}.
\end{itemize}
Then, we have for each $j \in [d]$,
\begin{itemize}
    \item {\bf Part 1.} 
    \begin{align*} 
        \underbrace{ \frac{ \d s_{x} }{ \d x_j } }_{n \times 1} = A_{*, j}
    \end{align*}
    \item {\bf Part 2.}
    \begin{align*}
        \underbrace{ \frac{ \d s_x^{-1} }{ \d x_j } }_{n \times 1} = - s_x^{-2} \circ A_{*, j}
    \end{align*}
    \item {\bf Part 3.}
     \begin{align*}
        \underbrace{ \frac{ \d s_x^{-2} }{ \d x_j } }_{n \times 1} = -2 \cdot s_x^{-3} \circ A_{*, j}
    \end{align*}
    \item {\bf Part 4.}
    \begin{align*}
        \frac{\d S_x^{-1}}{\d x_j} =  \diag(- s_x^{-2} \circ A_{*, j}),
    \end{align*}
    \item {\bf Part 5.}
    \begin{align*}
        \underbrace{ \frac{\d S_x^{-2}}{\d x_j} }_{n \times n} =  -2\diag(S_x^{-3} A_{*, j})
 \end{align*}
    \item {\bf Part 6.}
    \begin{align*}
        \underbrace{ \frac{ \d A^\top S_x^{-2} A }{ \d x_j } }_{d \times d} = -2 A^\top \diag(S_x^{-3} A_{*, j}) A
    \end{align*}
\end{itemize}
\end{lemma}
\begin{proof}

{\bf Proof of Part 1.}

We have
\begin{align*}
    \frac{ \d s_{x} }{ \d x_j } 
    = & ~ \frac{ \d (Ax - b) }{ \d x_j }\\
    = & ~ \frac{ \d Ax }{ \d x_j } - \frac{ \d b }{ \d x_j }\\
    = & ~ \frac{ \d Ax }{ \d x_j } \\
    = & ~ A \frac{ \d x }{ \d x_j } \\
    = & ~ A_{*, j},
\end{align*}
where the first step follows from the definition of $s_x$ (see Definition~\ref{def:s_x}), the second step follows from the difference rule in Fact~\ref{fac:derivative_rules}, the third step follows from $\frac{ \d b }{ \d x_j } = 0$, the fourth step follows from the constant multiple rule in Fact~\ref{fac:derivative_rules}, and the last step follows from simple algebra.

{\bf Proof of Part 2.}
\begin{align*}
    \underbrace{\frac{ \d s_x^{-1} }{ \d x_j }}_{n \times 1}
    = & ~ -1 \cdot \underbrace{s_x^{-2}}_{n \times 1} \circ \underbrace{\frac{ \d s_x }{ \d x_j }}_{n \times 1} \\
    = & ~ - s_x^{-2} \circ A_{*, j},
\end{align*}
where the first step follows from the power rule in Fact~\ref{fac:derivative_rules} and the second step follows from the {\bf Part 1}.

{\bf Proof of Part 3.}
\begin{align*}
    \underbrace{\frac{ \d s_x^{-2} }{ \d x_j }}_{n \times 1} 
    = &~ 2 \cdot \underbrace{s_x^{-1}}_{n \times 1} \circ \underbrace{\frac{ \d s_x^{-1} }{ \d x_j }}_{n \times 1}\\
    = &~ 2 \cdot s_x^{-1} \circ - s_x^{-2} \circ A_{*, j} \\
    = &~ -2 \cdot s_x^{-3} \circ A_{*, j},
\end{align*}
where the first step follows from the product rule in Fact~\ref{fac:derivative_rules}, the second step follows from the {\bf Part 2}, and the last step follows from simple algebra.

{\bf Proof of Part 4.}
\begin{align}\label{eq:gradient:S-1}
    \underbrace{\frac{\d S_x^{-1}}{\d x_j}}_{n \times n}  
    = & ~  \underbrace{\frac{\d ( \diag (s_x)^{-1})}{\d x_j}}_{n \times n}  \notag\\
    = & ~  {\frac{\d(\diag(  s_x^{-1})}{\d x_j}} \notag \\
    = & ~  \diag( {\frac{\d s_x^{-1}}{\d x_j}} ) \notag \\
    = & ~  \diag(- s_x^{-2} \circ A_{*, j}),
\end{align}
where the first step follows from simple algebra, the definition of $S_x$ (see Definition~\ref{def:s_x}), the second step follows from the properties of diagonal matrices, the third step follows from the matrix calculus, and the last step follows from the {\bf Part 2}.

Therefore, we have
\begin{align*}
    \underbrace{ \frac{\d S_x^{-1}}{\d x_j} }_{n \times n} 
    = & ~  \diag(- s_x^{-1} \circ A_{*,j} \circ s_x^{-1})  \\
    = & ~  - \diag(s_x^{-1} \circ A_{*,j}) \diag(s_x^{-1})  \\
    = & ~  - \diag(s_x^{-1} \circ A_{*,j}) S_x^{-1} \\
    = & ~  - \diag(S_x^{-1} A_{*,j}) S_x^{-1} \\
    = & ~  - \diag(A_{x, *,j}) S_x^{-1},
\end{align*}
where the first step follows from Eq.~\eqref{eq:gradient:S-1}, the second step follows from Fact~\ref{fac:vector}, the third step follows from the definition of $S_x$ (see from the Lemma statement), the fourth step follows from the definition of $S_x$ (see from the Lemma statement), and the last step follows from the definition of $A_x$ (see from the Lemma statement).

{\bf Proof of Part 5.}
\begin{align*}
    \underbrace{\frac{\d S_x^{-2}}{\d x_j}}_{n \times n}  
    = &~  \underbrace{\frac{\d ( S_x^{-1} \cdot S_x^{-1})}{\d x_j}}_{n \times n}  \\
    = &~  {\frac{\d(\diag(  s_x^{-1}) \cdot \diag(  s_x^{-1}))}{\d x_j}}  \\
    = &~  {\frac{\d(\diag(  s_x^{-2}) )}{\d x_j}}  \\    
    = &~  \diag( {\frac{\d s_x^{-2}}{\d x_j}} ) \\
    = &~  -2\diag(S_x^{-3} A_{*, j}),     
\end{align*}
where the first step follows from matrix multiplication, the second step follows from the definition of $S_x$ (see Definition~\ref{def:s_x}), the third and forth step follows from the properties of diagonal matrices, and the last step follows from the {\bf Part 3}.

{\bf Proof of Part 6.}
\begin{align*}
    \underbrace{ \frac{ \d A^\top S_x^{-2} A }{ \d x_j } }_{d \times d}  
    = &~ \underbrace{ A^\top \frac{ \d S_x^{-2} A }{ \d x_j } }_{d \times d}\\
    = &~ A^\top  \frac{ \d S_x^{-2}  }{ \d x_j } A \\
    = &~ -2 A^\top \diag(S_x^{-3} A_{*, j}) A,
\end{align*}
where the first and second step follows from the constant multiple rule in Fact~\ref{fac:derivative_rules}, the last step follows from the {\bf Part 5}.
\end{proof}

\subsection{Gradients Related to Matrix $A_x$}
\label{sub:gradient_s_A:A}

\begin{lemma}\label{lem:gradient_A}
    If the following conditions hold
\begin{itemize}
    \item Let $A_x \in \R^{n \times d} $ be defined as in Definition~\ref{def:A_x}.
\end{itemize}
Then, we have for each $j \in [d]$,
\begin{itemize}
    \item {\bf Part 1.}
    \begin{align*}
        \underbrace{ \frac{ \d A_x }{\d x_j} }_{n \times d} = - \diag(A_{x, *,j}) A_x
    \end{align*}
    \item {\bf Part 2.}
    \begin{align*}
        \frac{ \d A_x^\top A_x }{ \d x_j } = -  2A_x^\top \diag(A_{x, *,j}) A_x 
    \end{align*}
    \item {\bf Part 3.}
    \begin{align*}
        \frac{ \d ( A_x^\top A_x )^{-1} }{ \d x_j } = 2 (A_x^\top A_x)^{-2} \cdot A_x^\top \diag(A_{x, *,j}) A_x
    \end{align*}
\end{itemize}
\end{lemma}
\begin{proof}
    
{\bf Proof of Part 1.}

    \begin{align*}
        \underbrace{ \frac{ \d A_x }{\d x_j} }_{n \times d} 
        = &~   \underbrace{ \frac{ \d (S_x^{-1} A) }{\d x_j} }_{n \times d}  \\
        = &~   \frac{ \d S_x^{-1} }{\d x_j} A \\
        = &~  - \diag(A_{x, *,j}) S_x^{-1} A \\
        = &~ - \diag(A_{x, *,j}) A_x
    \end{align*}
where the first step follows from the definition of $A_x$ (see Definition~\ref{def:A_x}), the second step follows from the constant multiple rule in Fact~\ref{fac:derivative_rules}, the third step follows from the {\bf Part 4} of Lemma~\ref{lem:gradient_s_and_S}, and the last step follows from the definition of $A_x$ (see Definition~\ref{def:A_x}).

{\bf Proof of Part 2.}
    \begin{align*}
         \underbrace{ \frac{ \d A_x^\top A_x }{ \d x_j } }_{d \times d}
        = &~   \underbrace{ \frac{ \d A_x^\top}{\d x_j} A_x}_{d \times d} +  \underbrace{A_x^\top \frac{ \d A_x}{\d x_j} }_{d \times d}\\
        = &~  (\frac{ \d A_x}{\d x_j})^\top A_x + A_x^\top \frac{ \d A_x}{\d x_j} \\
        = &~    (- \diag(A_{x, *,j}) A_x)^\top A_x +  A_x^\top (- \diag(A_{x, *,j}) A_x)\\
        = &~  -  A_x^\top \diag(A_{x, *,j}) A_x -  A_x^\top  \diag(A_{x, *,j}) A_x \\
        = &~  -  2A_x^\top \diag(A_{x, *,j}) A_x 
    \end{align*}
where the first step follows from the product rule in Fact~\ref{fac:derivative_rules}, the second step follows from the matrix calculus, the third step follows from the {\bf Part 1}, the fourth step follows from simple algebra, and the last step follows from simple algebra.

{\bf Proof of Part 3.}
    \begin{align*}
    \underbrace{ \frac{ \d ( A_x^\top A_x )^{-1} }{ \d x_j }}_{d \times d} 
    =&~ -1 \cdot (A_x^\top A_x)^{-2} \cdot \underbrace{ \frac{ \d A_x^\top A_x }{ \d x_j } }_{d \times d} \\
    =&~ 2 (A_x^\top A_x)^{-2} \cdot A_x^\top \diag(A_{x, *,j}) A_x
    \end{align*}
where the first step follows from the power rule in Fact~\ref{fac:derivative_rules}, the second step follows from the {\bf Part 2}.
\end{proof}

\subsection{Gradients Related to Vector $a_{x, i}$}
\label{sub:gradient_s_A:a}

\begin{lemma}\label{lem:gradient_a}
    If the following conditions hold
\begin{itemize}
    \item Let $A_x \in \R^{n \times d} $ be defined as in Definition~\ref{def:A_x}.
    \item Let $A_{x, i, j} \in \R$ denote the entry of $A_x \in \R^{n\times d}$ located at the $i$-th row and $j$-th column.
    \item Let $a_i^\top \in \R^{1 \times d}$ be defined as in Definition~\ref{def:sigma}.
\end{itemize}
Then, we have for each $j \in [d]$,
\begin{itemize}
    \item {\bf Part 1.} For each $i \in [n]$
    \begin{align*}
        \frac{\d a_{x,i} }{\d x_j} = -A_{x, i,j} a_{x,i}
    \end{align*}
    \item {\bf Part 2.} For each $i \in [n]$
    \begin{align*}
        \frac{\d a_{x,i} a_{x,i}^\top }{\d x_j} = - 2 A_{x, i,j} a_{x,i} a^\top_{x,i} 
    \end{align*}
    \item {\bf Part 3.} 
    For each $i \in [n]$, we have
    \begin{align*}
    \frac{\d a_{x,l} a_{x,i}^\top }{\d x_j} =  - (A_{x, i,j} + A_{x, l,j}) a_{x,l} a_{x,i}^\top 
    \end{align*}
\end{itemize}
\end{lemma}
\begin{proof}

{\bf Proof of Part 1.}
    \begin{align*}
        \underbrace{ \frac{\d a_{x,i} }{\d x_j}}_{d \times 1}
        = & ~   \underbrace{(\frac{\d A_x }{\d x_j})_{i,*}}_{d \times 1}  \\
        = & ~ (- \diag(A_{x, *,j}) A_x )_{i,*}\\
        = & ~ -A_{x, i,j} a_{x,i}\\
    \end{align*}
where the first step follows from the definition of $a_{x,i}$ (see Definition~\ref{def:sigma}), the second step follows from the {\bf Part 1} of Lemma~\ref{lem:gradient_A}, and the third step follows from simple  algebra.

{\bf Proof of Part 2.} 
    \begin{align*}
        \underbrace{ \frac{\d a_{x,i} a^\top_{x,i}}{\d x_j}}_{d \times d}
        = & ~   \underbrace{ a_{x,i} \frac{\d  a^\top_{x,i}}{\d x_j}}_{d \times d} + \underbrace{ \frac{\d a_{x,i}}{\d x_j}  a^\top_{x,i}}_{d \times d} \\
        = & ~  -a_{x,i} (A_{x, i,j} A_{x, i,*})^\top    - A_{x, i,j} A_{x, i,*}  a^\top_{x,i}\\
        = & ~  -a_{x,i} (A_{x, i,j} a_{x,i})^\top    - A_{x, i,j} a_{x,i}  a^\top_{x,i}\\
        =  & ~   - 2 A_{x, i,j} a_{x,i} a^\top_{x,i} 
    \end{align*}
where the first step follows from the product rule in Fact~\ref{fac:derivative_rules}, the second step follows from the {\bf Part 1} , the third step follows from the definition of $a_{x,i}$ (see Definition~\ref{def:A_x}),  and the last step follows from simple  algebra.

{\bf Proof of Part 3.} 
    \begin{align*}
        \underbrace{ \frac{\d a_{x,l} a^\top_{x,i}}{\d x_j}}_{d \times d}
        = & ~   \underbrace{ a_{x,l} \frac{\d  a^\top_{x,i}}{\d x_j}}_{d \times d} + \underbrace{ \frac{\d a_{x,l}}{\d x_j}  a^\top_{x,i}}_{d \times d} \\
        = & ~  -a_{x,l} (A_{x, i,j} A_{x, i,*})^\top    - A_{x, l,j} A_{x, l,*}  a^\top_{x,i}\\
        = & ~  -a_{x,l} (A_{x, i,j} a_{x,i})^\top    - A_{x, l,j} a_{x,l}  a^\top_{x,i}\\
        = & ~ -a_{x,l} A_{x, i,j} a_{x,i}^\top    - A_{x, l,j} a_{x,l}  a^\top_{x,i}\\
        = & ~ - (A_{x, i,j} + A_{x, l,j}) a_{x,l} a_{x,i}^\top
    \end{align*}
where the first step follows from the product rule in Fact~\ref{fac:derivative_rules}, the second step follows from the {\bf Part 1} , the third step follows from the definition of $a_{x,i}$ (see Definition~\ref{def:A_x}),  the fourth step follows from simple algebra, and the last step follows from simple  algebra.
\end{proof}

\subsection{Gradients Related to Scalar $A_{x, i, j}$ and Vector $A_{x, *, j}$}
\label{sub:gradient_s_A:AA}

\begin{lemma}\label{lem:gradient_Axij_and_Axj}
    If the following conditions hold
\begin{itemize}
    \item Let $A_x \in \R^{n \times d} $ be defined as in Definition~\ref{def:A_x}.
    \item Let $A_{x, i, j} \in \R$ denote the entry of $A_x \in \R^{n\times d}$ located at the $i$-th row and $j$-th column.
        \item Let $A_{x, *, j}  \in \R^n$ denote the $j$-th column of $A_x  \in \R^{n \times d} $.
\end{itemize}
Then, we have for each $j \in [d]$,
\begin{itemize}
    \item {\bf Part 1.}
    \begin{align*}
        \underbrace{ \frac{\d A_{x,i,j}}{\d x_j} }_{ \mathrm{scalar} } =  -  A_{x,i,j}^2 
    \end{align*}
    \item {\bf Part 2.}
    \begin{align*}
        \underbrace{ \frac{\d A_{x,i,j}}{\d x_k} }_{ \mathrm{scalar} } =  - A_{x, i,k} A_{x,i,j}
    \end{align*}    
    \item {\bf Part 3.}
    \begin{align*}
        \frac{ \d A_{x,*,j} }{\d x_j} = - A_{x, *,j}^{\circ 2}
    \end{align*}
    \item {\bf Part 4.}
    \begin{align*}
        \frac{ \d A_{x,*,j} }{\d x_k} = - A_{x, *,k} \circ A_{x,*,j}
    \end{align*}
\end{itemize}
\end{lemma}
\begin{proof}
{\bf Proof of Part 1.}
    \begin{align*}
        \underbrace{ \frac{\d A_{x,i,j}}{\d x_j} }_{ \mathrm{scalar} } 
        = &~ \underbrace{ (\frac{\d A_x}{\d x_j})_{i,j} }_{ \mathrm{scalar} } \\
        = &~  ( - \diag(A_{x, *,j}) A_x )_{i,j} \\
        = &~   -  A_{x,i,j}^2,
    \end{align*}
where the first step follows from matrix calculus, the second step follows from the {\bf Part 1} of Lemma~\ref{lem:gradient_A}, and the third step follows from simple algebra.

{\bf Proof of Part 2.}
    \begin{align*}
        \underbrace{ \frac{\d A_{x,i,j}}{\d x_k} }_{ \mathrm{scalar} } 
        = &~ \underbrace{ (\frac{\d A_x}{\d x_k})_{i,j} }_{ \mathrm{scalar} } \\
        = &~  ( - \diag(A_{x, *,k}) A_x )_{i,j} \\
        = &~   - A_{x, i,k} A_{x,i,j},
    \end{align*}
where the first step follows from matrix calculus, the second step follows from the {\bf Part 1} of Lemma~\ref{lem:gradient_A}, the third step follows from simple algebra. 

{\bf Proof of Part 3.}
    \begin{align*}
        \underbrace{ \frac{\d A_{x,*,j}}{\d x_j} }_{ 1 \times d } 
        = &~ \underbrace{ (\frac{\d A_x}{\d x_j})_{*,j} }_{ 1 \times d } \\
        = &~  ( - \diag(A_{x, *,j}) A_x )_{*,j} \\
        = &~   - A_{x, *,j}^{\circ 2},
    \end{align*}
where the first step follows from matrix calculus, the second step follows from the {\bf Part 1} of Lemma~\ref{lem:gradient_A}, the third step follows from simple algebra. 

{\bf Proof of Part 4.}
    \begin{align*}
        \underbrace{ \frac{\d A_{x,*,j}}{\d x_k} }_{ 1 \times d } 
        = &~ \underbrace{ (\frac{\d A_x}{\d x_k})_{*,j} }_{ 1 \times d } \\
        = &~  ( - \diag(A_{x, *,k}) A_x )_{*,j} \\
        = &~   - A_{x, *,k} \circ A_{x,*,j},
    \end{align*}
where the first step follows from matrix calculus, the second step follows from the {\bf Part 1} of Lemma~\ref{lem:gradient_A}, the third step follows from simple algebra. 
\end{proof}

\subsection{Gradient for Scalar $\sigma_{i,i}$ and Scalar $\sigma_{i,j}$}

\label{sub:gradient_hessian_sigma_Sigma:scalar}

\begin{lemma}\label{lem:first_derivative_of_leverage_score}
If the following conditions hold
\begin{itemize}
    \item Let $A_x \in \R^{n \times d} $ be defined as in Definition~\ref{def:A_x}.
    \item Let $\sigma_{*, i}(x) \in \R^{n}$, $\sigma_{i, i} (x)  \in \R$, and $\sigma_{i, l} (x)  \in \R$ be defined as in Definition~\ref{def:sigma}.
        \item Let $A_{x, *, j}  \in \R^n$ denote the $j$-th column of $A_x  \in \R^{n \times d} $.
\end{itemize}
Then we have for each $j \in [d]$
\begin{itemize}
    \item Part 1. For each $i \in [n]$,  
    \begin{align*}
        \frac{ \d \sigma_{i,i}( x ) }{ \d x_j } = 2 \langle \sigma_{*,i}^{\circ 2}(x), A_{x,*,j} \rangle
    \end{align*}
    \item Part 2. For each $i \in [n]$, $l \in [n]$ 
    \begin{align*}
        \frac{ \d \sigma_{i,l}( x ) }{ \d x_j } = 2 \langle \sigma_{*,i}(x) \circ \sigma_{*,l}(x), A_{x,*,j} \rangle   
\end{align*}
 
\end{itemize}

\end{lemma}

\begin{proof}

{\bf Proof of Part 1.}

We know 
\begin{align*}
    \frac{\d \sigma_{i,i} ( x )}{ \d x_j }
    = & ~ \frac{ \d a_{x,i}^\top ( A_x^\top A_x )^{-1}  a_{x,i} } { \d x_j } \\
   = & ~ \frac{ \d \langle a_{x,i} a_{x,i}^\top, (A_x^\top A_x)^{-1} \rangle }{\d x_j} \\
   = & ~ \langle \frac{ \d a_{x,i} a_{x,i}^\top }{\d x_j }, (A_x^\top A_x)^{-1} \rangle + \langle a_{x,i} a_{x,i}^\top, \frac{\d  (A_x^\top A_x)^{-1} }{\d x_j} \rangle,
\end{align*}
where the initial step arises from Definition~\ref{def:sigma}, the subsequent step is derived from Fact~\ref{fac:vector}, and the final step comes from Fact~\ref{fac:derivative_rules}.

For the first term, in above, we have
\begin{align*} 
\langle \frac{ \d a_{x,i} a_{x,i}^\top }{\d x_j }, (A_x^\top A_x)^{-1} \rangle
= & ~ \langle - 2 \cdot A_{x,i,j}  \cdot  a_{x,i} a_{x,i}^\top , (A_x^\top A_x)^{-1} \rangle\\
= & ~ -2 A_{x,i,j} \cdot \langle a_{x,i} a_{x,i}^\top, (A_x^\top A_x)^{-1} \rangle \\
= & ~ - 2A_{x,i,j} \cdot \sigma_{i,i}(x),
\end{align*}
where the initial step arises from {\bf Part 2} of Lemma~\ref{lem:gradient_a}, the subsequent step is derived from Fact~\ref{fac:vector}, and the final step follows from Definition~\ref{def:sigma}.

For the second term, we have
\begin{align*}
 \langle a_{x,i} a_{x,i}^\top, \frac{\d  (A_x^\top A_x)^{-1} }{\d x_j} \rangle
 = & ~ \langle a_{x,i} a_{x,i}^\top, 2 ( A_x^\top A_x )^{-1} \cdot A_x^\top \diag (  A_{x,*,j} ) A_{x} \cdot  ( A_x^\top A_x )^{-1}  \rangle\\
 = & ~ 2 \langle a_{x,i} a_{x,i}^\top, ( A_x^\top A_x )^{-1} \cdot A_x^\top \diag (  A_{x,*,j} ) A_{x} \cdot  ( A_x^\top A_x )^{-1} \rangle \\
 = & ~ 2 a_{x,i}^\top ( A_x^\top A_x )^{-1} \cdot A_x^\top \diag (  A_{x,*,j} ) A_{x} \cdot  ( A_x^\top A_x )^{-1} a_{x,i} \\
 = & ~ 2 a_{x,i}^\top ( A_x^\top A_x )^{-1} \cdot ( \sum_{l=1}^n a_{x,l} a_{x,l}^\top A_{x,l,j} ) \cdot  ( A_x^\top A_x )^{-1} a_{x,i} \\
 = & ~  2 \sum_{l=1}^n \sigma_{l,i}(x)^2 A_{x,l,j} \\
 = & ~ 2 \langle \sigma_{*,i}^{\circ 2}(x), A_{x,*,j} \rangle,
\end{align*}
where the initial step arises from {\bf Part 3} of Lemma~\ref{lem:gradient_A}, the subsequent step are derived from Fact~\ref{fac:vector}, the next step is based on property of inner product (see Fact~\ref{fac:inner_product_matrix}), the fourth step follows from the definition of $\diag(\cdot)$, the following step is derived from Definition~\ref{def:sigma}, and the final step comes from the definition of $\langle \cdot, \cdot \rangle$.

Thus, we complete the proof.

{\bf Proof of Part 2.}

We know 
\begin{align*}
    \frac{\d \sigma_{i,l} ( x )}{ \d x_j }
    = & ~ \frac{ \d a_{x,i}^\top ( A_x^\top A_x )^{-1} a_{x,l}  } { \d x_j } \\
   = & ~ \frac{ \d \langle a_{x,l} a_{x,i}^\top, (A_x^\top A_x)^{-1} \rangle }{\d x_j} \\
   = & ~ \langle \frac{ \d a_{x,l} a_{x,i}^\top }{\d x_j }, (A_x^\top A_x)^{-1} \rangle + \langle a_{x,l} a_{x,i}^\top, \frac{\d  (A_x^\top A_x)^{-1} }{\d x_j} \rangle,
\end{align*}
where the initial step arises from Definition~\ref{def:sigma}, the subsequent step is derived from Fact~\ref{fac:vector}, and the final step follows from Fact~\ref{fac:derivative_rules}.

For the first term, in above, we have
\begin{align*} 
\langle \frac{ \d a_{x,l} a_{x,i}^\top }{\d x_j }, (A_x^\top A_x)^{-1} \rangle
= & ~ \langle - (A_{x,l,j} + A_{x,i,j}) a_{x,l} a_{x,i}^\top, (A_x^\top A_x)^{-1} \rangle\\
= & ~ - (A_{x,l,j} + A_{x,i,j}) \cdot \langle a_{x,l} a_{x,i}^\top, (A_x^\top A_x)^{-1} \rangle \\
= & ~ - (A_{x,l,j} + A_{x,i,j}) \cdot \sigma_{i,l}(x),
\end{align*}
where the initial step arises from {\bf Part 3} of Lemma~\ref{lem:gradient_a}, the subsequent step is derived from Fact~\ref{fac:vector}, and the final step follows from Definition~\ref{def:sigma}.

About the second term, we get
\begin{align*}
 \langle a_{x,l} a_{x,i}^\top, \frac{\d  (A_x^\top A_x)^{-1} }{\d x_j} \rangle
 = & ~ \langle a_{x,l} a_{x,i}^\top, 2 ( A_x^\top A_x )^{-1} \cdot A_x^\top \diag (  A_{x,*,j} ) A_{x} \cdot  ( A_x^\top A_x )^{-1}  \rangle\\
 = & ~ 2 \langle a_{x,l} a_{x,i}^\top, ( A_x^\top A_x )^{-1} \cdot A_x^\top \diag (  A_{x,*,j} ) A_{x} \cdot  ( A_x^\top A_x )^{-1} \rangle \\
 = & ~ 2 a_{x,l}^\top ( A_x^\top A_x )^{-1} \cdot A_x^\top \diag (  A_{x,*,j} ) A_{x} \cdot  ( A_x^\top A_x )^{-1} a_{x,i} \\
 = & ~ 2 a_{x,l}^\top ( A_x^\top A_x )^{-1} \cdot ( \sum_{l=1}^n a_{x,l} a_{x,l}^\top A_{x,l,j} ) \cdot  ( A_x^\top A_x )^{-1} a_{x,i} \\
 = & ~  2 \sum_{l=1}^n \sigma_{l,l}(x) \sigma_{l,i}(x) A_{x,l,j} \\
 = & ~ 2 \langle \sigma_{*,i}(x) \circ \sigma_{*,l}(x), A_{x,*,j} \rangle,
\end{align*}
where the 1st step is due to {\bf Part 3} of Lemma~\ref{lem:gradient_A}, the 2nd step arises from Fact~\ref{fac:vector}, the 3rd step can be seen by Fact~\ref{fac:vector}, the fourth step follows from the definition of $\diag(\cdot)$, the fifth step follows from Definition~\ref{def:sigma}, and the last step follows from the definition of $\langle \cdot, \cdot \rangle$.
\end{proof}

\subsection{Gradient for Vector $\sigma_{*,i}$}

\label{sub:gradient_hessian_sigma_Sigma:vector}

\begin{lemma}\label{lem:gradient_sigma_*i}
If the following conditions hold
\begin{itemize}
    \item Let $A_x \in \R^{n \times d} $ be defined as in Definition~\ref{def:A_x}.
    \item Let $\sigma_{*, *}(x) \in \R^{n \times n}$, $\sigma_{*, i}(x) \in \R^n$, $\sigma_{i, i} (x)\in \R$, and $\sigma_{i, l} (x)\in \R$ be defined as in Definition~\ref{def:sigma}.
        \item Let $A_{x, *, j}  \in \R^n$ denote the $j$-th column of $A_x  \in \R^{n \times d} $.
    \item Let $A_{x, i, j}  \in \R$ denote the entry of $A_x\in \R^{n \times d}$ located at the $i$-th row and $j$-th column.
\end{itemize}
Then, we have 
    \begin{align*}
        \frac{ \d \sigma_{*,i}(x) }{ \d x_j} = -  A_{x,*,j} \circ \sigma_{*,i}(x) +  2 \sigma_{*,*}(x)  (\sigma_{*,i}(x) \circ A_{x,*,j}) - \sigma_{*,i}(x) A_{x,i,j}
    \end{align*}
\end{lemma}

\begin{proof}
We have
\begin{align*}
    & ~ \frac{ \d \sigma_{*,i}(x) }{ \d x_j} \\
    = & ~ \frac{ \d A_x (A_x^\top A_x)^{-1} a_{x,i}}{ \d x_j} \\
    = & ~ \frac{ \d A_x }{ \d x_j} (A_x^\top A_x)^{-1} a_{x,i} +  A_x \frac{ \d(A_x^\top A_x)^{-1}}{ \d x_j} a_{x,i} + A_x (A_x^\top A_x)^{-1} \frac{ \d a_{x,i}}{ \d x_j} \\
    = & ~ - \diag( A_{x,*,j} ) A_x  (A_x^\top A_x)^{-1} a_{x,i} +  2 A_x ( A_x^\top A_x )^{-1} \cdot A_x^\top \diag (  A_{x,*,j} ) A_{x} \cdot  ( A_x^\top A_x )^{-1}  a_{x,i} \\
    - & ~ A_x (A_x^\top A_x)^{-1} A_{x,i,j} a_{x,i} \\
    = & ~ - \diag( A_{x,*,j} ) A_x  (A_x^\top A_x)^{-1} a_{x,i} +  2 \sigma_{*,*}(x) \diag (  A_{x,*,j} ) A_{x} \cdot  ( A_x^\top A_x )^{-1}  a_{x,i} - A_x (A_x^\top A_x)^{-1} A_{x,i,j} a_{x,i} \\
    = & ~ - \diag( A_{x,*,j} ) \sigma_{*,i}(x) +  2 \sigma_{*,*}(x) \diag (  A_{x,*,j} ) \sigma_{*,i}(x) - \sigma_{*,i}(x) A_{x,i,j}\\
    = & ~ -  A_{x,*,j} \circ \sigma_{*,i}(x) +  2 \sigma_{*,*}(x)  (\sigma_{*,i}(x) \circ A_{x,*,j}) - \sigma_{*,i}(x) A_{x,i,j},
\end{align*}
where the 1st step is by how we define $\sigma_{*,i}(x)$ (see Definition~\ref{def:sigma}), the 2nd step is by the product rule, the third step follows from {\bf Part 1} and {\bf Part 3} from Lemma~\ref{lem:gradient_A} and {\bf Part 1} from Lemma~\ref{lem:gradient_a}, the fourth step follows from the definition of $\sigma_{*,*}(x)$ (see Definition~\ref{def:sigma}), the fifth step follows from the definition of $\sigma_{*,i}(x)$ (see Definition~\ref{def:sigma}), and the last step follows from Fact~\ref{fac:vector}.
\end{proof}

\subsection{Gradient for Vector $p(x)$}\label{sub:gradient_p}
\begin{lemma}\label{lem:p(x)}
If the following conditions hold
\begin{itemize}
        \item Let $b \in \R^n$ 
        be defined as in Definition~\ref{def:loss_lb}.
        \item Let $c \in \R^d$ be defined as in Definition~\ref{def:L_c}.
        \item Let $A_x \in \R^{n \times d} $ be defined as in Definition~\ref{def:A_x}.
        \item Let $\sigma_{*,*}^{\circ 2}(x) \in \R^{n \times n}$, $\sigma_{*,*}(x)  \in \R^{n \times n}$, and $\sigma_{*, i}(x)  \in \R^n $ be defined as in Definition~\ref{def:sigma}.
        \item Let $\Sigma(x)  \in \R^{n \times n} $ be defined as in Definition~\ref{def:Sigma}.
        \item Let $A_{x, *, j}  \in \R^n$ denote the $j$-th column of $A_x \in \R^{n \times d}$.
        \item Let $g(x) \in \R^d $ be defined as in Definition~\ref{def:g(x)}
        \item Let $p(x)  = \sigma_{*, i}(x) - b \in \R^n$.
        \item Let $q(x)  = g(x) - c  \in \R^d$.
\end{itemize}
Then, we have 
    \begin{align*}
        \frac{ \d p(x) }{ \d x_j}  = -  A_{x,*,j} \circ \sigma_{*,i}(x) +  2 \sigma_{*,*}(x)  (\sigma_{*,i}(x) \circ A_{x,*,j}) - \sigma_{*,i}(x) A_{x,i,j} 
    \end{align*}
\end{lemma}

\begin{proof}
We have
\begin{align*}
    \frac{  \d p(x)}{ \d x_j} =& ~ \frac{\d \sigma_{*,i}(x) - b}{\d x_j} \\
    = & ~ \frac{\d \sigma_{*,i}(x)}{\d x_j} \\
    = & ~ -  A_{x,*,j} \circ \sigma_{*,i}(x) +  2 \sigma_{*,*}(x)  (\sigma_{*,i}(x) \circ A_{x,*,j}) - \sigma_{*,i}(x) A_{x,i,j},
\end{align*}
where the first step follows from the definition of p(x) (see lemma statement), the second step follows from $\frac{\d b}{\d x_j} = 0$, and the final step follows from Lemma~\ref{lem:gradient_sigma_*i}.
\end{proof}

\subsection{Gradient for Matrix $\sigma_{*,*}(x)$ and $\sigma_{*,*}^{\circ 2}(x)$}
\label{sub:gradient_hessian_sigma_Sigma:matrix}

\begin{lemma}\label{lem:gradient_sigma_**}
If the following conditions hold
\begin{itemize}
    \item Let $A_x \in \R^{n \times d} $ be defined as in Definition~\ref{def:A_x}.
    \item Let $\sigma_{*, *} \in \R^{n \times n}$, $\sigma_{*, *}^{\circ 2}\in \R^{n \times n}$, $\sigma_{*, i}(x)\in \R^{ n}$, $\sigma_{i, i} (x)\in \R$, and $\sigma_{i, l} (x)\in \R$ be defined as in Definition~\ref{def:sigma}.
        \item Let $A_{x, *, j}  \in \R^n$ denote the $j$-th column of $A_x  \in \R^{n \times d} $.
\end{itemize}
Then, we have
\begin{itemize}
    \item Part 1. For all $j \in [d]$,
    \begin{align*}
        \frac{ \d \sigma_{*,*}(x) }{ \d x_j} = 2 \sigma_{*,*}(x) \diag (  A_{x,*,j} ) \sigma_{*,*}(x) - \diag( A_{x,*,j} )  \sigma_{*,*}(x) - \sigma_{*,*}(x) \diag( A_{x,*,j} )  
    \end{align*}
    \item Part 2. For all $j \in [d]$,
    \begin{align*}
        \frac{ \d \sigma_{*,*}^{\circ 2}(x) }{ \d x_j} =& ~ 4 \sigma^2_{*,*}(x) \diag (  A_{x,*,j} ) \sigma_{*,*}(x)  \\
   & ~ - 2 \diag( A_{x,*,j} )  \sigma_{*,*}^{\circ 2}(x) \\
    & ~ - 2\sigma^{\circ 2}_{*,*}(x) \diag( A_{x,*,j} )
    \end{align*}
\end{itemize}
\end{lemma}

\begin{proof}
{\bf Proof of Part 1.}

We have
\begin{align*}
    \frac{ \d \sigma_{*,*}(x) }{ \d x_j}
    = & ~  \frac{ \d A_x (A_x^\top A_x)^{-1} A_x^\top }{ \d x_j}  \\
    = & ~  \frac{ \d A_x }{ \d x_j} (A_x^\top A_x)^{-1} A_x^\top  + A_x \frac{ \d (A_x^\top A_x)^{-1} }{ \d x_j} A_x^\top  + A_x (A_x^\top A_x)^{-1} \frac{ \d A_x^\top }{ \d x_j} \\
    = & ~  - \diag( A_{x,*,j} )  A_x  (A_x^\top A_x)^{-1} A_x^\top  + 2 A_x ( A_x^\top A_x )^{-1} \cdot A_x^\top \diag (  A_{x,*,j} ) A_{x} \cdot  ( A_x^\top A_x )^{-1} A_x^\top  \\
    + & ~ A_x (A_x^\top A_x)^{-1} (-\diag( A_{x,*,j} ) A_x)^\top\\
    = & ~  - \diag( A_{x,*,j} )  A_x  (A_x^\top A_x)^{-1} A_x^\top  + 2 A_x ( A_x^\top A_x )^{-1} \cdot A_x^\top \diag (  A_{x,*,j} ) A_{x} \cdot  ( A_x^\top A_x )^{-1} A_x^\top  \\
    - & ~ A_x (A_x^\top A_x)^{-1} A_x^\top \diag( A_{x,*,j} ) \\
    = & ~  - \diag( A_{x,*,j} )  \sigma_{*,*}(x)  + 2 \sigma_{*,*}(x) \diag (  A_{x,*,j} ) \sigma_{*,*}(x) - \sigma_{*,*}(x) \diag( A_{x,*,j} ) \\
    = & ~ 2 \sigma_{*,*}(x) \diag (  A_{x,*,j} ) \sigma_{*,*}(x) - \diag( A_{x,*,j} )  \sigma_{*,*}(x) - \sigma_{*,*}(x) \diag( A_{x,*,j} ),
\end{align*}
where the initial step arises from the definition of $\sigma_{*,*}(x)$ (refer to Definition~\ref{def:sigma}), the subsequent step follows from the product rule, the next step is derived from {\bf Part 1} and {\bf Part 3} of Lemma~\ref{lem:gradient_A}, the following step is based on the properties $(AB)^\top = B^\top A^\top$ and the symmetry of $\diag(x)$, where $x$ is a vector, the subsequent step arises from the definition of $\sigma_{*,*}(x)$ (refer to Definition~\ref{def:sigma}), and the final step involves reorganizing the terms.

{\bf Proof of Part 2.}
\begin{align*}
    \frac{ \d \sigma_{*,*}^{\circ 2}(x) }{\d x}
    = & ~  2 \sigma_{*,*}(x) \circ  \frac{ \d \sigma_{*,*}(x) }{\d x} \\
    = & ~ 2 \sigma_{*,*}(x) \circ  (2 \sigma_{*,*}(x) \diag (  A_{x,*,j} ) \sigma_{*,*}(x) - \diag( A_{x,*,j} )  \sigma_{*,*}(x) - \sigma_{*,*}(x) \diag( A_{x,*,j} ))  \\
    =  & ~ 4 \sigma^2_{*,*}(x) \diag (  A_{x,*,j} ) \sigma_{*,*}(x)  \\
   & ~ - 2 \diag( A_{x,*,j} )  \sigma_{*,*}^{\circ 2}(x) \\
    & ~ - 2\sigma^{\circ 2}_{*,*}(x) \diag( A_{x,*,j} )
\end{align*}
where the first step follows from the definition of $\sigma_{*,*}(x)$ (refer to Definition~\ref{def:sigma}) and the power rule in Fact~\ref{fac:derivative_rules}, the second step follows from the {\bf Part1}, and the final step follows from simple algebra. 

\end{proof}

\subsection{Gradient for Scalar $f(x)_i$} \label{sub:f(x)}

\begin{lemma}\label{lem:f(x)}
    If the following conditions hold
    \begin{itemize}
        \item Let $f(x) \in \R^n$ be defined as in Definition~\ref{def:f(x)}
        \item Let $A_x \in \R^{n \times d} $ be defined as in Definition~\ref{def:A_x}.
        \item Let $\sigma_{*, i}(x) \in \R^n  $ and $\sigma_{i, i} (x) \in \R$ be defined as in Definition~\ref{def:sigma}.
        \item Let $A_{x, *, j}  \in \R^n$ denote the $j$-th column of $A_x  \in \R^{n \times d} $.
        \item Let $A_{x, i, j} \in \R$ denote the entry of $A_x \in \R^{n \times d}$ located at the $i$-th row and $j$-th column.
    \end{itemize}

    Then, we have for all $j \in [d]$,
    \begin{align*}
        \frac{ \d f(x)_i}{ \d x_j} = 2 \sigma_{i,*}(x) \diag (  A_{x,*,j} ) \sigma_{*,i}(x) -  2A_{x,i,j} \sigma_{i,i}(x)
    \end{align*}
\end{lemma}
\begin{proof}
    
We have
\begin{align*}
    \frac{ \d f(x)_i}{ \d x_j}
    = & ~ \frac{ \d \Sigma(x)_{i,i}}{ \d x_j}\\
    = & ~ \frac{ \d (\sigma_{*,*}(x) \circ I_n)_{i,i}}{ \d x_j}\\
    = & ~ (\frac{ \d (\sigma_{*,*}(x) \circ I_n)}{ \d x_j})_{i,i}\\
    = & ~ (\frac{ \d (\sigma_{*,*}(x))}{ \d x_j} \circ I_n)_{i,i}\\
    = & ~ ((2 \sigma_{*,*}(x) \diag (  A_{x,*,j} ) \sigma_{*,*}(x) - \diag( A_{x,*,j} )  \sigma_{*,*}(x) - \sigma_{*,*}(x) \diag( A_{x,*,j} )) \circ I_n)_{i,i}\\
    = & ~ (2 \sigma_{*,*}(x) \diag (  A_{x,*,j} ) \sigma_{*,*}(x) - \diag( A_{x,*,j} )  \sigma_{*,*}(x) - \sigma_{*,*}(x) \diag( A_{x,*,j} ))_{i,i}\\
    = & ~ (2 \sigma_{*,*}(x) \diag (  A_{x,*,j} ) \sigma_{*,*}(x))_{i,i} - (\diag( A_{x,*,j} )  \sigma_{*,*}(x))_{i,i} - (\sigma_{*,*}(x) \diag( A_{x,*,j} ))_{i,i}\\
    = & ~ 2 \sigma_{i,*}(x) \diag (  A_{x,*,j} ) \sigma_{*,i}(x) -  A_{x,i,j} \sigma_{i,i}(x) - \sigma_{i,i}(x) A_{x,i,j} \\
    = & ~ 2 \sigma_{i,*}(x) \diag (  A_{x,*,j} ) \sigma_{*,i}(x) -  2A_{x,i,j} \sigma_{i,i}(x),
\end{align*}
where the first step follows from the definition of $f(x)_i$ (see Definition~\ref{def:f(x)}), the second step follows from the definition of $\Sigma(x)$ (see Definition~\ref{def:Sigma}), the third step follows from simple algebra, the fourth step follows from the product rule, the fifth step follows from Lemma~\ref{lem:gradient_sigma_**}, the sixth step follows from $(I_n)_{i, i} = 1$, the seventh step follows from simple algebra, the eighth and the final steps follow from simple algebra.
\end{proof}

\subsection{Gradient for Scalar $L_b(x)$}
\label{sub:L_b}

\begin{lemma}\label{lem:L_b}
If the following conditions hold
    \begin{itemize}
        \item Let $L_b \in \R$ and $  b \in \R^n$ be defined as in Definition~\ref{def:loss_lb}.
        \item Let $A_x \in \R^{n \times d} $ be defined as in Definition~\ref{def:A_x}.
        \item Let $\sigma_{*,*}^{\circ 2}(x) \in \R^{n \times n}$, $\sigma_{*,*}(x)  \in \R^{n \times n}$, and $\sigma_{*, i}(x)  \in \R^n $ be defined as in Definition~\ref{def:sigma}.
        \item Let $\Sigma(x)  \in \R^{n \times n} $ be defined as in Definition~\ref{def:Sigma}.
        \item Let $A_{x, *, j}  \in \R^n$ denote the $j$-th column of $A_x  \in \R^{n \times d} $.
        \item Let $p(x) = \sigma_{*, i}(x) - b  \in \R^n$.
    \end{itemize}

Then, we have for all $j \in [d]$,
\begin{itemize}
    \item Part 1.  
            \begin{align*}
     \frac{ \d L_b(x) }{ \d x_j} = 2 \underbrace{A_{x,*,j}^\top}_{1 \times n} (\underbrace{\sigma_{*,*}^{\circ 2}(x)}_{n \times n} - \underbrace{\Sigma(x)}_{n \times n}) \underbrace{p(x)}_{n \times 1}
        \end{align*}
    \item Part 2. 
    \begin{align*}
       \underbrace{ \frac{\d L_b(x)}{\d x} }_{d \times 1}
        = 2 \underbrace{A_{x}^\top}_{d \times n} (\underbrace{\sigma_{*,*}^{\circ 2}(x)}_{n \times n} - \underbrace{\Sigma(x)}_{n \times n}) \underbrace{p(x)}_{n \times 1}
    \end{align*}
\end{itemize}

\end{lemma}

\begin{proof}

{\bf Proof of Part 1.}

    We have
    \begin{align*}
       \underbrace{ \frac{ \d L_b(x) }{ \d x_j} }_{\mathrm{scalar}} = &~ \underbrace{ \frac{ \d 0.5 \| f(x) - b \|_2^2  }{ \d x_j}  }_{\mathrm{scalar}}  \\
        =&~ \frac{ \d 0.5 \sum^n_{i = 1} ( f(x)_i - b_i )^2  }{ \d x_j}  \\
        =&~ 0.5 \sum^n_{i = 1} \frac{ \d  ( f(x)_i - b_i )^2  }{ \d x_j}  \\
        =&~ 0.5 \sum^n_{i = 1} 2( f(x)_i - b_i ) \frac{ \d   f(x)_i - b_i   }{ \d x_j}  \\
        =&~  \sum^n_{i = 1} ( f(x)_i - b_i ) \frac{ \d   f(x)_i - b_i   }{ \d x_j}  \\
        =&~  \sum^n_{i = 1} ( f(x)_i - b_i ) \frac{ \d   f(x)_i  }{ \d x_j}  \\
        =&~  \sum^n_{i = 1} ( f(x)_i - b_i ) (2 \sigma_{i,*}(x) \diag( A_{x,*,j}) \sigma_{*,i}(x) - 2 A_{x,i,j}   \sigma_{i,i}(x) )  \\
        = &~  \sum^n_{i = 1} ( \sigma_{i,i}(x) - b_i ) (2 \sigma_{i,*}(x) \diag( A_{x,*,j}) \sigma_{*,i}(x) - 2 A_{x,i,j}   \sigma_{i,i}(x) )  
    \end{align*}
    where the first step follows from the Definition~\ref{def:f(x)}, the second step and the third step follow from simple algebra, the fourth step follows from the power rule in Fact~\ref{fac:derivative_rules},  the fifth step follows simple algebra, the sixth step follows from $\frac{ \d b }{ \d x_j } = 0$, the seventh step follows lemma~\ref{lem:f(x)}, and the last step follows from the definition of $f(x)_i$ (see Definition~\ref{def:f(x)} and the definition of $\Sigma(x)$).

    Therefore, we have
    \begin{align}\label{eq:Lb/xj}
        \frac{ \d L_b(x) }{ \d x_j} 
        = & ~ \sum^n_{i = 1} p(x)_i (2 \sigma_{*,i}(x)^\top \diag( A_{x,*,j}) \sigma_{*,i}(x) - 2 A_{x,i,j}   \sigma_{i,i}(x) ) \notag\\
        = & ~ \sum^n_{i = 1}  p(x)_i (2  \underbrace{( \sigma_{*,i}^{\circ 2}(x) )^\top}_{1 \times n} \underbrace{A_{x,*,j}}_{n \times 1} - 2 A_{x,i,j}   \sigma_{i,i}(x) ) \notag\\
        = & ~ 2 \sum^n_{i = 1}  p(x)_i \underbrace{( \sigma_{*,i}^{\circ 2}(x) )^\top}_{1 \times n} \underbrace{A_{x,*,j}}_{n \times 1} - p(x)_i A_{x,i,j}   \sigma_{i,i}(x)  \notag\\
        = & ~ 2 \sum^n_{i = 1}  (p(x)_i \underbrace{( \sigma_{*,i}^{\circ 2}(x) )^\top}_{1 \times n} \underbrace{A_{x,*,j}}_{n \times 1} - p(x)_i A_{x,i,j}   \sigma_{i,i}(x))^\top  \notag\\
        = & ~ 2 \sum^n_{i = 1}  \underbrace{A_{x,*,j}^\top}_{1 \times n} \underbrace{\sigma_{*,i}^{\circ 2}(x) }_{n \times 1} p(x)_i - p(x)_i A_{x,i,j}   \sigma_{i,i}(x) \notag\\
        = & ~ 2 \sum^n_{i = 1}  \underbrace{A_{x,*,j}^\top}_{1 \times n} \underbrace{\sigma_{*,i}^{\circ 2}(x) }_{n \times 1} p(x)_i - 2 \sum^n_{i = 1} p(x)_i A_{x,i,j}   \sigma_{i,i}(x),
    \end{align}
    where the second step follows from Fact~\ref{fac:vector}, the third step follows from simple algebra, the fourth step follows from $a^\top = a$, for all $a \in \R$, the fifth step follows from $(AB)^\top = B^\top A^\top$, and the last step follows from simple algebra.

    Considering the first term of Eq.~\eqref{eq:Lb/xj}, we have
    \begin{align*}
        2 \sum^n_{i = 1}  \underbrace{A_{x,*,j}^\top}_{1 \times n} \underbrace{\sigma_{*,i}^{\circ 2}(x) }_{n \times 1} p(x)_i
        = & ~ 2 \underbrace{A_{x,*,j}^\top}_{1 \times n} \sum^n_{i = 1} \underbrace{\sigma_{*,i}^{\circ 2}(x) }_{n \times 1} p(x)_i\\
        = & ~ 2 \underbrace{A_{x,*,j}^\top}_{1 \times n} \underbrace{\sigma_{*,*}^{\circ 2}(x) }_{n \times n} \underbrace{p(x)}_{n \times 1},
    \end{align*}
    where the first step follows from the fact that $A_{x,*,j}^\top$ is independent of $i$ and the second step follows from the definition of the linear combination. 

    Considering the second term of Eq.~\eqref{eq:Lb/xj}, we have
    \begin{align*}
        2 \sum^n_{i = 1} p(x)_i A_{x,i,j}   \sigma_{i,i}(x)
        = & ~ 2 \langle \underbrace{A_{x,*,j}}_{n \times 1}, \underbrace{\Sigma(x)}_{n \times n} \underbrace{p(x)}_{n \times 1} \rangle\\
        = & ~ 2 \underbrace{A_{x,*,j}^\top}_{1 \times n} \underbrace{\Sigma(x)}_{n \times n} \underbrace{p(x)}_{n \times 1},
    \end{align*}
    where the first step follows from the definition of the inner product and $(\Sigma(x) p(x))_i = \sigma_{i,i}(x) p(x)_i$ by the definition of $\Sigma(x)$ and the second step follows from Fact~\ref{fac:vector}.

    Combining the first term and the second term together, we have
    \begin{align*}
        \frac{ \d L_b(x) }{ \d x_j} 
        = & ~ 2 \underbrace{A_{x,*,j}^\top}_{1 \times n} \underbrace{\sigma_{*,*}^{\circ 2}(x) }_{n \times n} \underbrace{p(x)}_{n \times 1} - 2 \underbrace{A_{x,*,j}^\top}_{1 \times n} \underbrace{\Sigma(x)}_{n \times n} \underbrace{p(x)}_{n \times 1}\\
        = & ~  2 \underbrace{A_{x,*,j}^\top}_{1 \times n} (\underbrace{\sigma_{*,*}^{\circ 2}(x)}_{n \times n} - \underbrace{\Sigma(x)}_{n \times n}) \underbrace{p(x)}_{n \times 1},
    \end{align*}
    where the second step follows from simple algebra.

    {\bf Proof of Part 2.}

    From {\bf Part 1}, we have
    \begin{align*}
        \frac{\d L_b(x)}{\d x} 
        = & ~ 2 A_{x}^\top \sigma_{*,*}^{\circ 2}(x) p(x) - 2A_{x}^\top \Sigma(x) p(x)\\
        = & ~ 2 A_{x}^\top (\sigma_{*,*}^{\circ 2}(x) - \Sigma(x)) p(x),
     \end{align*}
     where the second step follows from simple algebra.
\end{proof}

\subsection{Gradient for $\Sigma(x)$}\label{sub:sigma}

\begin{lemma}\label{lem:Sigma}
If the following conditions hold
    \begin{itemize}
        \item Let $b \in \R^n$ 
        be defined as in Definition~\ref{def:loss_lb}.
        \item Let $c \in \R^d$ be defined as in Definition~\ref{def:L_c}.
        \item Let $A_x \in \R^{n \times d} $ be defined as in Definition~\ref{def:A_x}.
        \item Let $\sigma_{*,*}^{\circ 2}(x) \in \R^{n \times n}$, $\sigma_{*,*}(x)  \in \R^{n \times n}$, and $\sigma_{*, i}(x)  \in \R^n $ be defined as in Definition~\ref{def:sigma}.
        \item Let $\Sigma(x)  \in \R^{n \times n} $ be defined as in Definition~\ref{def:Sigma}.
        \item Let $A_{x, *, j}  \in \R^n$ denote the $j$-th column of $A_x  \in \R^{n \times d} $.
        \item Let $g(x) \in \R^d $ be defined as in Definition~\ref{def:g(x)}
        \item Let $p(x) = \sigma_{*, i}(x) - b  \in \R^n$.
        \item Let $q(x) = g(x) - c \in \R^d $.
    \end{itemize}
Then, we have
    \begin{align*}
    \frac{\d \Sigma(x)}{\d x_j} =&~ 2 \sigma_{*,*}(x) \diag (  A_{x,*,j} )\Sigma(x)\\
       &~  -2 \diag( A_{x,*,j} )  \Sigma(x).
    \end{align*}
\end{lemma}
\begin{proof}
    \begin{align*}
        \frac{\d \Sigma(x)}{\d x_j} = &~ \frac{\d \sigma_{*,*}(x) \circ I_n}{\d x_j} \\
        =&~ \frac{\d \sigma_{*,*}(x) }{\d x_j} \circ I_n \\
        =&~ 2 \sigma_{*,*}(x) \diag (  A_{x,*,j} ) \sigma_{*,*}(x)\circ I_n\\
       &~  - \diag( A_{x,*,j} )  \sigma_{*,*}(x)  \circ I_n\\
       &~ - \sigma_{*,*}(x) \diag( A_{x,*,j}) \circ I_n \\
         =&~ 2 \sigma_{*,*}(x) \diag (  A_{x,*,j} )\Sigma(x)\\
       &~  - 2\diag( A_{x,*,j} )  \Sigma(x)\\
        \end{align*}
where the first step follows from the definition of $\Sigma(x)$(see Definition \ref{def:Sigma}), the second step follows from constant multiple rule of Hadamard product(see Fact \ref{fac:derivative_rules}), the third step follows from the definition of $\Sigma(x)$(see Definition \ref{def:Sigma}) and the final step follows from simple algebra.
\end{proof}

\begin{definition}\label{def:A_i}
    We define 
        \begin{align*}
A_1 := & ~ -10 A_x^\top \sigma_{*,*}^{\circ 2}(x) \diag(p(x))    A_{x, *,j}  \\
A_2 := & ~ -2  A_x^\top \Sigma(x) \diag(p(x)) A_{x, *,j}   \\
A_3 := & ~ + 8 A_x^\top \sigma_{*,*}(x)  \sigma_{*,*}^{\circ 2}(x) \diag( p(x) )  A_{x,*,j}
 \\
A_4 := & ~ - 4 A_x^\top \Sigma(x)  \sigma_{*,*}(x) \diag( p(x) )  A_{x, *,j}\\
A_5 := & ~ - 2 A_x^\top \sigma_{*,*}^{\circ 2}(x) \diag(\sigma_{*,i}(x) )A_{x, *,j}\\
A_6 := & ~ + 2 A_x^\top  \Sigma(x) \diag(\sigma_{*,i}(x)) A_{x, *,j}\\
A_7 := & ~ + 4 A_x^\top  \sigma_{*,*}^{\circ 2}(x) \sigma_{*,*}(x)   \diag(\sigma_{*,i}(x))A_{x, *,j}\\
A_8 := & ~ - 4 A_x^\top  \Sigma(x)  \sigma_{*,*}(x)    \diag(\sigma_{*,i}(x))A_{x, *,j}\\
A_9 := & ~ - 2A_{x}^\top \sigma_{*,*}^{\circ 2}(x) \sigma_{*,i}(x) A_{x,i,j}^\top = - 2A_{x}^\top \sigma_{*,*}^{\circ 2}(x) \sigma_{*,i}(x) e_i^\top A_{x,*,j}  \\
A_{10} := & ~ +2A_{x}^\top \Sigma(x) \sigma_{*,i}(x)  A_{x,i,j}^\top = +2A_{x}^\top \Sigma(x) \sigma_{*,i}(x)  e_i^\top A_{x,*,j}.
        \end{align*}        
and
        \begin{align*}
\wt{A}_1 := & ~ -10 A_x^\top \sigma_{*,*}^{\circ 2}(x) \diag(p(x))    A_{x}  \\
\wt{A}_2 := & ~ -2  A_x^\top \Sigma(x) \diag(p(x)) A_x   \\
\wt{A}_3 := & ~ + 8 A_x^\top \sigma_{*,*}(x)  \sigma_{*,*}^{\circ 2}(x) \diag( p(x) )  A_x
 \\
\wt{A}_4 := & ~ - 4 A_x^\top \Sigma(x)  \sigma_{*,*}(x) \diag( p(x) )  A_x\\
\wt{A}_5 := & ~ - 2 A_x^\top \sigma_{*,*}^{\circ 2}(x) \diag(\sigma_{*,i}(x) )A_x\\
\wt{A}_6 := & ~ + 2 A_x^\top  \Sigma(x) \diag(\sigma_{*,i}(x)) A_x\\
\wt{A}_7 := & ~ + 4 A_x^\top  \sigma_{*,*}^{\circ 2}(x) \sigma_{*,*}(x)   \diag(\sigma_{*,i}(x))A_x\\
\wt{A}_8 := & ~ - 4 A_x^\top  \Sigma(x)  \sigma_{*,*}(x)    \diag(\sigma_{*,i}(x))A_x\\
\wt{A}_9 := & ~  - 2A_{x}^\top \sigma_{*,*}^{\circ 2}(x) \sigma_{*,i}(x) e_i^\top A_x  \\
\wt{A}_{10} := & ~ +2A_{x}^\top \Sigma(x) \sigma_{*,i}(x)  e_i^\top A_x.
        \end{align*} 
\end{definition}

\subsection{Gradient for $g(x)$}\label{sub:g(x)}

\begin{lemma}\label{lem:g(x)}
    If the following conditions hold
    \begin{itemize}
        \item Let $b \in \R^n$ 
        be defined as in Definition~\ref{def:loss_lb}.
        \item Let $c \in \R^d$ be defined as in Definition~\ref{def:L_c}.
        \item Let $A_x \in \R^{n \times d} $ be defined as in Definition~\ref{def:A_x}.
        \item Let $\sigma_{*,*}^{\circ 2}(x) \in \R^{n \times n}$, $\sigma_{*,*}(x)  \in \R^{n \times n}$, and $\sigma_{*, i}(x)  \in \R^n $ be defined as in Definition~\ref{def:sigma}.
        \item Let $\Sigma(x)  \in \R^{n \times n} $ be defined as in Definition~\ref{def:Sigma}.
        \item Let $A_{x, *, j}  \in \R^n$ denote the $j$-th column of $A_x \in \R^{n \times d}$. 
        \item Let $g(x) \in \R^d $ be defined as in Definition~\ref{def:g(x)}
        \item Let $p(x) = \sigma_{*, i}(x) - b \in \R^n$.
        \item Let $q(x)  = g(x) - c \in \R^d$.
        \item Let $A_l  \in \R^d$ be defined in Definition~\ref{def:A_i} 
    \end{itemize}

    Then, for all $j \in [d]$, we have,
    \begin{itemize}

        \item Part 1.
        \begin{align*}
            \frac{\d g(x)}{\d x_j} =& ~ \sum_{l=1}^{10} A_l
        \end{align*}

        \item Part 2.
        \begin{align*}
            \frac{\d g(x)}{\d x} = \sum_{i = 1}^{10} \wt{A}_i.
        \end{align*}

    \end{itemize}

\end{lemma}

\begin{proof}

To get $\frac{\d g(x)}{\d x_j}$, we have
\begin{align*}
   \underbrace{ \frac{\d g(x)}{\d x_j} }_{d \times 1}
    = & ~ \frac{\d (2 A_{x}^\top (\sigma_{*,*}^{\circ 2}(x) - \Sigma(x)) p(x))}{\d x_j}\\
    = & ~\underbrace{ \frac{\d 2 A_{x}^\top}{\d x_j} }_{d\times n} \underbrace{ (\sigma_{*,*}^{\circ 2}(x) - \Sigma(x))}_{n\times n } \underbrace{ p(x)}_{n \times 1} + 2  \underbrace{A_{x}^\top}_{d\times n}  \underbrace{\frac{\d \sigma_{*,*}^{\circ 2}(x) - \Sigma(x)}{\d x_j}}_{n\times n }  \underbrace{p(x)}_{n \times 1} + 2  \underbrace{A_{x}^\top}_{d\times n}  \underbrace{(\sigma_{*,*}^{\circ 2}(x) - \Sigma(x))}_{n\times n }  \underbrace{\frac{\d p(x)}{\d x_j}}_{n \times 1}
\end{align*}
where the first step follows from the definition of $g(x)$ (see Definition~\ref{def:g(x)} and {\bf Part 3} of lemma~\ref{lem:L_b} and the second step follows from the product rule in Fact~\ref{fac:derivative_rules}.

We define 
\begin{align*}
    C_1 := & ~ 2 (\frac{\d A_{x}}{\d x_j} )^\top  \sigma_{*,*}^{\circ 2}(x)   p(x)    \\
    C_2: = & ~ - 2 ( \frac{\d A_{x}}{\d x_j} )^\top \Sigma(x)      p(x) \\
    C_3 := & ~ + 2 A_{x}^\top  \frac{\d \sigma_{*,*}^{\circ 2}(x) }{\d x_j}   p(x) \\
    C_4 := & ~ - 2 A_{x}^\top    \frac{\d \Sigma(x)}{\d x_j}    p(x) \\
    C_5 := & ~ + 2 A_{x}^\top    (\sigma_{*,*}^{\circ 2}(x))  \frac{\d p(x)}{\d x_j}\\
    C_6 := & ~ - 2 A_{x}^\top    (\Sigma(x))   \frac{\d p(x)}{\d x_j}
\end{align*}

Consider $C_1$ we have.
\begin{align*}
    C_1   = & ~ 2 A_x^\top \cdot - \diag(A_{x, *,j})   \sigma_{*,*}^{\circ 2}(x)   p(x) \\
        = & ~ - 2A_x^\top  \sigma_{*,*}^{\circ 2}(x)  \diag(A_{x, *,j}) p(x) \\
        = & ~ - 2A_x^\top  \sigma_{*,*}^{\circ 2}(x)  \diag(p(x))  A_{x, *,j} \\
        = & ~ 0.2 A_1
\end{align*}
Where the first step follows from {\bf Part 1} lemma~\ref{lem:gradient_A}, the second step and the third step follows from simple algebra, and the final step follows from definition of $A_1$ (see Definition~\ref{def:A_i}).

Consider $C_2$ we have.
\begin{align*}
    C_2    = & ~ -2 A_x^\top \cdot - \diag(A_{x, *,j})   \Sigma(x)   p(x) \\
        = & ~ 2A_x^\top  \Sigma(x) \diag(A_{x, *,j}) p(x) \\
        = & ~ 2A_x^\top  \Sigma(x) \diag(p(x)) A_{x, *,j} \\
        = & ~ -A_2
\end{align*}
Where the first step follows from {\bf Part 1} lemma~\ref{lem:gradient_A}, the second step and the third step follows from simple algebra, and the final step follows from definition of $A_2$ (see Definition~\ref{def:A_i}).

Consider $C_3$ we have.
\begin{align*}
    C_3 = &~ 8 A_x^\top \sigma^{\circ 2}_{*,*}(x) \diag (  A_{x,*,j} ) \sigma_{*,*}(x) p(x)  \\
          & ~ - 4 A_x^\top \diag( A_{x,*,j} )  \sigma_{*,*}^{\circ 2}(x) p(x) \\
          & ~ - 4 A_x^\top \sigma_{*,*}^{\circ 2}(x) \diag( A_{x,*,j} ) p(x) \\
        =  & ~ + 8 A_x^\top \sigma_{*,*}(x)  \sigma_{*,*}^{\circ 2}(x) \diag( p(x) )  A_{x,*,j} \\
        & ~  -8 A_x^\top \sigma_{*,*}^{\circ 2}(x) \diag(p(x))    A_{x, *,j}  \\ 
        = & ~A_3 + 0.8A_1
\end{align*}
Where the first step follows from {\bf Part 2} of lemma~\ref{lem:gradient_sigma_**}, the second step follows from simple algebra, and the final step follows from definition of $A_1$ and $A_3$ (see Definition~\ref{def:A_i}).

Consider $C_4$ we have.
\begin{align*}
    C_4 = &~ -4 A_x^\top \sigma_{*,*}(x) \diag (  A_{x,*,j} )\Sigma(x) p(x)  \\
          & ~ + 4 A_x^\top\diag( A_{x,*,j} )  \Sigma(x) p(x) \\
        = &~ -4 A_x^\top \sigma_{*,*}(x) \Sigma(x) \diag (  A_{x,*,j} ) p(x)  \\
          & ~ + 4 A_x^\top  \Sigma(x) \diag( A_{x,*,j} ) p(x) \\
        =  &~ - 4 A_x^\top \Sigma(x)  \sigma_{*,*}(x) \diag( p(x) )  A_{x, *,j}\\
       &~- 4  A_x^\top \Sigma(x) \diag(p(x)) A_{x, *,j}   \\
       =&~ A_4 + 2 A_2
\end{align*}
Where the first step follows from {\bf Part 2} of lemma~\ref{lem:gradient_sigma_**}, the second step follows and the third step follows from Fact~\ref{fac:vector}, and the final step follows from definition of $A_2$ and $A_4$ (see Definition~\ref{def:A_i}).

Consider $C_5$ we have.
\begin{align*}
    C_5 =&~  +2 A_{x}^\top    \sigma_{*,*}^{\circ 2}(x)  \cdot (- A_{x,*,j} \circ \sigma_{*,i}(x) +  2 \sigma_{*,*}(x)  (\sigma_{*,i}(x) \circ A_{x,*,j}) - \sigma_{*,i}(x) A_{x,i,j} ) \\
    = & ~ - 2 A_{x}^\top    \sigma_{*,*}^{\circ 2}(x)  A_{x,*,j} \circ \sigma_{*,i}(x)  \\
    & ~ + 4 A_{x}^\top \sigma_{*,*}^{\circ 2}(x)  \sigma_{*,*}(x)  (\sigma_{*,i}(x) \circ A_{x,*,j})\\
    & ~ -2 A_{x}^\top    \sigma_{*,*}^{\circ 2}(x) \sigma_{*,i}(x) A_{x,i,j}^\top\\
    = &~ - 2 A_x^\top \sigma_{*,*}^{\circ 2}(x) \diag(\sigma_{*,i}(x) )A_{x, *,j} \\
    &~ +4A_x^\top  \sigma_{*,*}^{\circ 2}(x)  \sigma_{*,*}(x)    \diag(\sigma_{*,i}(x))A_{x, *,j}\\
    &~ -2 A_{x}^\top \sigma_{*,*}^{\circ 2}(x) \sigma_{*,i}(x) A_{x,i,j}^\top \\
    = &~ A_5 +A_7+ A_9
\end{align*}
Where the first step follows from of lemma~\ref{lem:p(x)}, the second step follows from simple algebra, the third step follows from Fact~\ref{fac:vector}, and the final step follows from definition of $A_5$ and $A_7$and $A_9$ (see Definition~\ref{def:A_i}).

Consider $C_6$ we have.
\begin{align*}
    C_6 =&~   -2 A_{x}^\top    (\Sigma(x))    \cdot (- A_{x,*,j} \circ \sigma_{*,i}(x) +  2 \sigma_{*,*}(x)  (\sigma_{*,i}(x) \circ A_{x,*,j}) - \sigma_{*,i}(x) A_{x,i,j} ) \\
    =&~ +  2 A_{x}^\top    (\Sigma(x))  A_{x,*,j} \circ \sigma_{*,i}(x) \\
     &~ -4 A_{x}^\top    (\Sigma(x)) \sigma_{*,*}(x)  (\sigma_{*,i}(x) \circ A_{x,*,j})\\
     &~ +2 A_{x}^\top    (\Sigma(x)) \sigma_{*,i}(x) A_{x,i,j}\\
    = &~ + 2 A_x^\top  \Sigma(x) \diag(\sigma_{*,i}(x)) A_{x, *,j}\\
    &~ - 4 A_x^\top  \Sigma(x)  \sigma_{*,*}(x)    \diag(\sigma_{*,i}(x))A_{x, *,j}\\
    &~ +2 A_{x}^\top \Sigma(x) \sigma_{*,i}(x)  A_{x,i,j}^\top \\
    = &~ A_6 +A_8+ A_{10}
\end{align*}
Where the first step follows from of lemma~\ref{lem:p(x)}, the second step follows from simple algebra, the third step follows from Fact~\ref{fac:vector}, and the final step follows from the definition of $A_6$, $A_8$, and $A_{10}$ (see Definition~\ref{def:A_i}).

Therefore, we have 
\begin{align*}
    \frac{\d g(x)}{\d x_j} 
    = & ~ C_1+ C_2 + C_3 + C_4 + C_5 + C_6\\
    =& ~ A_1 +A_2 +A_3 +A_4+ A_5 +A_6+A_7 +A_8+A_9+A_{10}
\end{align*}
where the final step follows from definition of $A_i$ (see Definition~\ref{def:A_i}).

Thus, we have
\begin{align*}
    \frac{\d g(x)}{\d x} = \sum_{i = 1}^{10} \wt{A}_i.
\end{align*}
\end{proof}

\subsection{Gradient for $L_c(x)$}\label{sub:l_c}

\begin{definition}
For each $l \in [10]$, we define
\begin{align*}
    B_l := ( 2 (g(x)_{j_0} - c_{j_0}) A_l)_{j_0}.
\end{align*}
 \end{definition}

\begin{lemma}[Formal version of Lemma~\ref{lem:g(x)_informal}]\label{lem:L_c}
If the following conditions hold
    \begin{itemize}
        \item Let $b \in \R^n$  
        be defined as in Definition~\ref{def:loss_lb}.
        \item Let $c \in \R^d$ be defined as in Definition~\ref{def:L_c}.
        \item Let $A_x \in \R^{n \times d} $ be defined as in Definition~\ref{def:A_x}.
        \item Let $\sigma_{*,*}^{\circ 2}(x) \in \R^{n \times n}$, $\sigma_{*,*}(x)  \in \R^{n \times n}$, and $\sigma_{*, i}(x)  \in \R^n $ be defined as in Definition~\ref{def:sigma}.
        \item Let $\Sigma(x)  \in \R^{n \times n} $ be defined as in Definition~\ref{def:Sigma}.
        \item Let $A_{x, *, j}  \in \R^n$ denote the $j$-th column of $A_x$.
        \item Let $g(x) \in \R^d $ be defined as in Definition~\ref{def:g(x)}
        \item Let $p(x) = \sigma_{*, i}(x) - b \in \R^n$.
        \item Let $q(x) = g(x) - c \in \R^d$.
        \item Let $L_{c,j_0}:= (g(x)_{j_0} - c_{j_0})^2$ for $j_0 \in [d]$
    \end{itemize}

 Then, we have
    \begin{itemize}
        \item Part 1. For all $j_0, j \in [d]$,
        \begin{align*}
            \frac{\d L_{c,j_0} (x)}{\d x_{j} } = \sum_{l=1}^{10} B_l
        \end{align*}
        \item Part 2. $\frac{\d L_{c} (x)}{\d x } = (\sum_{i = 1}^{10} \wt{A}_i)^\top q(x)$
    \end{itemize}
\end{lemma}

\begin{proof}

We have
\begin{align*}
    \frac{\d L_{c,j_0} (x)}{\d x_{j} } 
    = & ~ \frac{\d (g(x)_{j_0} - c_{j_0})^2}{\d x_{j} } \\
    = & ~ 2 (g(x)_{j_0} - c_{j_0}) \frac{\d (g(x)_{j_0} - c_{j_0})}{\d x_{j} } \\
    = & ~ 2 (g(x)_{j_0} - c_{j_0}) \frac{ \d g(x)_{j_0} }{\d x_{j} } \\
    = & ~ 2 (g(x)_{j_0} - c_{j_0}) (\frac{ \d g(x) }{\d x_{j} })_{j_0} \\
    = & ~ 2 (g(x)_{j_0} - c_{j_0}) (\sum_{l=1}^{10} A_l)_{j_0},
\end{align*}
where the first step follows from the definition of $L_{c,j_0}$(see the Lemma statement), the second step follows from the chain rule, the third step follows from $\frac{ \d c_{j_0} }{\d x_{j} } = 0$, the fourth step follows from the simple definition from matrix calculus, and the last step follows from Lemma~\ref{lem:g(x)}.

For convenience, for each $l \in [10]$, we define
\begin{align*}
    B_l = ( 2 (g(x)_{j_0} - c_{j_0}) A_l)_{j_0}.
\end{align*}

Therefore, we have
\begin{align*}
    \frac{\d L_{c,j_0} (x)}{\d x_{j} }  = \sum_{l=1}^{10} B_l.
\end{align*}

Also, we have
\begin{align*}
    \frac{\d L_c(x)}{\d x} 
    = & ~ \frac{\d 0.5 \cdot \| g(x) - c\|_2^2}{\d x} \\
    = & ~ \frac{\d 0.5  (g(x) - c)^\top (g(x) - c)}{\d x} \\
    = & ~ \frac{\d (g(x) - c)^\top }{\d x} (g(x) - c)\\
    = & ~ (\frac{\d g(x) }{\d x})^\top (g(x) - c)\\
    = & ~ (\sum_{i = 1}^{10} \wt{A}_i)^\top q(x),
\end{align*}
where the first step follows from the Definition of $L_c(x)$ (see Definition~\ref{def:L_c}), the second step follows from $\|x\|_2^2 = x^\top x$, the third step follows from the product rule, the fourth step follows from $\frac{\d c}{\d x} = 0$, and the last step follows from {\bf Part 2} of Lemma~\ref{lem:g(x)}.
\end{proof}

\section{Hessian}\label{sec:hessian}

In Section~\ref{sub:gradient_Al}, we present the gradient for $A_l$. In Section~\ref{sub::L_c_Hessian}, we present the Hessian for $L_c(x)$.

\subsection{Gradient for $A_l$}\label{sub:gradient_Al}

\begin{lemma}[Formal version of Lemma~\ref{lem:gradient_Al_informal}]\label{lem:gradient_Al}
    For each $l \in [10]$, let $A_l$ be defined as in Definition~\ref{def:A_i}. Let $x_{j_2} \in \R$. Then, the gradient of $A_l$ can be expressed as:
    \begin{itemize}
        \item $\frac{\d A_{1}}{\d x_{j_2}} = \sum_{h = 1}^8 A_{1, h} := \mathsf{A}_1$,
        \item $\frac{\d A_{2}}{\d x_{j_2}} = \sum_{h = 1}^7 A_{2, h} := \mathsf{A}_2$,
        \item $\frac{\d A_{3}}{\d x_{j_2}} = \sum_{h = 1}^{11} A_{3, h} := \mathsf{A}_3$,
        \item $\frac{\d A_{4}}{\d x_{j_2}} = \sum_{h = 1}^{10} A_{4, h} := \mathsf{A}_4$,
        \item $\frac{\d A_{5}}{\d x_{j_2}} = \sum_{h = 1}^8 A_{5, h} := \mathsf{A}_5$,
        \item $\frac{\d A_{6}}{\d x_{j_2}} = \sum_{h = 1}^8 A_{5, h} := \mathsf{A}_6$,
        \item $\frac{\d A_{7}}{\d x_{j_2}} = \sum_{h = 1}^{11} A_{7, h} := \mathsf{A}_7$,
        \item $\frac{\d A_{8}}{\d x_{j_2}} = \sum_{h = 1}^{10} A_{8, h} := \mathsf{A}_8$,
        \item $\frac{\d A_{9}}{\d x_{j_2}} = \sum_{h = 1}^{8} A_{9, h} := \mathsf{A}_9$, and
        \item $\frac{\d A_{10}}{\d x_{j_2}} = \sum_{h = 1}^{7} A_{10, h} := \mathsf{A}_{10}$.
    \end{itemize}
\end{lemma}
\begin{proof}
    For the gradient of $A_1$, we have
    \begin{align*}
        & ~ \frac{\d A_{1}}{\d x_{j_2}} \\
        = & ~ \frac{\d -10 A_x^\top \sigma_{*,*}^{\circ 2}(x) \diag(p(x)) A_{x, *,j}}{\d x_{j_2}} \\
        = & ~ -10 \frac{\d A_x^\top}{\d x_{j_2}} \sigma_{*,*}^{\circ 2}(x) \diag(p(x)) A_{x, *,j} \\
        & ~ - 10 A_x^\top \frac{\d \sigma_{*,*}^{\circ 2}(x)}{\d x_{j_2}} \diag(p(x)) A_{x, *,j} \\
        & ~ - 10 A_x^\top \sigma_{*,*}^{\circ 2}(x) \frac{\d \diag(p(x))}{\d x_{j_2}} A_{x, *,j} \\
        & ~ - 10 A_x^\top \sigma_{*,*}^{\circ 2}(x) \diag(p(x)) \frac{\d A_{x, *,j}}{\d x_{j_2}} \\
        = & ~ +10 A_x^\top \diag(A_{x, *, j_2}) \sigma_{*,*}^{\circ 2}(x) \diag(p(x)) A_{x, *,j} \\
        & ~ - 10 A_x^\top (4 \sigma^2_{*,*}(x) \diag (  A_{x,*,j_2} ) \sigma_{*,*}(x) - 2 \diag( A_{x,*,j_2} )  \sigma_{*,*}^{\circ 2}(x) - 2\sigma^{\circ 2}_{*,*}(x) \diag( A_{x,*,j_2} )) \diag(p(x)) A_{x, *,j} \\
        & ~ - 10 A_x^\top \sigma_{*,*}^{\circ 2}(x) (\diag(-  A_{x,*,j_2} \circ \sigma_{*,i}(x) +  2 \sigma_{*,*}(x)  (\sigma_{*,i}(x) \circ A_{x,*,j_2}) - \sigma_{*,i}(x) A_{x,i,j_2})) A_{x, *,j} \\
        & ~ - 10 A_x^\top \sigma_{*,*}^{\circ 2}(x) \diag(p(x)) (- A_{x, *,j_2} \circ A_{x, *,j}),
    \end{align*}
    where the first step follows from the definition of $A_1$ (see Definition~\ref{def:A_i}), the second step follows from the product rule, the third step follows from combining Lemma~\ref{lem:gradient_A}, Lemma~\ref{lem:gradient_sigma_**}, Lemma~\ref{lem:p(x)}, and Lemma~\ref{lem:gradient_Axij_and_Axj}.

    For simplicity, we denote $A_{1, h}$ to be the $h$-th term of $\frac{\d A_{1}}{\d x_{j_2}}$.

    Therefore, we have
    \begin{align*}
        \frac{\d A_{1}}{\d x_{j_2}} = \sum_{h = 1}^8 A_{1, h},
    \end{align*}
    where
    \begin{itemize}
        \item $A_{1, 1} = 10 A_x^\top \diag(A_{x, *, j_2}) \sigma_{*,*}^{\circ 2}(x) \diag(p(x)) A_{x, *,j}$,
        \item $A_{1, 2} = - 40 A_x^\top \sigma^2_{*,*}(x) \diag (  A_{x,*,j_2} ) \sigma_{*,*}(x) A_{x, *,j}$,
        \item $A_{1, 3} = 20 A_x^\top \diag( A_{x,*,j_2} )  \sigma_{*,*}^{\circ 2}(x)  A_{x, *,j}$,
        \item $A_{1, 4} = 20 A_x^\top \sigma^{\circ 2}_{*,*}(x) \diag( A_{x,*,j_2} ) \diag(p(x)) A_{x, *,j}$,
        \item $A_{1, 5} = 10 A_x^\top \sigma_{*,*}^{\circ 2}(x) \diag(  A_{x,*,j_2} \circ \sigma_{*,i}(x)) A_{x, *,j}$,
        \item $A_{1, 6} = - 20 A_x^\top \sigma_{*,*}^{\circ 2}(x) \diag(\sigma_{*,*}(x)  (\sigma_{*,i}(x) \circ A_{x,*,j_2})) A_{x, *,j}$,
        \item $A_{1, 7} = 10 A_x^\top \sigma_{*,*}^{\circ 2}(x) \diag(\sigma_{*,i}(x) A_{x,i,j_2}) A_{x, *,j}$, and
        \item $A_{1, 8} = 10 A_x^\top \sigma_{*,*}^{\circ 2}(x) \diag(p(x)) A_{x, *,j_2} \circ A_{x, *,j}$.
    \end{itemize}

    For the gradient of $A_2$, we have
    \begin{align*}
        & ~ \frac{\d A_{2}}{\d x_{j_2}} \\
        = & ~ \frac{\d -2  A_x^\top \Sigma(x) \diag(p(x)) A_{x, *,j}}{\d x_{j_2}} \\
        = & ~ -2 \frac{\d A_x^\top}{\d x_{j_2}} \Sigma(x) \diag(p(x)) A_{x, *,j} \\
        & ~ - 2 A_x^\top \frac{\d \Sigma(x)}{\d x_{j_2}} \diag(p(x)) A_{x, *,j} \\
        & ~ - 2 A_x^\top \Sigma(x) \frac{\d \diag(p(x))}{\d x_{j_2}} A_{x, *,j} \\
        & ~ - 2 A_x^\top \Sigma(x) \diag(p(x)) \frac{\d A_{x, *,j}}{\d x_{j_2}} \\
        = & ~ +2 A_x^\top \diag(A_{x, *, j_2}) \Sigma(x) \diag(p(x)) A_{x, *,j} \\
        & ~ - 4 A_x^\top (\sigma_{*,*}(x) \diag (  A_{x,*,j} )\Sigma(x) - \diag( A_{x,*,j} )  \Sigma(x)) \diag(p(x)) A_{x, *,j} \\
        & ~ - 2 A_x^\top \Sigma(x) (\diag(-  A_{x,*,j_2} \circ \sigma_{*,i}(x) +  2 \sigma_{*,*}(x)  (\sigma_{*,i}(x) \circ A_{x,*,j_2}) - \sigma_{*,i}(x) A_{x,i,j_2})) A_{x, *,j} \\
        & ~ - 2 A_x^\top \Sigma(x) \diag(p(x)) (- A_{x, *,j_2} \circ A_{x, *,j}),
    \end{align*}
    where the first step follows from the definition of $A_2$ (see Definition~\ref{def:A_i}), the second step follows from the product rule, the third step follows from combining Lemma~\ref{lem:gradient_A}, Lemma~\ref{lem:Sigma}, Lemma~\ref{lem:p(x)}, and Lemma~\ref{lem:gradient_Axij_and_Axj}.

    For simplicity, we denote $A_{2, h}$ to be the $h$-th term of $\frac{\d A_{2}}{\d x_{j_2}}$.

    Therefore, we have
    \begin{align*}
        \frac{\d A_{2}}{\d x_{j_2}} = \sum_{h = 1}^7 A_{2, h},
    \end{align*}
    where 
    \begin{itemize}
        \item $A_{2, 1} = 2 A_x^\top \diag(A_{x, *, j_2}) \Sigma(x) \diag(p(x)) A_{x, *,j}$,
        \item $A_{2, 2} = - 4 A_x^\top \sigma_{*,*}(x) \diag (  A_{x,*,j} )\Sigma(x) \diag(p(x)) A_{x, *,j}$, 
        \item $A_{2, 3} = 4 A_x^\top \diag( A_{x,*,j} )  \Sigma(x) \diag(p(x)) A_{x, *,j}$, 
        \item $A_{2, 4} = 2 A_x^\top \Sigma(x) \diag(  A_{x,*,j_2} \circ \sigma_{*,i}(x)) A_{x, *,j}$,
        \item $A_{2, 5} = - 4 A_x^\top \Sigma(x) \diag(\sigma_{*,*}(x)  (\sigma_{*,i}(x) \circ A_{x,*,j_2})) A_{x, *,j}$,
        \item $A_{2, 6} = 2 A_x^\top \Sigma(x) \diag( \sigma_{*,i}(x) A_{x,i,j_2})) A_{x, *,j}$, and
        \item $A_{2, 7} = 2 A_x^\top \Sigma(x) \diag(p(x)) A_{x, *,j_2} \circ A_{x, *,j}$.
    \end{itemize}

    For the gradient of $A_3$, we have
    \begin{align*}
        & ~ \frac{\d A_{3}}{\d x_{j_2}} \\
        = & ~ 8\frac{\d A_x^\top \sigma_{*,*}(x)  \sigma_{*,*}^{\circ 2}(x) \diag( p(x) )  A_{x,*,j}}{\d x_{j_2}} \\
        = & ~ + 8\frac{\d A_x^\top }{\d x_{j_2}} \sigma_{*,*}(x)  \sigma_{*,*}^{\circ 2}(x) \diag( p(x) )  A_{x,*,j} \\
        & ~ + 8 A_x^\top \frac{\d \sigma_{*,*}(x)}{\d x_{j_2}}   \sigma_{*,*}^{\circ 2}(x) \diag( p(x) )  A_{x,*,j}\\
        & ~ + 8 A_x^\top \sigma_{*,*}(x)  \frac{\d \sigma_{*,*}^{\circ 2}(x)}{\d x_{j_2}} \diag( p(x) )  A_{x,*,j} \\
        & ~ + 8 A_x^\top \sigma_{*,*}(x)  \sigma_{*,*}^{\circ 2}(x) \frac{\d \diag( p(x) )}{\d x_{j_2}}   A_{x,*,j}\\
        & ~ + 8 A_x^\top \sigma_{*,*}(x)  \sigma_{*,*}^{\circ 2}(x) \diag( p(x) )  \frac{\d A_{x,*,j}}{\d x_{j_2}} \\
        = & ~ - 8 A_x^\top \diag(A_{x, *, j_2}) \sigma_{*,*}(x)  \sigma_{*,*}^{\circ 2}(x) \diag( p(x) )  A_{x,*,j} \\
        & ~ + 8 A_x^\top (2 \sigma_{*,*}(x) \diag (  A_{x,*,j_2} ) \sigma_{*,*}(x) - \diag( A_{x,*,j_2} )  \sigma_{*,*}(x) - \sigma_{*,*}(x) \diag( A_{x,*,j_2} ))   \sigma_{*,*}^{\circ 2}(x) \diag( p(x) )  A_{x,*,j}\\
        & ~ + 8 A_x^\top \sigma_{*,*}(x)  (4 \sigma^2_{*,*}(x) \diag (  A_{x,*,j} ) \sigma_{*,*}(x) - 2 \diag( A_{x,*,j} )  \sigma_{*,*}^{\circ 2}(x) - 2\sigma^{\circ 2}_{*,*}(x) \diag( A_{x,*,j} )) \diag( p(x) )  A_{x,*,j} \\
        & ~ + 8 A_x^\top \sigma_{*,*}(x)  \sigma_{*,*}^{\circ 2}(x) ((\diag(-  A_{x,*,j_2} \circ \sigma_{*,i}(x) +  2 \sigma_{*,*}(x)  (\sigma_{*,i}(x) \circ A_{x,*,j_2}) - \sigma_{*,i}(x) A_{x,i,j_2})))   A_{x,*,j}\\
        & ~ + 8 A_x^\top \sigma_{*,*}(x)  \sigma_{*,*}^{\circ 2}(x) \diag( p(x) )  (- A_{x, *,j_2} \circ A_{x, *,j}),
    \end{align*}
    where the first step follows from the definition of $A_3$ (see Definition~\ref{def:A_i}), the second step follows from the product rule, the third step follows from combining Lemma~\ref{lem:gradient_A}, Lemma~\ref{lem:gradient_sigma_**}, Lemma~\ref{lem:p(x)}, and Lemma~\ref{lem:gradient_Axij_and_Axj}.

    For simplicity, we denote $A_{3, h}$ to be the $h$-th term of $\frac{\d A_{3}}{\d x_{j_2}}$.

    Therefore, we have
    \begin{align*}
        \frac{\d A_{3}}{\d x_{j_2}} = \sum_{h = 1}^{11} A_{3, h},
    \end{align*}
    where
    \begin{itemize}
        \item $A_{3, 1} = - 8 A_x^\top \diag(A_{x, *, j_2}) \sigma_{*,*}(x)  \sigma_{*,*}^{\circ 2}(x) \diag( p(x) )  A_{x,*,j}$
        \item $A_{3, 2} = 16 A_x^\top \sigma_{*,*}(x) \diag (  A_{x,*,j_2} ) \sigma_{*,*}(x) \sigma_{*,*}^{\circ 2}(x) \diag( p(x) )  A_{x,*,j}$
        \item $A_{3, 3} = -8 A_x^\top \diag( A_{x,*,j_2} )  \sigma_{*,*}(x) \sigma_{*,*}^{\circ 2}(x) \diag( p(x) )  A_{x,*,j}$
        \item $A_{3, 4} = -8 A_x^\top \sigma_{*,*}(x) \diag( A_{x,*,j_2} ) \sigma_{*,*}^{\circ 2}(x) \diag( p(x) )  A_{x,*,j}$
        \item $A_{3, 5} = 32 A_x^\top \sigma_{*,*}(x) \sigma^2_{*,*}(x) \diag (  A_{x,*,j} ) \sigma_{*,*}(x) \diag( p(x) )  A_{x,*,j}$
        \item $A_{3, 6} = -16 A_x^\top \sigma_{*,*}(x) \diag( A_{x,*,j} )  \sigma_{*,*}^{\circ 2}(x) \diag( p(x) )  A_{x,*,j}$
        \item $A_{3, 7} = -16 A_x^\top \sigma_{*,*}(x)  \sigma^{\circ 2}_{*,*}(x) \diag( A_{x,*,j} ) \diag( p(x) )  A_{x,*,j}$
        \item $A_{3, 8} = -8 A_x^\top \sigma_{*,*}(x)  \sigma_{*,*}^{\circ 2}(x) \diag(  A_{x,*,j_2} \circ \sigma_{*,i}(x) )   A_{x,*,j}$
        \item $A_{3, 9} = 16 A_x^\top \sigma_{*,*}(x)  \sigma_{*,*}^{\circ 2}(x) \diag(\sigma_{*,*}(x)  (\sigma_{*,i}(x) \circ A_{x,*,j_2}) )   A_{x,*,j}$
        \item $A_{3, 10} = -8 A_x^\top \sigma_{*,*}(x)  \sigma_{*,*}^{\circ 2}(x) \diag(\sigma_{*,i}(x) A_{x,i,j_2})   A_{x,*,j}$
        \item $A_{3, 11} = -8 A_x^\top \sigma_{*,*}(x)  \sigma_{*,*}^{\circ 2}(x) \diag( p(x) )  A_{x, *,j_2} \circ A_{x, *,j}$
    \end{itemize}

    For the gradient of $A_4$, we have
    \begin{align*}
        \frac{\d A_{4}}{\d x_{j_2}}
        = & ~ \frac{\d - 4 A_x^\top \Sigma(x)  \sigma_{*,*}(x) \diag( p(x) )  A_{x, *,j}}{\d x_{j_2}}\\
        = & ~ - 4 \frac{\d A_x^\top }{\d x_{j_2}} \Sigma(x)  \sigma_{*,*}(x) \diag( p(x) )  A_{x, *,j}\\
        & ~ - 4 A_x^\top \frac{\d \Sigma(x)}{\d x_{j_2}}  \sigma_{*,*}(x) \diag( p(x) )  A_{x, *,j}\\
        & ~ - 4 A_x^\top \Sigma(x)  \frac{\d \sigma_{*,*}(x)}{\d x_{j_2}} \diag( p(x) )  A_{x, *,j}\\
        & ~ - 4 A_x^\top \Sigma(x)  \sigma_{*,*}(x) \frac{\d \diag( p(x) )}{\d x_{j_2}}  A_{x, *,j}\\
        & ~ - 4 A_x^\top \Sigma(x)  \sigma_{*,*}(x) \diag( p(x) )  \frac{\d A_{x, *,j}}{\d x_{j_2}}\\
        = & ~ + 4 A_x^\top \diag(A_{x, *, j_2}) \Sigma(x)  \sigma_{*,*}(x) \diag( p(x) )  A_{x, *,j}\\
        & ~ - 8 A_x^\top (\sigma_{*,*}(x) \diag (  A_{x,*,j} )\Sigma(x) - \diag( A_{x,*,j} )  \Sigma(x))  \sigma_{*,*}(x) \diag( p(x) )  A_{x, *,j}\\
        & ~ - 4 A_x^\top \Sigma(x)  (2 \sigma_{*,*}(x) \diag (  A_{x,*,j_2} ) \sigma_{*,*}(x) - \diag( A_{x,*,j_2} )  \sigma_{*,*}(x) - \sigma_{*,*}(x) \diag( A_{x,*,j_2} )) \diag( p(x) )  A_{x, *,j}\\
        & ~ - 4 A_x^\top \Sigma(x)  \sigma_{*,*}(x) (\diag(-  A_{x,*,j_2} \circ \sigma_{*,i}(x) +  2 \sigma_{*,*}(x)  (\sigma_{*,i}(x) \circ A_{x,*,j_2}) - \sigma_{*,i}(x) A_{x,i,j_2}))  A_{x, *,j}\\
        & ~ - 4 A_x^\top \Sigma(x)  \sigma_{*,*}(x) \diag( p(x) )  (- A_{x, *,j_2} \circ A_{x, *,j}),
    \end{align*}
    where the first step follows from the definition of $A_4$ (see Definition~\ref{def:A_i}), the second step follows from the product rule, the third step follows from combining Lemma~\ref{lem:gradient_A}, Lemma~\ref{lem:Sigma}, Lemma~\ref{lem:gradient_sigma_**}, Lemma~\ref{lem:p(x)}, and Lemma~\ref{lem:gradient_Axij_and_Axj}.

    For simplicity, we denote $A_{4, h}$ to be the $h$-th term of $\frac{\d A_{4}}{\d x_{j_2}}$.

    Therefore, we have
    \begin{align*}
        \frac{\d A_{4}}{\d x_{j_2}} = \sum_{h = 1}^{10} A_{4, h},
    \end{align*}
    where
    \begin{itemize}
        \item $A_{4, 1} = + 4 A_x^\top \diag(A_{x, *, j_2}) \Sigma(x)  \sigma_{*,*}(x) \diag( p(x) )  A_{x, *,j}$
        \item $A_{4, 2} = - 8 A_x^\top \sigma_{*,*}(x) \diag (  A_{x,*,j} )\Sigma(x)  \sigma_{*,*}(x) \diag( p(x) )  A_{x, *,j}$
        \item $A_{4, 3} = + 8 A_x^\top \diag( A_{x,*,j} )  \Sigma(x)  \sigma_{*,*}(x) \diag( p(x) )  A_{x, *,j}$
        \item $A_{4, 4} = - 8 A_x^\top \Sigma(x) \sigma_{*,*}(x) \diag (  A_{x,*,j_2} ) \sigma_{*,*}(x) \diag( p(x) )  A_{x, *,j}$
        \item $A_{4, 5} =  4 A_x^\top \Sigma(x) \diag( A_{x,*,j_2} )  \sigma_{*,*}(x) \diag( p(x) )  A_{x, *,j}$
        \item $A_{4, 6} = 4 A_x^\top \Sigma(x)  \sigma_{*,*}(x) \diag( A_{x,*,j_2} ) \diag( p(x) )  A_{x, *,j}$
        \item $A_{4, 7} = 4 A_x^\top \Sigma(x)  \sigma_{*,*}(x) \diag(A_{x,*,j_2} \circ \sigma_{*,i}(x))  A_{x, *,j}$
        \item $A_{4, 8} = - 8 A_x^\top \Sigma(x)  \sigma_{*,*}(x) \diag( \sigma_{*,*}(x)  (\sigma_{*,i}(x) \circ A_{x,*,j_2}) )  A_{x, *,j}$
        \item $A_{4, 9} = 4 A_x^\top \Sigma(x)  \sigma_{*,*}(x) \diag(\sigma_{*,i}(x) A_{x,i,j_2})  A_{x, *,j}$
        \item $A_{4, 10} = 4 A_x^\top \Sigma(x)  \sigma_{*,*}(x) \diag( p(x) )  A_{x, *,j_2} \circ A_{x, *,j}$
    \end{itemize}

     For the gradient of $A_5$, we have
    \begin{align*}
        & ~ \frac{\d A_{5}}{\d x_{j_2}} \\
        = & ~ \frac{\d - 2 A_x^\top \sigma_{*,*}^{\circ 2}(x) \diag(\sigma_{*,i}(x) )A_{x, *,j}}{\d x_{j_2}} \\
        = & ~ - 2 \frac{\d A_x^\top}{\d x_{j_2}} \sigma_{*,*}^{\circ 2}(x) \diag(\sigma_{*,i}(x)) A_{x, *,j} \\
        & ~ - 2 A_x^\top \frac{\d \sigma_{*,*}^{\circ 2}(x)}{\d x_{j_2}} \diag(\sigma_{*,i}(x)) A_{x, *,j} \\
        & ~ - 2 A_x^\top \sigma_{*,*}^{\circ 2}(x) \frac{\d \diag(\sigma_{*,i}(x))}{\d x_{j_2}} A_{x, *,j} \\
        & ~ - 2 A_x^\top \sigma_{*,*}^{\circ 2}(x) \diag(\sigma_{*,i}(x)) \frac{\d A_{x, *,j}}{\d x_{j_2}} \\
        = & ~ + 2 A_x^\top \diag(A_{x, *, j_2}) \sigma_{*,*}^{\circ 2}(x) \diag(\sigma_{*,i}(x)) A_{x, *,j} \\
        & ~ - 2 A_x^\top (4 \sigma^2_{*,*}(x) \diag (  A_{x,*,j_2} ) \sigma_{*,*}(x) - 2 \diag( A_{x,*,j_2} )  \sigma_{*,*}^{\circ 2}(x) - 2\sigma^{\circ 2}_{*,*}(x) \diag( A_{x,*,j_2} )) \diag(\sigma_{*,i}(x)) A_{x, *,j} \\
        & ~ - 2 A_x^\top \sigma_{*,*}^{\circ 2}(x) (\diag(-  A_{x,*,j_2} \circ \sigma_{*,i}(x) +  2 \sigma_{*,*}(x)  (\sigma_{*,i}(x) \circ A_{x,*,j_2}) - \sigma_{*,i}(x) A_{x,i,j_2})) A_{x, *,j} \\
        & ~ - 2 A_x^\top \sigma_{*,*}^{\circ 2}(x) \diag(\sigma_{*,i}(x) ) (- A_{x, *,j_2} \circ A_{x, *,j}),
    \end{align*}
    where the first step follows from the definition of $A_5$ (see Definition~\ref{def:A_i}), the second step follows from the product rule, the third step follows from combining Lemma~\ref{lem:gradient_A}, Lemma~\ref{lem:gradient_sigma_**}, Lemma~\ref{lem:gradient_sigma_*i}, and Lemma~\ref{lem:gradient_Axij_and_Axj}.

    For simplicity, we denote $A_{5, h}$ to be the $h$-th term of $\frac{\d A_{5}}{\d x_{j_2}}$.

    Therefore, we have
    \begin{align*}
        \frac{\d A_{5}}{\d x_{j_2}} = \sum_{h = 1}^8 A_{5, h},
    \end{align*}
    where

    \begin{itemize}
        \item $A_{5, 1} =  + 2 A_x^\top \diag(A_{x, *, j_2}) \sigma_{*,*}^{\circ 2}(x) \diag(\sigma_{*,i}(x)) A_{x, *,j} $,
        \item $A_{5, 2} = - 8 A_x^\top  \sigma^2_{*,*}(x) \diag (  A_{x,*,j_2} ) \sigma_{*,*}(x)  \diag(\sigma_{*,i}(x)) A_{x, *,j} $,
        \item $A_{5, 3} = + 4 A_x^\top   \diag( A_{x,*,j_2} )  \sigma_{*,*}^{\circ 2}(x)  \diag(\sigma_{*,i}(x)) A_{x, *,j} $,
        \item $A_{5, 4} = + 4 A_x^\top  \sigma^{\circ 2}_{*,*}(x) \diag( A_{x,*,j_2} )) \diag(\sigma_{*,i}(x) A_{x, *,j} $,
        \item $A_{5, 5} = + 2 A_x^\top \sigma_{*,*}^{\circ 2}(x) \diag(  A_{x,*,j_2} \circ \sigma_{*,i}(x) ) A_{x, *,j} $,
        \item $A_{5, 6} =  - 4 A_x^\top \sigma_{*,*}^{\circ 2}(x) \diag( \sigma_{*,*}(x)  (\sigma_{*,i}(x) \circ A_{x,*,j_2})  A_{x, *,j} $,
        \item $A_{5, 7} = + 2 A_x^\top \sigma_{*,*}^{\circ 2}(x) \diag(  \sigma_{*,i}(x) A_{x,i,j_2}) A_{x, *,j} $, and
        \item $A_{5, 8} = + 2 A_x^\top \sigma_{*,*}^{\circ 2}(x) \diag(\sigma_{*,i}(x) )  A_{x, *,j_2} \circ A_{x, *,j} $.
    \end{itemize}

     For the gradient of $A_6$, we have
    \begin{align*}
        & ~ \frac{\d A_{6}}{\d x_{j_2}} \\
        = & ~ \frac{\d + 2 A_x^\top  \Sigma(x) \diag(\sigma_{*,i}(x)) A_{x, *,j}}{\d x_{j_2}} \\
        = & ~ + 2 \frac{\d A_x^\top}{\d x_{j_2}} \Sigma(x) \diag(\sigma_{*,i}(x)) A_{x, *,j} \\
        & ~ + 2 A_x^\top \frac{\d \Sigma(x)}{\d x_{j_2}} \diag(\sigma_{*,i}(x)) A_{x, *,j} \\
        & ~ + 2 A_x^\top \Sigma(x) \frac{\d \diag(\sigma_{*,i}(x))}{\d x_{j_2}} A_{x, *,j} \\
        & ~ + 2 A_x^\top \Sigma(x) \diag(\sigma_{*,i}(x)) \frac{\d A_{x, *,j}}{\d x_{j_2}} \\
        = & ~ - 2 A_x^\top \diag(A_{x, *, j_2}) \Sigma(x) \diag(\sigma_{*,i}(x)) A_{x, *,j} \\
        & ~ - 2 A_x^\top (2 \sigma_{*,*}(x) \diag (  A_{x,*,j} )\Sigma(x) - 2 \diag( A_{x,*,j} )  \Sigma(x)) \diag(\sigma_{*,i}(x)) A_{x, *,j} \\
        & ~ - 2 A_x^\top \Sigma(x) (\diag(-  A_{x,*,j_2} \circ \sigma_{*,i}(x) +  2 \sigma_{*,*}(x)  (\sigma_{*,i}(x) \circ A_{x,*,j_2}) - \sigma_{*,i}(x) A_{x,i,j_2})) A_{x, *,j} \\
        & ~ - 2 A_x^\top \Sigma(x) \diag(\sigma_{*,i}(x) ) (- A_{x, *,j_2} \circ A_{x, *,j}),
    \end{align*}
    where the first step follows from the definition of $A_6$ (see Definition~\ref{def:A_i}), the second step follows from the product rule, the third step follows from combining Lemma~\ref{lem:gradient_A}, Lemma~\ref{lem:Sigma}, Lemma~\ref{lem:gradient_sigma_*i}, and Lemma~\ref{lem:gradient_Axij_and_Axj}.

    For simplicity, we denote $A_{6, h}$ to be the $h$-th term of $\frac{\d A_{6}}{\d x_{j_2}}$.

    Therefore, we have
    \begin{align*}
        \frac{\d A_{6}}{\d x_{j_2}} = \sum_{h = 1}^7 A_{6, h},
    \end{align*}
    where
    \begin{itemize}
        \item $A_{6, 1} =  - 2 A_x^\top \diag(A_{x, *, j_2}) \Sigma(x) \diag(\sigma_{*,i}(x)) A_{x, *,j}$,
        \item $A_{6, 2} = - 4 A_x^\top \sigma_{*,*}(x) \diag (  A_{x,*,j} )\Sigma(x) \diag(\sigma_{*,i}(x)) A_{x, *,j} $,
        \item $A_{6, 3} = + 4 A_x^\top \diag( A_{x,*,j} )  \Sigma(x) \diag(\sigma_{*,i}(x)) A_{x, *,j} $,
        \item $A_{6, 4} = + 2 A_x^\top \Sigma(x) \diag(  A_{x,*,j_2} \circ \sigma_{*,i}(x)) A_{x, *,j}$,
        \item $A_{6, 5} = - 4 A_x^\top \Sigma(x) \diag(\sigma_{*,*}(x)  (\sigma_{*,i}(x) \circ A_{x,*,j_2})) A_{x, *,j}$,
        \item $A_{6, 6} = + 2 A_x^\top \Sigma(x) \diag(\sigma_{*,i}(x) A_{x,i,j_2}) A_{x, *,j}$, and
        \item $A_{6, 7} = + 2 A_x^\top \Sigma(x) \diag(\sigma_{*,i}(x) ) A_{x, *,j_2} \circ A_{x, *,j}$.
    \end{itemize}

     For the gradient of $A_7$, we have
    \begin{align*}
        & ~ \frac{\d A_{7}}{\d x_{j_2}} \\
        = & ~ \frac{\d + 4 A_x^\top  \sigma_{*,*}^{\circ 2}(x) \sigma_{*,*}(x) \diag(\sigma_{*,i}(x))A_{x, *,j}}{\d x_{j_2}} \\
        = & ~ + 4 \frac{\d A_x^\top}{\d x_{j_2}}  \sigma_{*,*}^{\circ 2}(x) \sigma_{*,*}(x) \diag(\sigma_{*,i}(x))A_{x, *,j} \\
        & ~ + 4 A_x^\top  \frac{\d \sigma_{*,*}^{\circ 2}(x) }{\d x_{j_2}} \sigma_{*,*}(x) \diag(\sigma_{*,i}(x))A_{x, *,j} \\
        & ~ + 4 A_x^\top  \sigma_{*,*}^{\circ 2}(x) \frac{\d \sigma_{*,*}(x)}{\d x_{j_2}} \diag(\sigma_{*,i}(x))A_{x, *,j} \\
        & ~ + 4 A_x^\top  \sigma_{*,*}^{\circ 2}(x) \sigma_{*,*}(x) \frac{\d \diag(\sigma_{*,i}(x))}{\d x_{j_2}} A_{x, *,j} \\
        & ~ + 4 A_x^\top  \sigma_{*,*}^{\circ 2}(x) \sigma_{*,*}(x) \diag(\sigma_{*,i}(x)) \frac{\d A_{x, *,j}}{\d x_{j_2}} \\
        = & ~ - 4 A_x^\top \diag(A_{x, *, j_2})  \sigma_{*,*}^{\circ 2}(x) \sigma_{*,*}(x) \diag(\sigma_{*,i}(x))A_{x, *,j} \\
        & ~ + 4 A_x^\top  (4 \sigma^2_{*,*}(x) \diag (  A_{x,*,j_2} ) \sigma_{*,*}(x)  - 2 \diag( A_{x,*,j_2} )  \sigma_{*,*}^{\circ 2}(x) - 2\sigma^{\circ 2}_{*,*}(x) \diag( A_{x,*,j_2} )) \sigma_{*,*}(x) \diag(\sigma_{*,i}(x))A_{x, *,j} \\
        & ~ + 4 A_x^\top  \sigma_{*,*}^{\circ 2}(x) (2 \sigma_{*,*}(x) \diag (  A_{x,*,j_2} ) \sigma_{*,*}(x) - \diag( A_{x,*,j_2} )  \sigma_{*,*}(x) - \sigma_{*,*}(x) \diag( A_{x,*,j_2} )) \diag(\sigma_{*,i}(x))A_{x, *,j} \\
        & ~ + 4 A_x^\top  \sigma_{*,*}^{\circ 2}(x) \sigma_{*,*}(x) \diag(-  A_{x,*,j_2} \circ \sigma_{*,i}(x) +  2 \sigma_{*,*}(x)  (\sigma_{*,i}(x) \circ A_{x,*,j_2}) - \sigma_{*,i}(x) A_{x,i,j_2}) A_{x, *,j} \\
        & ~ + 4 A_x^\top  \sigma_{*,*}^{\circ 2}(x) \sigma_{*,*}(x) \diag(\sigma_{*,i}(x)) (- A_{x, *,j_2} \circ A_{x, *,j}),
    \end{align*}
    where the first step follows from the definition of $A_7$ (see Definition~\ref{def:A_i}), the second step follows from the product rule, the third step follows from combining Lemma~\ref{lem:gradient_A}, Lemma~\ref{lem:gradient_sigma_**}, Lemma~\ref{lem:gradient_sigma_*i}, and Lemma~\ref{lem:gradient_Axij_and_Axj}.

    For simplicity, we denote $A_{7, h}$ to be the $h$-th term of $\frac{\d A_{7}}{\d x_{j_2}}$.

    Therefore, we have
    \begin{align*}
        \frac{\d A_{7}}{\d x_{j_2}} = \sum_{h = 1}^{11} A_{7, h},
    \end{align*}
    where
    \begin{itemize}
        \item $A_{7, 1} = - 4 A_x^\top \diag(A_{x, *, j_2})  \sigma_{*,*}^{\circ 2}(x) \sigma_{*,*}(x) \diag(\sigma_{*,i}(x))A_{x, *,j}$,
        \item $A_{7, 2} = + 16 A_x^\top  \sigma^2_{*,*}(x) \diag (  A_{x,*,j_2} ) \sigma_{*,*}(x) \sigma_{*,*}(x) \diag(\sigma_{*,i}(x))A_{x, *,j} $,
        \item $A_{7, 3} = -8  A_x^\top  \diag( A_{x,*,j_2} )  \sigma_{*,*}^{\circ 2}(x) \sigma_{*,*}(x) \diag(\sigma_{*,i}(x))A_{x, *,j} $,
        \item $A_{7, 4} = -8 A_x^\top  \sigma^{\circ 2}_{*,*}(x) \diag( A_{x,*,j_2} ) \sigma_{*,*}(x) \diag(\sigma_{*,i}(x))A_{x, *,j} $,
        \item $A_{7, 5} = + 8 A_x^\top  \sigma_{*,*}^{\circ 2}(x)  \sigma_{*,*}(x) \diag (  A_{x,*,j_2} ) \sigma_{*,*}(x) \diag(\sigma_{*,i}(x))A_{x, *,j}$,
        \item $A_{7, 6} = - 4 A_x^\top  \sigma_{*,*}^{\circ 2}(x) \diag( A_{x,*,j_2} )  \sigma_{*,*}(x)  \diag(\sigma_{*,i}(x))A_{x, *,j}$,
        \item $A_{7, 7} = - 4 A_x^\top  \sigma_{*,*}^{\circ 2}(x) \sigma_{*,*}(x) \diag( A_{x,*,j_2} ) \diag(\sigma_{*,i}(x))A_{x, *,j}$,
        \item $A_{7, 8} = - 4 A_x^\top  \sigma_{*,*}^{\circ 2}(x) \sigma_{*,*}(x)  \diag(A_{x,*,j_2} \circ \sigma_{*,i}(x))  A_{x, *,j}$,
        \item $A_{7, 9} = + 8 A_x^\top  \sigma_{*,*}^{\circ 2}(x) \sigma_{*,*}(x) \diag(\sigma_{*,*}(x)  (\sigma_{*,i}(x) \circ A_{x,*,j_2})) A_{x, *,j}$,
        \item $A_{7, 10} = - 4 A_x^\top  \sigma_{*,*}^{\circ 2}(x) \sigma_{*,*}(x) \diag(\sigma_{*,i}(x) A_{x,i,j_2}) A_{x, *,j}$, and
        \item $A_{7, 11} = - 4 A_x^\top  \sigma_{*,*}^{\circ 2}(x) \sigma_{*,*}(x) \diag(\sigma_{*,i}(x)) A_{x, *,j_2} \circ A_{x, *,j}$.
    \end{itemize}

         For the gradient of $A_8$, we have
    \begin{align*}
        & ~ \frac{\d A_{8}}{\d x_{j_2}} \\
        = & ~ \frac{\d - 4 A_x^\top  \Sigma(x)  \sigma_{*,*}(x)    \diag(\sigma_{*,i}(x))A_{x, *,j}}{\d x_{j_2}} \\
        = & ~ - 4 \frac{\d A_x^\top}{\d x_{j_2}}  \Sigma(x)  \sigma_{*,*}(x)    \diag(\sigma_{*,i}(x))A_{x, *,j} \\
        & ~ - 4  A_x^\top  \frac{\d \Sigma(x)}{\d x_{j_2}}  \sigma_{*,*}(x)    \diag(\sigma_{*,i}(x))A_{x, *,j} \\
        & ~ - 4 A_x^\top  \Sigma(x)  \frac{\d \sigma_{*,*}(x)}{\d x_{j_2}}    \diag(\sigma_{*,i}(x))A_{x, *,j} \\
        & ~ - 4 A_x^\top  \Sigma(x)  \sigma_{*,*}(x)    \frac{\d \diag(\sigma_{*,i}(x))}{\d x_{j_2}} A_{x, *,j} \\
        & ~ - 4 A_x^\top  \Sigma(x)  \sigma_{*,*}(x)    \diag(\sigma_{*,i}(x)) \frac{\d A_{x, *,j}}{\d x_{j_2}} \\
        = & ~ + 4 A_x^\top \diag(A_{x, *, j_2})   \Sigma(x)  \sigma_{*,*}(x)    \diag(\sigma_{*,i}(x))A_{x, *,j} \\
        & ~ - 4  A_x^\top  (2 \sigma_{*,*}(x) \diag (  A_{x,*,j_2} )\Sigma(x) -2 \diag( A_{x,*,j_2} )  \Sigma(x))  \sigma_{*,*}(x)    \diag(\sigma_{*,i}(x))A_{x, *,j} \\
        & ~ - 4 A_x^\top  \Sigma(x)  (2 \sigma_{*,*}(x) \diag (  A_{x,*,j_2} ) \sigma_{*,*}(x) - \diag( A_{x,*,j_2} )  \sigma_{*,*}(x) - \sigma_{*,*}(x) \diag( A_{x,*,j_2} ))    \diag(\sigma_{*,i}(x))A_{x, *,j} \\
        & ~ - 4 A_x^\top  \Sigma(x)  \sigma_{*,*}(x)    \diag(-  A_{x,*,j_2} \circ \sigma_{*,i}(x) +  2 \sigma_{*,*}(x)  (\sigma_{*,i}(x) \circ A_{x,*,j_2}) - \sigma_{*,i}(x) A_{x,i,j_2}) A_{x, *,j} \\
        & ~ - 4 A_x^\top  \Sigma(x)  \sigma_{*,*}(x)    \diag(\sigma_{*,i}(x)) (- A_{x, *,j_2} \circ A_{x, *,j}),
    \end{align*}
    where the first step follows from the definition of $A_8$ (see Definition~\ref{def:A_i}), the second step follows from the product rule, the third step follows from combining Lemma~\ref{lem:gradient_A}, Lemma~\ref{lem:Sigma}, Lemma~\ref{lem:gradient_sigma_**}, Lemma~\ref{lem:gradient_sigma_*i}, and Lemma~\ref{lem:gradient_Axij_and_Axj}.

    For simplicity, we denote $A_{8, h}$ to be the $h$-th term of $\frac{\d A_{8}}{\d x_{j_2}}$.

    Therefore, we have
    \begin{align*}
        \frac{\d A_{8}}{\d x_{j_2}} = \sum_{h = 1}^{10} A_{8, h},
    \end{align*}
    where
    \begin{itemize}
        \item $A_{8, 1} = + 4 A_x^\top \diag(A_{x, *, j_2})   \Sigma(x)  \sigma_{*,*}(x)    \diag(\sigma_{*,i}(x))A_{x, *,j}$,
        \item $A_{8, 2} = - 8  A_x^\top  \sigma_{*,*}(x) \diag (  A_{x,*,j_2} )\Sigma(x)   \sigma_{*,*}(x)    \diag(\sigma_{*,i}(x))A_{x, *,j}$,
        \item $A_{8, 3} = + 8  A_x^\top  \diag( A_{x,*,j_2} )  \Sigma(x)  \sigma_{*,*}(x)    \diag(\sigma_{*,i}(x))A_{x, *,j}$,
        \item $A_{8, 4} = - 8 A_x^\top  \Sigma(x)  \sigma_{*,*}(x) \diag (  A_{x,*,j_2} ) \sigma_{*,*}(x)   \diag(\sigma_{*,i}(x))A_{x, *,j}$,
        \item $A_{8, 5} = + 4 A_x^\top  \Sigma(x)   \diag( A_{x,*,j_2} )  \sigma_{*,*}(x)  \diag(\sigma_{*,i}(x))A_{x, *,j}$,
        \item $A_{8, 6} = + 4 A_x^\top  \Sigma(x)  \sigma_{*,*}(x) \diag( A_{x,*,j_2} ) \diag(\sigma_{*,i}(x))A_{x, *,j}$,
        \item $A_{8, 7} =  4 A_x^\top  \Sigma(x)  \sigma_{*,*}(x)    \diag( A_{x,*,j_2} \circ \sigma_{*,i}(x) ) A_{x, *,j}$,
        \item $A_{8, 8} = - 8 A_x^\top  \Sigma(x)  \sigma_{*,*}(x)    \diag(\sigma_{*,*}(x)  (\sigma_{*,i}(x) \circ A_{x,*,j_2})) A_{x, *,j}$,
        \item $A_{8, 9} = 4 A_x^\top  \Sigma(x)  \sigma_{*,*}(x)    \diag(\sigma_{*,i}(x) A_{x,i,j_2}) A_{x, *,j}$, and
        \item $A_{8, 10} = 4 A_x^\top  \Sigma(x)  \sigma_{*,*}(x)    \diag(\sigma_{*,i}(x)) A_{x, *,j_2} \circ A_{x, *,j}$.
    \end{itemize}

    For the gradient of $A_9$, we have
    \begin{align*}
        & ~ \frac{\d A_{9}}{\d x_{j_2}} \\
        = & ~ \frac{\d - 2A_{x}^\top \sigma_{*,*}^{\circ 2}(x) \sigma_{*,i}(x) A_{x,i,j}^\top }{\d x_{j_2}} \\
        = & ~ - 2\frac{\d A_{x}^\top}{\d x_{j_2}}  \sigma_{*,*}^{\circ 2}(x) \sigma_{*,i}(x) A_{x,i,j}^\top \\
        & ~ - 2A_{x}^\top \frac{\d \sigma_{*,*}^{\circ 2}(x)  }{\d x_{j_2}} \sigma_{*,i}(x) A_{x,i,j}^\top \\
        & ~ - 2A_{x}^\top \sigma_{*,*}^{\circ 2}(x) \frac{\d \sigma_{*,i}(x)}{\d x_{j_2}} A_{x,i,j}^\top  \\
        & ~ - 2A_{x}^\top \sigma_{*,*}^{\circ 2}(x) \sigma_{*,i}(x) \frac{\d A_{x,i,j}^\top }{\d x_{j_2}} \\
        = & ~ + 2 A_x^\top \diag(A_{x, *,j_2}) \sigma_{*,*}^{\circ 2}(x) \sigma_{*,i}(x) A_{x,i,j}^\top \\
        & ~ - 2A_{x}^\top (4 \sigma^2_{*,*}(x) \diag (  A_{x,*,j_2} ) \sigma_{*,*}(x)  - 2 \diag( A_{x,*,j_2} )  \sigma_{*,*}^{\circ 2}(x) - 2\sigma^{\circ 2}_{*,*}(x) \diag( A_{x,*,j_2} )) \sigma_{*,i}(x) A_{x,i,j}^\top \\
        & ~ - 2A_{x}^\top \sigma_{*,*}^{\circ 2}(x) (-  A_{x,*,j_2} \circ \sigma_{*,i}(x) +  2 \sigma_{*,*}(x)  (\sigma_{*,i}(x) \circ A_{x,*,j_2}) - \sigma_{*,i}(x) A_{x,i,j_2}) A_{x,i,j}^\top  \\
        & ~ + 2A_{x}^\top \sigma_{*,*}^{\circ 2}(x) \sigma_{*,i}(x) A_{x, i,j_2} A_{x,i,j},
    \end{align*}
    where the first step follows from the definition of $A_9$ (see Definition~\ref{def:A_i}), the second step follows from the product rule, the third step follows from combining Lemma~\ref{lem:gradient_A}, Lemma~\ref{lem:gradient_sigma_**}, Lemma~\ref{lem:gradient_sigma_*i}, and Lemma~\ref{lem:gradient_Axij_and_Axj}.

    For simplicity, we denote $A_{9, h}$ to be the $h$-th term of $\frac{\d A_{9}}{\d x_{j_2}}$.

    Therefore, we have
    \begin{align*}
        \frac{\d A_{9}}{\d x_{j_2}} = \sum_{h = 1}^{8} A_{9, h},
    \end{align*}
    where
    \begin{itemize}
        \item $A_{9, 1} = + 2 A_x^\top \diag(A_{x, *,j_2}) \sigma_{*,*}^{\circ 2}(x) \sigma_{*,i}(x) A_{x,i,j}^\top$,
        \item $A_{9, 2} = - 8A_{x}^\top \sigma^2_{*,*}(x) \diag (  A_{x,*,j_2} ) \sigma_{*,*}(x) \sigma_{*,i}(x) A_{x,i,j}^\top $,
        \item $A_{9, 3} = + 4 A_{x}^\top \diag( A_{x,*,j_2} )  \sigma_{*,*}^{\circ 2}(x) \sigma_{*,i}(x) A_{x,i,j}^\top $,
        \item $A_{9, 4} = + 4A_{x}^\top \sigma^{\circ 2}_{*,*}(x) \diag( A_{x,*,j_2} ) \sigma_{*,i}(x) A_{x,i,j}^\top $,
        \item $A_{9, 5} = + 2A_{x}^\top \sigma_{*,*}^{\circ 2}(x) A_{x,*,j_2} \circ \sigma_{*,i}(x)  A_{x,i,j}^\top$,
        \item $A_{9, 6} = - 4A_{x}^\top \sigma_{*,*}^{\circ 2}(x) \sigma_{*,*}(x)  (\sigma_{*,i}(x) \circ A_{x,*,j_2})  A_{x,i,j}^\top$,
        \item $A_{9, 7} = + 2A_{x}^\top \sigma_{*,*}^{\circ 2}(x)  \sigma_{*,i}(x) A_{x,i,j_2} A_{x,i,j}^\top$, and
        \item $A_{9, 8} = +2A_{x}^\top \sigma_{*,*}^{\circ 2}(x) \sigma_{*,i}(x) A_{x, i,j_2} A_{x,i,j}$.
    \end{itemize}
    For the gradient of $A_{10}$, we have
    \begin{align*}
        & ~ \frac{\d A_{10}}{\d x_{j_2}} \\
        = & ~ \frac{\d +2A_{x}^\top \Sigma(x) \sigma_{*,i}(x)  A_{x,i,j}^\top }{\d x_{j_2}} \\
        = & ~ 2 \frac{\d A_{x}^\top }{\d x_{j_2}} \Sigma(x) \sigma_{*,i}(x)  A_{x,i,j}^\top  \\
        & ~ +2 A_{x}^\top \frac{\d \Sigma(x) }{\d x_{j_2}} \sigma_{*,i}(x)  A_{x,i,j}^\top \\
        & ~ +2 A_{x}^\top \Sigma(x) \frac{\d \sigma_{*,i}(x) }{\d x_{j_2}}  A_{x,i,j}^\top \\
        & ~ +2 A_{x}^\top \Sigma(x) \sigma_{*,i}(x)  \frac{\d A_{x,i,j}^\top }{\d x_{j_2}} \\
        = & ~ -2 A_x^\top \diag(A_{x, *,j_2}) \Sigma(x) \sigma_{*,i}(x)  A_{x,i,j}^\top  \\
        & ~ +2 A_{x}^\top (2 \sigma_{*,*}(x) \diag (  A_{x,*,j_2} )\Sigma(x) -2 \diag( A_{x,*,j_2} )  \Sigma(x)) \sigma_{*,i}(x)  A_{x,i,j}^\top \\
        & ~ +2 A_{x}^\top \Sigma(x) (-  A_{x,*,j_2} \circ \sigma_{*,i}(x) +  2 \sigma_{*,*}(x)  (\sigma_{*,i}(x) \circ A_{x,*,j_2}) - \sigma_{*,i}(x) A_{x,i,j_2})  A_{x,i,j}^\top \\
        & ~ -2 A_{x}^\top \Sigma(x) \sigma_{*,i}(x)  A_{x, i,j_2} A_{x,i,j},
    \end{align*}
    where the first step follows from the definition of $A_{10}$ (see Definition~\ref{def:A_i}), the second step follows from the product rule, the third step follows from combining Lemma~\ref{lem:gradient_A}, Lemma~\ref{lem:Sigma}, Lemma~\ref{lem:gradient_sigma_*i}, and Lemma~\ref{lem:gradient_Axij_and_Axj}.

    For simplicity, we denote $A_{10, h}$ to be the $h$-th term of $\frac{\d A_{10}}{\d x_{j_2}}$.

    Therefore, we have
    \begin{align*}
        \frac{\d A_{10}}{\d x_{j_2}} = \sum_{h = 1}^{7} A_{10, h},
    \end{align*}
    where
    \begin{itemize}
        \item $A_{10, 1} = -2 A_x^\top \diag(A_{x, *,j_2}) \Sigma(x) \sigma_{*,i}(x)  A_{x,i,j}^\top$,
        \item $A_{10, 2} = +4 A_{x}^\top \sigma_{*,*}(x) \diag (  A_{x,*,j_2} )\Sigma(x) \sigma_{*,i}(x)  A_{x,i,j}^\top$,
        \item $A_{10, 3} = -4 A_{x}^\top \diag( A_{x,*,j_2} )  \Sigma(x) \sigma_{*,i}(x)  A_{x,i,j}^\top$,
        \item $A_{10, 4} = -2 A_{x}^\top \Sigma(x) A_{x,*,j_2} \circ \sigma_{*,i}(x)  A_{x,i,j}^\top$,
        \item $A_{10, 5} = +4 A_{x}^\top \Sigma(x) \sigma_{*,*}(x)  (\sigma_{*,i}(x) \circ A_{x,*,j_2}) A_{x,i,j}^\top$,
        \item $A_{10, 6} = -2 A_{x}^\top \Sigma(x) \sigma_{*,i}(x) A_{x,i,j_2}  A_{x,i,j}^\top$, and
        \item $A_{10, 7} = -2 A_{x}^\top \Sigma(x) \sigma_{*,i}(x)  A_{x, i,j_2} A_{x,i,j}$.
    \end{itemize}
\end{proof}

\subsection{Hessian for $L_c(x)$}\label{sub::L_c_Hessian}

\begin{lemma}[Formal version of Lemma~\ref{lem:L_c_Hessian_informal}]\label{lem:L_c_Hessian}
If the following conditions hold
    \begin{itemize}
        \item Let $b \in \R^n$  
        be defined as in Definition~\ref{def:loss_lb}.
        \item Let $c \in \R^d$ be defined as in Definition~\ref{def:L_c}.
        \item Let $A_x \in \R^{n \times d} $ be defined as in Definition~\ref{def:A_x}.
        \item Let $\sigma_{*,*}^{\circ 2}(x) \in \R^{n \times n}$, $\sigma_{*,*}(x)  \in \R^{n \times n}$, and $\sigma_{*, i}(x)  \in \R^n $ be defined as in Definition~\ref{def:sigma}.
        \item Let $\Sigma(x)  \in \R^{n \times n} $ be defined as in Definition~\ref{def:Sigma}.
        \item Let $A_{x, *, j}  \in \R^n$ denote the $j$-th column of $A_x$.
        \item Let $g(x) \in \R^d $ be defined as in Definition~\ref{def:g(x)}
        \item Let $p(x) = \sigma_{*, i}(x) - b \in \R^n$.
        \item Let $q(x) = g(x) - c \in \R^d$.
        \item Let $L_{c,j_0}:= (g(x)_{j_0} - c_{j_0})^2$
        \item Let $A_l$ be defined as in Definition~\ref{def:A_i}.
        \item Let $\wt{A}_l$ be $A_l$ where all $j$ is replaced by $j_2$.
    \end{itemize}

 Then, we have 
    \begin{itemize}
        \item Part 1. For all $j \in [d]$,
        \begin{align*}
            \frac{\d^2 L_{c,j_0} (x)}{\d x_{j} \d x_{j_2} } = 2(\sum_{l=1}^{10} \wt{A}_l \sum_{l=1}^{10} A_l)_{j_0}  + ((g(x) - c)  \sum_{l=1}^{10} \mathsf{A}_l )_{j_0}
        \end{align*}
        \item Part 2. For all $j \in [d]$,
        \begin{align*}
            \frac{\d^2 L_{c} (x)}{\d x_{j} \d x_{j_2} } = (\sum_{l=1}^{10} \wt{A}_l)^\top \sum_{l=1}^{10} A_l(g(x) - c)^\top  \sum_{l=1}^{10} \mathsf{A}_l
        \end{align*}
        \item Part 3. For all $j \in [d]$,
        \begin{align*}
            \frac{\d^2 L_{c} (x)}{\d x^2 } =A_{x}^\top B(x) A_{x} 
        \end{align*}
    \end{itemize}
\end{lemma}

\begin{proof}

We have
\begin{align*}
    \frac{\d^2 L_{c,j_0} (x)}{\d x_{j} \d x_{j_2} } 
    = & ~ \frac{\d^2 (g(x)_{j_0} - c_{j_0})^2}{\d x_{j} \d x_{j_2} }\\
    = & ~ \frac{\d 2 (g(x)_{j_0} - c_{j_0}) (\sum_{l=1}^{10} A_l)_{j_0} }{ \d x_{j_2} }\\
    = & ~ 2\frac{\d g(x)_{j_0} - c_{j_0}}{ \d x_{j_2} } (\sum_{l=1}^{10} A_l)_{j_0}  + 2 (g(x)_{j_0} - c_{j_0}) (\frac{\d \sum_{l=1}^{10} A_l }{ \d x_{j_2} })_{j_0}\\
    = & ~ 2(\sum_{l=1}^{10} \wt{A}_l)_{j_0} (\sum_{l=1}^{10} A_l)_{j_0}  + 2 (g(x)_{j_0} - c_{j_0}) (\frac{\d \sum_{l=1}^{10} A_l }{ \d x_{j_2} })_{j_0}\\
    = & ~ 2(\sum_{l=1}^{10} \wt{A}_l)_{j_0} (\sum_{l=1}^{10} A_l)_{j_0}  + 2 (g(x)_{j_0} - c_{j_0}) ( \sum_{l=1}^{10} \mathsf{A}_l )_{j_0}\\
    = & ~ 2(\sum_{l=1}^{10} \wt{A}_l \sum_{l=1}^{10} A_l)_{j_0}  + ((g(x) - c)  \sum_{l=1}^{10} \mathsf{A}_l )_{j_0},
\end{align*}
where the first step follows from the definition of $L_{c,j_0} (x)$ (see the lemma statement), the second step follows from the gradient of $g(x)$ (see Lemma~\ref{lem:g(x)}), the third step follows from the product rule, the fourth step follows from the gradient of $g(x)$ (see Lemma~\ref{lem:g(x)}), the fifth step follows from Lemma~\ref{lem:gradient_Al}, and the last step follows from simple algebra.

We have
\begin{align*}
    \frac{\d^2 L_{c} (x)}{\d x_{j} \d x_{j_2} } 
    = & ~ 0.5 \cdot \frac{\d^2 \| g(x) - c\|_2^2}{\d x_{j} \d x_{j_2} } \\
    = & ~ 0.5 \cdot \frac{\d^2 (g(x) - c)^\top (g(x) - c)}{\d x_{j} \d x_{j_2} } \\
    = & ~ \frac{\d (g(x) - c)^\top \frac{\d g(x)}{\d x_j}}{\d x_{j_2} } \\
    = & ~ \frac{\d (g(x) - c)^\top \sum_{l=1}^{10} A_l}{\d x_{j_2} } \\
    = & ~ \frac{\d g(x)^\top}{\d x_{j_2} }  \sum_{l=1}^{10} A_l + (g(x) - c)^\top \frac{\d \sum_{l=1}^{10} A_l}{\d x_{j_2} } \\
    = & ~ \underbrace{(\sum_{l=1}^{10} \wt{A}_l)^\top}_{1 \times d} \underbrace{\sum_{l=1}^{10} A_l}_{d \times 1} + \underbrace{(g(x) - c)^\top}_{1 \times d} \underbrace{\frac{\d \sum_{l=1}^{10} A_l}{\d x_{j_2} }}_{d \times 1} \\
    = & ~ \underbrace{(\sum_{l=1}^{10} \wt{A}_l)^\top}_{1 \times d} \underbrace{\sum_{l=1}^{10} A_l}_{d \times 1} + \underbrace{(g(x) - c)^\top}_{1 \times d} \underbrace{\sum_{l=1}^{10} \mathsf{A}_l}_{d \times 1},
\end{align*}
where the first step follows from the definition of $L_c(x)$ (see Definition~\ref{def:L_c}), the second step follows from the fact that $\|v\|_2^2 $ is equivalent to $ v^\top v $, the third step follows from the product rule, the fourth step follows from the gradient of $g(x)$ (see Lemma~\ref{lem:g(x)}), the fifth step follows from the product rule, the sixth step follows from the gradient of $g(x)$ (see Lemma~\ref{lem:g(x)}), the last step follows from Lemma~\ref{lem:gradient_Al}.

We first consider
\begin{align*}
    \underbrace{(\sum_{l=1}^{10} \wt{A}_l)^\top}_{1 \times d} \underbrace{\sum_{l=1}^{10} A_l}_{d \times 1}.
\end{align*}
Note that from Definition~\ref{def:A_i}, for each $l \in [10]$, $A_l \in \R^d$ is equal to some term $D_l \in \R^{d \times n}$ multiplying with $A_{x, *, j} \in \R^n$. Similarly, for each $l \in [10]$, $\wt{A}_l \in \R^d$ is equal to some term $D_l \in \R^{d \times n}$ multiplying with $A_{x, *, j_2} \in \R^n$. Therefore, we have
\begin{align*}
    \underbrace{(\sum_{l=1}^{10} \wt{A}_l)^\top}_{1 \times d} \underbrace{\sum_{l=1}^{10} A_l}_{d \times 1}
    = & ~ \underbrace{(\sum_{l=1}^{10} D_l A_{x, *, j_2})^\top}_{1 \times d} \underbrace{\sum_{l=1}^{10} D_l A_{x, *, j}}_{d \times 1}\\
    = & ~ \underbrace{\sum_{l=1}^{10} A_{x, *, j_2}^\top D_l^\top}_{1 \times d} \underbrace{\sum_{l=1}^{10} D_l A_{x, *, j}}_{d \times 1}\\
    = & ~ \underbrace{A_{x, *, j_2}^\top}_{1 \times n} \underbrace{\sum_{l=1}^{10} D_l^\top}_{n \times d} \underbrace{\sum_{l=1}^{10} D_l}_{d \times n} \underbrace{A_{x, *, j}}_{n \times 1}\\
    = & ~ \underbrace{A_{x, *, j_2}^\top}_{1 \times n}  \underbrace{B_1(x)}_{n \times n} \underbrace{A_{x, *, j}}_{n \times 1},
\end{align*}
where the first step follows from replace $\wt{A}_l$ and $A_l$ with their specific forms, the second
step follows from the fact~\ref{fac:matrix_norm}, the third step follows from the simple algebra, the last step follows from the definition of $B_1(x)$.

Second, we consider
\begin{align*}
    \underbrace{(g(x) - c)^\top}_{1 \times d} \underbrace{\sum_{l=1}^{10} \mathsf{A}_l}_{d \times 1}.
\end{align*}

We use the same technique to transform this to 
\begin{align*}
    \underbrace{A_{x, *, j_2}^\top}_{1 \times n}  \underbrace{B_2(x)}_{n \times n} \underbrace{A_{x, *, j}}_{n \times 1}.
\end{align*}

Combining everything together, we have
\begin{align*}
    \frac{\d^2 L_c(x)}{\d x^2} 
    = & ~ \underbrace{A_{x}^\top}_{d \times n}  \underbrace{B_1(x)}_{n \times n} \underbrace{A_{x}}_{n \times d} + \underbrace{A_{x}^\top}_{d \times n}  \underbrace{B_2(x)}_{n \times n} \underbrace{A_{x}}_{n \times d}\\
    = & ~ \underbrace{A_{x}^\top}_{d \times n}  \underbrace{(\underbrace{B_1(x)}_{n \times n} +  \underbrace{B_2(x)}_{n \times n})}_{:= B(x)} \underbrace{A_{x}}_{n \times d},
\end{align*}
where the first step follows from the above results, the last step follows from the simple algebra.
\end{proof}

\section{Hessian is Positive Definite}\label{sec:hessian_positive_definite}

In Section~\ref{sub:preliminary}, we present the preliminary. In Section~\ref{sec:preli:reg}, we present the regularizations. In Section~\ref{sec:psd:final}, we present the positive definite of the Hessian.

\subsection{Preliminary}\label{sub:preliminary}

\begin{lemma}[Lemma D.1 on page 40 of \cite{lsw+24}]
\label{lem:spectral_norm_bound}
If we have:
\begin{itemize}
    \item The spectral norm of $A$ is bounded by $R$.
    \item The $\ell_2$ norm of $x$ is bounded by $R$.
    \item $S(x)$ is defined as per Definition~\ref{def:s_x}.
    \item Let $\sigma_{*,*}^{\circ 2}(x) \in \R^{n \times n}$, $\sigma_{*,*}(x)  \in \R^{n \times n}$, and $\sigma_{*, i}(x)  \in \R^n $ be defined as in Definition~\ref{def:sigma}.
    \item The $\sigma_{\min }(A_x)$ is greater than or equal to $\beta$.
\end{itemize}

Then we have

\begin{itemize}
        \item Part 1. $\|\sigma_{*, *}(x)\| \leq 1$
        \item Part 2. $|\sigma_{i,i}(x)| \leq 1$
        \item Part 3. $\|\sigma_{*,i}(x)\|_2 \leq 1$
        \item Part 4. $\|A_x^{-1}\| \leq \beta^{-1}$
        \item Part 5. $\|(A_x^\top A_x)^{-1}\| \leq \beta^{-2}$
\end{itemize}
\end{lemma}
\begin{proof}
{\bf Proof of Part 1.}

If a matrix $P$ satisfy that $P^2=P$, then $P$ is called a projection matrix.

By property of projection matrix, we know that $\|P\| \leq 1$.

Since $\sigma_{*, *}(x) \sigma_{*, *}(x)=\sigma_{*, *}(x)$, thus $\sigma_{*, *}(x)$ is a projection matrix, thus, we have $\|\sigma_{*, *}(x)\| \leq 1$.

{\bf Proof of Part 2.}

It follows from Part 1 directly.

{\bf Proof of Part 3.}

It follows from Part 2 directly.

{\bf Proof of Part 4.}

We can show

\begin{align*}
\|{A_x}^{-1}\|=\sigma_{\min }(A_x)^{-1} \leq \beta^{-1}
\end{align*}

{\bf Proof of Part 5.}

We can show

\begin{align*}
\|({A_x}^{\top} A_x)^{-1}\|=\sigma_{\min }(A_x)^{-2} \leq \beta^{-2}
\end{align*}
\end{proof}

\begin{fact}[Schur’s inequality, Corollary 6 in \cite{l11}]\label{fac:schur_inequality}
Let $A$ and $B$ be two arbitrary matrices.
\begin{align*}
\|A \circ B\| \leq \|A\| \cdot \|B\|
\end{align*}
\end{fact}

\begin{lemma}\label{lem:more_spectral_norm_bound}
If the following conditions hold
\begin{itemize}
    \item Let $\|b\|_2 \leq 1$.
    \item Let $b \in \R^{n}$, $\sigma_{*, *}^{\circ 2} \in \R^{n \times n}$, and $\sigma_{*, i}(x) \in \R^{n}$ be defined as in Definition~\ref{def:sigma}.
    \item Let $p(x) = \sigma_{*, i}(x) - b \in \R^{n}$
    \item Let $\diag(p(x)) \in \R^{n \times n}$.
\end{itemize}

 Then, we have
 
    \begin{itemize}
        \item Part 1. $\|\sigma_{*, *}^{\circ 2}(x)\| \leq 1$
        \item Part 2. $\|\diag(p(x))\| \leq 2$
        \item Part 3. $\|\Sigma(x)\| \leq 1$
    \end{itemize}
\end{lemma}

\begin{proof}

{\bf Proof of Part 1.}

We have
\begin{align*}
\|\sigma_{*, *}^{\circ 2}(x)\| & = ~\| \sigma_{*, *}(x) \circ \sigma_{*, *}(x)\|\\
& \leq  ~\| \sigma_{*, *}(x) \| \cdot \| \sigma_{*, *}(x)\|\\
& \leq ~ 1
\end{align*}
where the first step follows from the definition of $\sigma_{*,*}^{\circ 2 }(x)$(see Definition~\ref{def:sigma}), the second step follows from the Fact~\ref{fac:schur_inequality} the last step follows from the Part 1 of Lemma~\ref{lem:spectral_norm_bound}.

{\bf Proof of Part 2.}

We have
\begin{align*}
\|\diag(p(x))\| & = ~ \|\diag(\sigma_{*, i}(x) - b)\| \\
& \leq ~ \|\diag(\sigma_{*, i}(x))\| + \|\diag(b)\| \\
& \leq ~ 1 + 1 \\
& = ~  2
\end{align*}
where the first step follows from the definition of $p(x)$(see the lemma statement), the second step follows from the Fact~\ref{fac:vector}, the third step follows from the definition of $(\sigma_{*, i}(x)$ and $b$(see the lemma statement), the last step follows from the simple algebra.

{\bf Proof of Part 3.}

This follows from the definition of $\Sigma(x)$
\end{proof}

\begin{lemma}\label{lem:spectral_B}

If the following conditions hold
\begin{itemize}
    \item Let $B_1(x), B_2(x), B(x) \in \R^{n \times n}$ be computed as in Lemma~\ref{lem:L_c_Hessian}. 
    \item Let $\|b\|_2 \leq 1$.
    \item Let $b \in \R^{n}$, $\sigma_{*, *}^{\circ 2} \in \R^{n \times n}$, and $\sigma_{*, i}(x) \in \R^{n}$ be defined as in Definition~\ref{def:sigma}.
    \item Let $p(x) = \sigma_{*, i}(x) - b \in \R^{n}$
    \item Let $\|A_x\|\leq R$ for some $R>0$ and $\sigma_{\min}(A_x)\geq \beta$, where $\beta \in (0, 0.1)$.
    \item Let $\Sigma(x) \in \R^{n \times n}$ be defined as in Definition~\ref{def:Sigma}.
\end{itemize}
    
Then, we have
    \begin{itemize}
        \item Part 1. $-4100 I \preceq B_1(x) \preceq 4100 I $
        \item Part 2. $-7660 \beta R I \preceq B_2(x) \preceq 7660 \beta R I $
        \item Part 3. $-12000 \beta R I \preceq B(x) \preceq 12000 \beta R I$
    \end{itemize}
\end{lemma}
\begin{proof}
    {\bf Proof of Part 1.}
    
    By Lemma~\ref{lem:L_c_Hessian}, we have
    \begin{align}\label{equ:B_1}
        B_1(x) = \underbrace{\sum_{l=1}^{10} D_l^\top}_{n \times d} \underbrace{\sum_{l=1}^{10} D_l}_{d \times n}.
    \end{align}

    Therefore, we have
    \begin{align}\label{eq:B_1x_bound}
        \|B_1(x)\|
        = & ~ \|\sum_{l=1}^{10} D_l^\top \sum_{l=1}^{10} D_l\| \notag\\
        \leq & ~ \|\sum_{l=1}^{10} D_l^\top\| \cdot \|\sum_{l=1}^{10} D_l\| \notag\\
        = & ~ \|\sum_{l=1}^{10} D_l\| \cdot \|\sum_{l=1}^{10} D_l\| \notag\\
        \leq & ~ (\sum_{l=1}^{10} \| D_l \|)^2,
    \end{align}
    Where the first step follows from Equation~\eqref{equ:B_1}, the second step, the third step and the fourth step follows from Fact \ref{fac:matrix_norm}.

    Note that by Lemma~\ref{lem:spectral_norm_bound} and Lemma~\ref{lem:more_spectral_norm_bound}, we have
    \begin{itemize}
        \item $\|\sigma_{*, *}(x)\| \leq 1$
        \item $|\sigma_{i,i}(x)| \leq 1$
        \item $\|\sigma_{*,i}(x)\|_2 \leq 1$
        \item $\|\sigma_{*, *}^{\circ 2}(x)\| \leq 1$
        \item $\|\diag(p(x))\| \leq 2$
\end{itemize}
    Therefore, since $A_l = D_l \cdot A_{x, *, j}$, by Definition of $A_l$ (see Definition~\ref{def:A_i}), we can get the bounds for $\|D_l\|$, for all $l \in [10]$, namely
    \begin{itemize}
        \item $\|D_1\| \leq 20$, 
        \item $\|D_2\| \leq 4$, 
        \item $\|D_3\| \leq 16$, 
        \item $\|D_4\| \leq 8$, 
        \item $\|D_5\| \leq 2$, 
        \item $\|D_6\| \leq 2$, 
        \item $\|D_7\| \leq 4$, 
        \item $\|D_8\| \leq 4$, 
        \item $\|D_9\| \leq 2$, and
        \item $\|D_{10}\| \leq 2$.
    \end{itemize}

    Combining with Eq.~\eqref{eq:B_1x_bound}, we have
    \begin{align*}
        \|B_1(x)\| \leq (20 + 4 + 16 + 8 + 2 + 2 + 4 + 4 + 2 + 2)^2 \leq 4100.
    \end{align*}

    {\bf Proof of Part 2.}
By Lemma~\ref{lem:L_b}, we have
\begin{align*}
        g(x) = \underbrace{ \frac{\d L_b(x)}{\d x} }_{d \times 1}
        = 2 \underbrace{A_{x}^\top}_{d \times n} (\underbrace{\sigma_{*,*}^{\circ 2}(x)}_{n \times n} - \underbrace{\Sigma(x)}_{n \times n}) \underbrace{p(x)}_{n \times 1}
\end{align*}

The spectral norm of $B_2(x)$ is bounded by the sum of the coefficients of $\mathsf{A}_l$ multiplying with $\|g(x) - c\|_2$.

\begin{align*}
        \|g(x) - c\|_2
        \leq & ~ \|g(x)\|_2 + \|c\|_2\\
        \leq & ~ \|2 A_{x}^\top \sigma_{*,*}^{\circ 2}(x) p(x) - 2A_{x}^\top \Sigma(x) p(x)\|_2 + 1\\
        \leq & ~ \|2 A_{x}^\top \sigma_{*,*}^{\circ 2}(x) p(x)\|_2 + \|2A_{x}^\top \Sigma(x) p(x)\|_2 + 1\\
        \leq & ~ 2\|A_{x}\| \|\sigma_{*,*}^{\circ 2}(x)\| \|p(x)\|_2 + 2\|A_{x}\| \|\Sigma(x)\| \|p(x)\|_2 + 1\\
        \leq & ~ 4 \beta R + 4 \beta R + 1\\
        = & ~ 8\beta R + 1\\
        \leq & ~ 10\beta R
\end{align*}
where the first step follows from the Fact~\ref{fac:matrix_norm}, the second step follows from the definition of $ g(x) $(see the definition above), the third step follows from the Fact~\ref{fac:matrix_norm}, the fourth step follows from the Fact~\ref{fac:matrix_norm}, the fifth step follows from the definition of $\|A_x\|$, $\sigma_{*,*}^{\circ 2}(x)$, $p(x)$ and $\Sigma(x)$(see the Lemma statement), the last step follows from simple algebra.
 
By Definition of $A_l$ (see Definition~\ref{def:A_i}), we can get the bounds for $A_l$, for all $l \in [10]$, namely
\begin{itemize}
        \item $\|A_1\| \leq 20 + 40 + 20 + 40 + 10 + 20 + 10 + 20 = 180$, 
        \item $\|A_2\| \leq 4 + 8 + 8 + 2 + 4 + 2 + 4 = 32$, 
        \item $\|A_3\| \leq 16 + 32 + 16 + 16 + 64 + 32 + 32 + 8 + 16 + 8 + 16 =256$, 
        \item $\|A_4\| \leq 8 + 16 + 16 + 16 + 8 + 8 + 4 + 8 + 4 + 8 =96$, 
        \item $\|A_5\| \leq 2 + 8 + 4 + 4 + 2 + 4 + 2 + 2 = 28$, 
        \item $\|A_6\| \leq 2 + 4 + 4 + 2 + 4 + 2 + 2 = 20$, 
        \item $\|A_7\| \leq 4 + 16 + 8 + 8 + 8 + 12 + 16 = 72$, 
        \item $\|A_8\| \leq 24 + 32 = 56$, 
        \item $\|A_9\| \leq 20 + 8 = 28$, and
        \item $\|A_{10}\| \leq 20$.
\end{itemize}

Thus, we have
\begin{align*}
    \|B_2(x)\|
    \leq & ~(180+32+256+96+28+20+72+56+28+20)\cdot 10\beta R \\
    < & ~ 790 \cdot 10\beta R\\
    = & ~ 7900 \beta R
\end{align*}
where the first step follows from the combination of the results above, the second step follows the simple algebra, the last step follows the simple algebra.

{\bf Proof of Part 3.}

It follows from combining {\bf Part 1} and {\bf Part 2}.
\end{proof}

\subsection{Hessian is Positive Definite}\label{sec:psd:final}

\begin{lemma}[Formal version of Lemma~\ref{lem:convex_informal}]\label{lem:convex}
If we have:
\begin{itemize}
    \item Let $A_x = S_x^{-1} A \in \R^{n \times d}$ be defined as in Definition~\ref{def:A_x}, where $\|S_x\| \geq \beta$, for $\beta \in (0, 0.1)$.
    \item Let $L_c(x) := 0.5 \cdot \| g(x) - c\|_2^2$ (see Definition~\ref{def:L_c}).
    \item Let $L_{\mathrm{reg}}(x)$ be defined as Definition~\ref{def:L_reg}.
    \item Let $L(x) = L_{c}(x) + L_{\mathrm{reg}}(x)$. 
    \item Let $W = \diag(w) \in \R^{n \times n}$. 
    \item Let $W^2 \in \R^{n \times n}$ denote the matrix that $i$-th diagonal entry is $w_{i,i}^2$.
    \item Let $\sigma_{\min}(A)$ denote the minimum singular value of $A$.
    \item Let $l > 0$ denote a scalar.
    \item Let $w_{i}^2 \geq -12000 \beta^3 R + l/\sigma_{\min}(A)^2$
\end{itemize}
Then, we have
    \begin{align*}
    \frac{\d^2 L}{\d x^2} \succeq l \cdot I_d
    \end{align*}
\end{lemma}

\begin{proof}

    By Lemma~\ref{lem:spectral_B}, we show
    \begin{align*}
        -12000 \beta R  I \preceq B(x) \preceq 12000 \beta R  I.
    \end{align*}

    Additionally, by Lemma~\ref{lem:L_c_Hessian}, we have
    \begin{align*}
        \frac{\d^2 L_{c} (x)}{\d x^2 } 
        = & ~ A_{x}^\top B(x) A_{x}\\
        = & ~ A^\top S_{x}^{-1} B(x) S_{x} A,
    \end{align*}

    This implies
    \begin{align}
        -12000 \beta^3 R I \preceq S_{x}^{-1} B(x) S_{x} \preceq 12000 \beta^3 R I
    \end{align}

    Therefore, by defining $G(x) := S_{x}^{-1} B(x) S_{x}$, we have
    \begin{align*}
        \frac{\d^2 L}{\d x^2} 
        = & ~ \frac{\d^2 L_{c}}{\d x^2} + \frac{\d^2 L_{\mathrm{reg}}
        }{\d x^2} \\
        = & ~ A^\top G(x) A + A^\top W^2 A \\
        = & ~ A^\top (G(x) + W^2) A
    \end{align*}
    where the first step follows from the definition of $L$(see the Lemma statement), the second step follows from the definition of $L_{c}$ and $L_{\mathrm{reg}}$(see the Lemma statement), the last step follows from simple algebra.

    Then we can write $\frac{\d^2 L}{\d x^2}$ as
    \begin{align*}
        \frac{\d^2 L}{\d x^2} = A^\top D A
    \end{align*}
    where
    \begin{align*}
        D = G(x) + W^2
    \end{align*}

    We can then bound $D$ as follows
    \begin{align*}
        D
        \succeq & ~ -12000 \beta^3 R I_n + w_{\min}^2 I_n \\
        = & ~ ( -12000 \beta^3 R + w_{\min}^2) I_n \\ 
        \succeq & ~ \frac{l}{\sigma_{\min}(A)^2} I_n
    \end{align*}
    where the first step follows from the definition of $D$ above, the second step follows from the simple algebra, and the last step follows from the definition of $w_{i}^2$ (see the Lemma statement).

    Since $D$ is positive definite, then we have 
    \begin{align*}
        A^\top D A \succeq \sigma_{\min}(D) \cdot \sigma_{\min}(A)^2 I_d \succeq l \cdot I_d
    \end{align*}
    Thus, Hessian is positive definite forever and thus $L$ is convex.
\end{proof}

\section{Hessian is Lipschitz Continuous}\label{sec:hessian_lipschitz_continuous}

In Section~\ref{sub:definition}, we present the definition. In Section~\ref{sub:basic_facts}, we present the basic facts.

\subsection{Definition}\label{sub:definition}

\begin{definition}[Hessian is $M$-Lipschitz]\label{def:hessian_lipschitz}

Consider a function $L : \mathbb{R}^d \rightarrow \mathbb{R}$. Let $M > 0$. We say that the Hessian matrix of $L$ is $M$-Lipschitz if for all $x$ and $y$ in $\mathbb{R}^d$,
    $\| \nabla^2 L(y) - \nabla^2 L(x) \| \leq M \cdot \| y - x \|_2$.
\end{definition}

\subsection{Basic Facts}\label{sub:basic_facts}

\begin{lemma}[Lemma E.1 in \cite{lsw+24}]\label{lem:basic_lips}
    If we have:
    \begin{itemize}
        \item Let $\|A\| \leq R$
        \item Let $\|A(x - \wh{x})\| \leq 0.01$
        \item Let $S_x$ be defined in Definition~\ref{def:s_x}
        \item Let $\|S_x\| \geq \beta$ where $\beta \in (0, 0.1)$ 
        \item Let $A_x$ be defined in Definition~\ref{def:A_x}.
        \item Define $\sigma_{*, *}(x)$ according to Definition~\ref{def:sigma}.
    \end{itemize}
    Then we have
    \begin{itemize}
        \item {\bf Part 1.} $\|S_x - S_{\wh{x}}\| \leq R \|x - \wh{x}\|_2$
        \item {\bf Part 2.} $\|S_x^{-1} - S_{\wh{x}}^{-1}\| \leq \beta^{-2} R \|x - \wh{x}\|_2$
        \item {\bf Part 3.} $\|A_x - A_{\wh{x}}\| \leq \beta^{-2} R^2 \|x - \wh{x}\|_2$
        \item {\bf Part 4.} $\|A_x^{-1} - A_{\wh{x}}^{-1}\| \leq \beta^{-4} R^2 \|x - \wh{x}\|_2$
        \item {\bf Part 5.} $\| (A_x^{\top } A_x )^{-1} - (A_{\wh{x}}^{\top } A_{\wh{x}} )^{-1}\| \leq 2\beta^{-5} R^2 \|x - \wh{x}\|_2$
        \item {\bf Part 6.} $\|\Sigma(x) - \Sigma(\wh{x})\| \leq \|\sigma_{*, *}(x) - \sigma_{*, *}(\wh{x})\| \leq 3\beta^{-7} R^3 \|x - \wh{x}\|_2$
        \item {\bf Part 7.} $|\sigma_{i,i}(x) - \sigma_{i,i}(\wh{x})| \leq  \| \sigma_{*, *}(x) - \sigma_{*, *}( \wh{x} ) \| \leq 3\beta^{-7} R^3 \|x - \wh{x}\|_2$
        \item {\bf Part 8.} $\|p(x) - p(\wh{x})\|_2 = \|\sigma_{*,i}(x) - \sigma_{*,i}(\wh{x})\|_2 \leq \| \sigma_{*, *}(x) - \sigma_{*, *}( \wh{x} ) \| \leq 3\beta^{-7} R^3 \|x - \wh{x}\|_2$
    \end{itemize}
\end{lemma}

\begin{lemma}\label{lem:more_lips}
    If we have:
    \begin{itemize}
        \item Let $\sigma_{*,*}^{\circ 2}(x) \in \R^{n \times n}$, $\sigma_{*,*}(x)  \in \R^{n \times n}$, and $\sigma_{*, i}(x)  \in \R^n $ be defined as in Definition~\ref{def:sigma}.
        \item Let $g(x) \in \R^d $ be defined as in Definition~\ref{def:g(x)}
        \item Let $\beta \in (0, 0.1)$ 
    \end{itemize}

    Then, we have
    \begin{itemize}
        \item Part 1. $\|\sigma_{*, *}^{\circ 2}(x) - \sigma_{*, *}^{\circ 2}(\wh{x})\| \leq 6\beta^{-7} R^3 \|x - \wh{x}\|_2$
        \item Part 2. $\|g(x) - g(\wh{x})\|_2 \leq 60 \beta^{-7} R^4 \|x - \wh{x}\|_2$
    \end{itemize}
\end{lemma}
\begin{proof}
    {\bf Proof of Part 1.}
    We have
    \begin{align*}
        & ~ \|\sigma_{*, *}^{\circ 2}(x) - \sigma_{*, *}^{\circ 2}(\wh{x})\|\\
        = & ~ \|\sigma_{*, *}(x) \circ \sigma_{*, *}(x) - \sigma_{*, *}(\wh{x}) \circ \sigma_{*, *}(\wh{x})\|\\
        \leq & ~ \|\sigma_{*, *}(x) \circ \sigma_{*, *}(x) - \sigma_{*, *}(x) \circ \sigma_{*, *}(\wh{x})\| + \|\sigma_{*, *}(x) \circ \sigma_{*, *}(\wh{x}) - \sigma_{*, *}(\wh{x}) \circ \sigma_{*, *}(\wh{x})\|\\
        = & ~ \|\sigma_{*, *}(x) \circ (\sigma_{*, *}(x) -  \sigma_{*, *}(\wh{x}))\| + \|(\sigma_{*, *}(x) - \sigma_{*, *}(\wh{x})) \circ \sigma_{*, *}(\wh{x})\|\\
        \leq & ~ \|\sigma_{*, *}(x)\| \|\sigma_{*, *}(x) -  \sigma_{*, *}(\wh{x})\| + \|\sigma_{*, *}(x) - \sigma_{*, *}(\wh{x})\| \|\sigma_{*, *}(\wh{x})\|\\
        = & ~ 2\|\sigma_{*, *}(x)\| \|\sigma_{*, *}(x) -  \sigma_{*, *}(\wh{x})\|\\
        \leq & ~ 6\beta^{-7} R^3 \|x - \wh{x}\|_2,
    \end{align*}
    Where the first step follows from the definition of $\sigma_{*,*}^{\circ 2 }(x)$(see the Definition~\ref{def:sigma}), the second step follows from Fact~\ref{fac:matrix_norm}, the third step follows from Fact~\ref{fac:vector} the fact that sigma is a symmetric matrix, the fourth step follows from Fact~\ref{fac:matrix_norm}, the fifth step follows from simple algebra, and the final step follows from {\bf Part 1} of Lemma~\ref{lem:spectral_norm_bound} and {\bf Part7} of Lemma~\ref{lem:basic_lips}.

    {\bf Proof of Part 2.}

    Now, we have
    \begin{align}\label{eq:g_gwh}
        & ~ \|g(x) - g(\wh{x})\|_2 \notag\\
        = & ~ 2\|(\underbrace{A_{x}^\top}_{d \times n} (\underbrace{\sigma_{*,*}^{\circ 2}(x)}_{n \times n} - \underbrace{\Sigma(x)}_{n \times n}) \underbrace{p(x)}_{n \times 1}) - (\underbrace{A_{\wh{x}}^\top}_{d \times n} (\underbrace{\sigma_{*,*}^{\circ 2}(\wh{x})}_{n \times n} - \underbrace{\Sigma(\wh{x})}_{n \times n}) \underbrace{p(\wh{x})}_{n \times 1})\|_2 \notag\\
        = & ~ 2\|\underbrace{A_{x}^\top}_{d \times n} \underbrace{\sigma_{*,*}^{\circ 2}(x)}_{n \times n}\underbrace{p(x)}_{n \times 1} - \underbrace{A_{x}^\top}_{d \times n} \underbrace{\Sigma(x)}_{n \times n} \underbrace{p(x)}_{n \times 1} - \underbrace{A_{\wh{x}}^\top}_{d \times n} \underbrace{\sigma_{*,*}^{\circ 2}(\wh{x})}_{n \times n}\underbrace{p(\wh{x})}_{n \times 1} + \underbrace{A_{\wh{x}}^\top}_{d \times n} \underbrace{\Sigma(\wh{x})}_{n \times n} \underbrace{p(\wh{x})}_{n \times 1}\|_2 \notag\\
        \leq & ~ 2\|\underbrace{A_{x}^\top}_{d \times n} \underbrace{\sigma_{*,*}^{\circ 2}(x)}_{n \times n}\underbrace{p(x)}_{n \times 1} - \underbrace{A_{\wh{x}}^\top}_{d \times n} \underbrace{\sigma_{*,*}^{\circ 2}(\wh{x})}_{n \times n}\underbrace{p(\wh{x})}_{n \times 1}\|_2 + 2\|\underbrace{A_{x}^\top}_{d \times n} \underbrace{\Sigma(x)}_{n \times n} \underbrace{p(x)}_{n \times 1} - \underbrace{A_{\wh{x}}^\top}_{d \times n} \underbrace{\Sigma(\wh{x})}_{n \times n} \underbrace{p(\wh{x})}_{n \times 1}\|_2,
    \end{align}
    Where the first step follows from the definition of $g(x)$(see the Definition~\ref{def:g(x)}), the second step follows from simple algebra, the last step follows from the Fact~\ref{fac:matrix_norm}.

    First, we consider
    \begin{align}\label{eq:a_sigma_p_a}
        & ~ \|A_{x}^\top \sigma_{*,*}^{\circ 2}(x) p(x) - A_{\wh{x}}^\top \sigma_{*,*}^{\circ 2}(\wh{x})p(\wh{x})\|_2 \notag\\
        \leq & ~ \|A_{x}^\top \sigma_{*,*}^{\circ 2}(x) p(x) - A_{\wh{x}}^\top \sigma_{*,*}^{\circ 2}(x)p(x)\|_2 + \|A_{\wh{x}}^\top \sigma_{*,*}^{\circ 2}(x) p(x) - A_{\wh{x}}^\top \sigma_{*,*}^{\circ 2}(\wh{x})p(x)\|_2 \notag\\
        & ~ + \|A_{\wh{x}}^\top \sigma_{*,*}^{\circ 2}(\wh{x}) p(x) - A_{\wh{x}}^\top \sigma_{*,*}^{\circ 2}(\wh{x})p(\wh{x})\|_2 \notag\\
        = & ~ \|(A_{x}^\top - A_{\wh{x}}^\top) \sigma_{*,*}^{\circ 2}(x)p(x)\|_2 + \|A_{\wh{x}}^\top (\sigma_{*,*}^{\circ 2}(x)  - \sigma_{*,*}^{\circ 2}(\wh{x})) p(x)\|_2 + \|A_{\wh{x}}^\top \sigma_{*,*}^{\circ 2}(\wh{x}) (p(x) - p(\wh{x}))\|_2 \notag\\
        \leq & ~ \|A_{x}^\top - A_{\wh{x}}^\top\| \|\sigma_{*,*}^{\circ 2}(x)\| \|p(x)\|_2 + \|A_{\wh{x}}^\top\| \|\sigma_{*,*}^{\circ 2}(x)  - \sigma_{*,*}^{\circ 2}(\wh{x})\| \|p(x)\|_2 \notag\\
        & ~ + \|A_{\wh{x}}^\top\| \|\sigma_{*,*}^{\circ 2}(\wh{x})\| \|(p(x) - p(\wh{x}))\|_2 \notag\\
        \leq & ~ 2 \beta^{-2} R^2 \|x - \wh{x}\|_2 + 12 \beta^{-7} R^4 \|x - \wh{x}\|_2 + 3 \beta^{-7} R^4 \|x - \wh{x}\|_2 \notag\\
        \leq & ~ 17 \beta^{-7} R^4 \|x - \wh{x}\|_2,
    \end{align}
    where the first step follows from the triangle inequality, the second step follows from simple algebra, the third step follows from the Fact~\ref{fac:matrix_norm}, the fourth step follows from the Lemma~\ref{lem:basic_lips}, Lemma~\ref{lem:spectral_norm_bound}, Lemma~\ref{lem:more_spectral_norm_bound} and {\bf Part 1}.

    Second, we consider
    \begin{align}\label{eq:a_sigma_p_b}
        & ~ \|A_{x}^\top \Sigma(x) p(x) - A_{\wh{x}}^\top \Sigma(\wh{x})p(\wh{x})\|_2 \notag\\
        \leq & ~ \|A_{x}^\top \Sigma(x) p(x) - A_{\wh{x}}^\top \Sigma(x)p(x)\|_2 + \|A_{\wh{x}}^\top \Sigma(x) p(x) - A_{\wh{x}}^\top \Sigma(\wh{x})p(x)\|_2 \notag\\
        & ~ + \|A_{\wh{x}}^\top \Sigma(\wh{x}) p(x) - A_{\wh{x}}^\top \Sigma(\wh{x})p(\wh{x})\|_2 \notag\\
        = & ~ \|(A_{x}^\top - A_{\wh{x}}^\top) \Sigma(x)p(x)\|_2 + \|A_{\wh{x}}^\top (\Sigma(x)  - \Sigma(\wh{x})) p(x)\|_2 + \|A_{\wh{x}}^\top \Sigma(\wh{x}) (p(x) - p(\wh{x}))\|_2 \notag\\
        \leq & ~ \|A_{x}^\top - A_{\wh{x}}^\top\| \|\Sigma(x)\| \|p(x)\|_2 + \|A_{\wh{x}}^\top\| \|\Sigma(x)  - \Sigma(\wh{x})\| \|p(x)\|_2 \notag\\
        & ~ + \|A_{\wh{x}}^\top\| \|\Sigma(\wh{x})\| \|(p(x) - p(\wh{x}))\|_2 \notag\\
        \leq & ~ 2\beta^{-2} R^2 \|x - \wh{x}\|_2 + 6 \beta^{-7} R^4 \|x - \wh{x}\|_2 + 3\beta^{-7} R^4 \|x - \wh{x}\|_2 \notag\\
        \leq & ~ 11 \beta^{-7} R^4 \|x - \wh{x}\|_2,
    \end{align}
    where the first step follows from the triangle inequality, the second step follows from simple algebra, the third step follows from the Fact~\ref{fac:matrix_norm}, the fourth step follows from the Lemma~\ref{lem:basic_lips}, Lemma~\ref{lem:spectral_norm_bound} and Lemma~\ref{lem:more_spectral_norm_bound}.

    Combining Eq.~\ref{eq:g_gwh}, Eq.~\eqref{eq:a_sigma_p_a} and Eq.~\eqref{eq:a_sigma_p_b}, we have
    \begin{align*}
        \|g(x) - g(\wh{x})\|_2 \leq 60 \beta^{-7} R^4 \|x - \wh{x}\|_2.
    \end{align*}
\end{proof}

\begin{lemma}\label{lem:b_bwh}
If the following conditions hold
\begin{itemize}
    \item Let $B_1(x), B_2(x), B(x) \in \R^{n \times n}$ be computed as in Lemma~\ref{lem:L_c_Hessian}. 
    \item Let $d_l \in \{\Sigma(x), \diag(p(x)), \sigma_{*, *}(x), \sigma_{*, *}^{\circ 2}(x), \diag(\sigma_{i, *}(x))\}$.
    \item Let $b \in \R^{n}$, $\sigma_{*, *}^{\circ 2} \in \R^{n \times n}$, and $\sigma_{*, i}(x) \in \R^{n}$ be defined as in Definition~\ref{def:sigma}.
    \item $\beta \in (0, 0.1)$.
\end{itemize}

    Then, we have
    \begin{align*}
        \|B(x) - B(\wh{x})\| \leq 1000000 \beta^{-7} R^4 \|x - \wh{x}\|_2
    \end{align*}
\end{lemma}
\begin{proof}
    We have
    \begin{align*}
        \|B(x) - B(\wh{x})\|
        = & ~ \|B_1(x) + B_2(x) - B_1(\wh{x}) - B_2(\wh{x})\|\\
        \leq & ~ \|B_1(x) - B_1(\wh{x})\| + \|B_2(x) - B_2(\wh{x})\|,
    \end{align*}
    Where the first step follows from the Definition of B(x) (see Lemma~\ref{lem:L_c_Hessian}), the second step follows from Fact~\ref{fac:matrix_norm}.

    First, we consider
    \begin{align*}
        & ~ \|B_1(x) - B_1(\wh{x})\|\\
        = & ~ \|\sum_{l=1}^{10} D_l^\top \sum_{l=1}^{10} D_l - \sum_{l=1}^{10} \wh{D_l}^\top \sum_{l=1}^{10} \wh{D_l}\|\\
        \leq & ~ \|\sum_{l=1}^{10} D_l^\top \sum_{l=1}^{10} D_l - \sum_{l=1}^{10} \wh{D_l}^\top \sum_{l=1}^{10} D_l\| + \|\sum_{l=1}^{10} \wh{D_l}^\top \sum_{l=1}^{10} D_l - \sum_{l=1}^{10} \wh{D_l}^\top \sum_{l=1}^{10} \wh{D_l}\|\\
        \leq & ~ \|\sum_{l=1}^{10} D_l^\top - \sum_{l=1}^{10} \wh{D_l}^\top\| \|\sum_{l=1}^{10} D_l\| + \|\sum_{l=1}^{10} \wh{D_l}^\top\| \|\sum_{l=1}^{10} D_l -  \sum_{l=1}^{10} \wh{D_l}\|\\
        = & ~ 2 \|\sum_{l=1}^{10} D_l^\top - \sum_{l=1}^{10} \wh{D_l}^\top\| \|\sum_{l=1}^{10} D_l\|\\
        \leq & ~ 2 \sum_{l=1}^{10} \|D_l - \wh{D_l}\| \sum_{l=1}^{10} \| D_l\|,
    \end{align*}
    Where the first step follows from the Definition of $B_1(x)$ (see Lemma~\ref{lem:L_c_Hessian}), the second step, the third step, and the fourth step follow from Fact~\ref{fac:matrix_norm}, and the final step follows from simple algebra.

    For all $d_l \in \{\Sigma(x), \diag(p(x)), \sigma_{*, *}(x), \sigma_{*, *}^{\circ 2}(x), \diag(\sigma_{i, *}(x))\}$, by Lemma~\ref{lem:more_lips} and Lemma~\ref{lem:basic_lips}, we have
    \begin{align*}
        \|d_l - \wh{d}_l\| \leq 6 R^3 \|x - \wh{x}\|_2.
    \end{align*}

    Also, by Lemma~\ref{lem:spectral_norm_bound} and Lemma~\ref{lem:more_spectral_norm_bound}, for all $d_l \in \{\Sigma(x), \diag(p(x)), \sigma_{*, *}(x), \sigma_{*, *}^{\circ 2}(x), \diag(\sigma_{i, *}(x))\}$, we have
    \begin{align*}
        \|d_l\| \leq 2.
    \end{align*}

    By the definition of $D_l$, we have
    \begin{align*}
        \|D_l - \wh{D_l}\|
        \leq & ~ 10\|d_ld_ld_l - \wh{d_l}\wh{d_l}\wh{d_l}\|\\
        \leq & ~ 10(\|d_ld_ld_l - d_ld_l\wh{d_l}\| + \|d_ld_l\wh{d_l} - d_l\wh{d_l}\wh{d_l}\| + \|d_l\wh{d_l}\wh{d_l} - \wh{d_l}\wh{d_l}\wh{d_l}\|)\\
        \leq & ~ 10(\|d_l\| \|d_l\| \|d_l - \wh{d_l}\| + \|d_l\| \|d_l - \wh{d_l}\| \|\wh{d_l}\| + \|d_l - \wh{d_l}\| \|\wh{d_l}\| \|\wh{d_l}\|)\\
        = & ~ 10(2 \|d_l\| \|d_l\| \|d_l - \wh{d_l}\| + \|d_l\| \|d_l - \wh{d_l}\| \|\wh{d_l}\|)\\
        \leq & ~ 10(2^3 \cdot 6 R^3 \|x - \wh{x}\|_2 + 4 \cdot 6 R^3 \|x - \wh{x}\|_2)\\
        = & ~ 720 R^3 \|x - \wh{x}\|_2,
    \end{align*}
    where the first step follows from the definition of $D_l$, the second step follows from the triangle inequality, the third step follows from the Fact~\ref{fac:matrix_norm}, the fourth step follows from the simple algebra, the fifth step follows from the definition of $\|d_l\|$ and $\|d_l - \wh{d}_l\|$ above, the last step follows from the simple algebra.

    Therefore, we have
    \begin{align}\label{eq:b1_b1wh}
        & ~ \|B_1(x) - B_1(\wh{x})\| \notag\\
        \leq & ~ 2 \sum_{l=1}^{10} \|D_l - \wh{D_l}\| \sum_{l=1}^{10} \| D_l\| \notag\\
        \leq & ~ 2 \sum_{l=1}^{10} 720 R^3 \|x - \wh{x}\|_2 \sum_{l=1}^{10} \| D_l\| \notag\\
        \leq & ~ 2 \sum_{l=1}^{10} 720 R^3 \|x - \wh{x}\|_2 (20 + 4 + 16 + 8 + 2 + 2 + 4 + 4 + 2 + 2) \notag\\
        = & ~ 921600 R^3 \|x - \wh{x}\|_2,
    \end{align}
    where the first step follows from the definition of $\|B_1(x) - B_1(\wh{x})\|$ above, the second step follows from the definition of $\|D_l - \wh{D_l}\|$ above, the third step follows from the definition of $\| D_l\|$ in Lemma~\ref{lem:spectral_B}, the last step follows from the simple algebra.

    Now, we consider
    \begin{align*}
        \|B_2(x) - B_2(\wh{x})\|.
    \end{align*}

    Similarly, $B_2(x)$ consists of $(g(x) - c)^\top$, $A_x^\top$, $\sigma_{*,*}^{\circ 2}(x)$, $\sigma_{*,*}(x)$, $\diag(\sigma_{*,i}(x))$, $\Sigma(x)$.

    For all $d_l \in \{(g(x) - c)^\top, A_x^\top, \sigma_{*,*}^{\circ 2}(x), \sigma_{*,*}(x), \diag(\sigma_{*,i}(x)), \Sigma(x)\}$, by Lemma~\ref{lem:more_lips} and Lemma~\ref{lem:basic_lips}, we have
    \begin{align*}
        \|d_l - \wh{d}_l\| \leq 60 \beta^{-7} R^4 \|x - \wh{x}\|_2.
    \end{align*}

    Also, by Lemma~\ref{lem:spectral_norm_bound} and Lemma~\ref{lem:more_spectral_norm_bound}, for all $d_l \in \{(g(x) - c)^\top, \sigma_{*,*}^{\circ 2}(x), \sigma_{*,*}(x), \diag(\sigma_{*,i}(x)), \Sigma(x)\}$, we have
    \begin{align*}
        \|d_l\| \leq 2.
    \end{align*}

    Therefore, we have
    \begin{align}\label{eq:b2_b2wh}
        \|B_2(x) - B_2(\wh{x}) \notag\|
        \leq & ~ 88 (2 \|d_l\| \|d_l\| \|d_l - \wh{d_l}\| + \|d_l\| \|d_l - \wh{d_l}\| \|\wh{d_l}\|) \notag\\
        \leq & ~ 88 (2^3 (60 \beta^{-7} R^4 \|x - \wh{x}\|_2) + 2^2 (60 \beta^{-7} R^4 \|x - \wh{x}\|_2)) \notag\\
        = & ~ 63360 \beta^{-7} R^4 \|x - \wh{x}\|_2,
    \end{align}
    where the first step follows from the Lemma~\ref{lem:gradient_Al} and the result above, the second step follows from the definition of $\|d_l\|$ and $\|d_l - \wh{d}_l\|$ above, the last step follows from the simple algebra.

    Combining with Eq.~\ref{eq:b1_b1wh} and Eq.~\ref{eq:b2_b2wh}, we have
    \begin{align*}
        \|B(x) - B(\wh{x})\| 
        \leq & ~ 921600 R^3 \|x - \wh{x}\|_2 + 63360 \beta^{-7} R^4 \|x - \wh{x}\|_2\\
        \leq & ~ 1000000 \beta^{-7} R^4 \|x - \wh{x}\|_2,
    \end{align*}
    where the first step follows from the combination of Eq.~\ref{eq:b1_b1wh} and Eq.~\ref{eq:b2_b2wh}, the last step follows from the simple algebra.
\end{proof}

\begin{lemma}[Formal version of Lemma~\ref{lem:lip_lc_informal}]\label{lem:lip_lc}
If the following conditions hold

\begin{itemize}
    \item Let $B_1(x), B_2(x), B(x) \in \R^{n \times n}$ be computed as in Lemma~\ref{lem:L_c_Hessian}. 
    \item Let $d_l \in \{\Sigma(x), \diag(p(x)), \sigma_{*, *}(x), \sigma_{*, *}^{\circ 2}(x), \diag(\sigma_{i, *}(x))\}$.
    \item Let $b \in \R^{n}$, $\sigma_{*, *}^{\circ 2} \in \R^{n \times n}$, and $\sigma_{*, i}(x) \in \R^{n}$ be defined as in Definition~\ref{def:sigma}.
    \item Let $\|A_x\|\leq R$ for some $R>0$ and $\sigma_{\min}(A_x)\geq \beta$, where $\beta \in (0, 0.1)$.
\end{itemize}

    Then, we have
    \begin{align*}
        \|\frac{\d^2 L}{\d x^2}(x) - \frac{\d^2 L}{\d x^2}(\wh{x})\| \leq 1024000 \beta^{-7} R^6 \|x - \wh{x}\|_2.
    \end{align*}
\end{lemma}
\begin{proof}
    We have
    \begin{align*}
        & ~ \|\frac{\d^2 L}{\d x^2}(x) - \frac{\d^2 L}{\d x^2}(\wh{x})\|\\
        = & ~ \|A_{x}^\top B(x) A_{x} - A_{\wh{x}}^\top B(\wh{x}) A_{\wh{x}}\|\\
        \leq & ~ \|A_{x}^\top B(x) A_{x} - A_{x}^\top B(x) A_{\wh{x}}\| + \|A_{x}^\top B(x) A_{\wh{x}} - A_{x}^\top B(\wh{x}) A_{\wh{x}}\| \\
        & ~ + \|A_{x}^\top B(\wh{x}) A_{\wh{x}} - A_{\wh{x}}^\top B(\wh{x}) A_{\wh{x}}\|\\
        \leq & ~ \|A_{x}^\top\| \|B(x)\| \|A_{x} - A_{\wh{x}}\| + \|A_{x}^\top\| \|B(x)  - B(\wh{x})\| \|A_{\wh{x}}\| + \|A_{x}^\top - A_{\wh{x}}^\top\| \|B(\wh{x})\| \|A_{\wh{x}}\|\\
        \leq & ~ R (12000 \beta R) (\beta^{-2} R^2 \|x - \wh{x}\|_2) + R^2 (1000000 \beta^{-7} R^4 \|x - \wh{x}\|_2) + (\beta^{-2} R^2 \|x - \wh{x}\|_2) (12000 \beta R) R\\
        \leq & ~  1024000 \beta^{-7} R^6 \|x - \wh{x}\|_2,
    \end{align*}
    where the first step follows from the definition of $\frac{\d^2 L}{\d x^2}(x)$ (see the Lemma~\ref{lem:L_c_Hessian}), the second step follows from the triangle inequality, the third step follows from the Fact~\ref{fac:matrix_norm}, the fourth step follows from Lemma~\ref{lem:b_bwh}, Lemma~\ref{lem:basic_lips} and Lemma~\ref{lem:spectral_norm_bound}, the last step follows from the simple algebra.
\end{proof}

\section{More Related Work}

\paragraph{Data privacy}

In recent years, there has been a surge of interest in applying differential privacy to machine learning. \cite{abadi2016deep} introduced differentially private stochastic gradient descent, which enables the training of deep neural networks with privacy guarantees. This work has been extended by numerous researchers, including \cite{papernot2018scalable}, who developed the PATE framework for private knowledge transfer between models. Additional works using differential privacy to protect data privacy include \cite{lls+24c,lsss24_dp_ntk,lssz24_dp_tree,dr14,syyz23_dp}.

\newpage

\ifdefined\isarxiv

\bibliographystyle{alpha}
\bibliography{ref}

\newcommand{\etalchar}[1]{$^{#1}$}
\begin{thebibliography}{DMIMW12}

\bibitem[ACG{\etalchar{+}}16]{abadi2016deep}
Martin Abadi, Andy Chu, Ian Goodfellow, H~Brendan McMahan, Ilya Mironov, Kunal
  Talwar, and Li~Zhang.
\newblock Deep learning with differential privacy.
\newblock In {\em Proceedings of the 2016 ACM SIGSAC conference on computer and
  communications security}, pages 308--318, 2016.

\bibitem[AKK{\etalchar{+}}20]{akk+20}
Naman Agarwal, Sham Kakade, Rahul Kidambi, Yin-Tat Lee, Praneeth Netrapalli,
  and Aaron Sidford.
\newblock Leverage score sampling for faster accelerated regression and erm.
\newblock In {\em Algorithmic Learning Theory}, pages 22--47. PMLR, 2020.

\bibitem[AM15]{am15}
Ahmed Alaoui and Michael~W Mahoney.
\newblock Fast randomized kernel ridge regression with statistical guarantees.
\newblock {\em Advances in neural information processing systems}, 28, 2015.

\bibitem[Ama98]{a98}
Shun-Ichi Amari.
\newblock Natural gradient works efficiently in learning.
\newblock {\em Neural computation}, 10(2):251--276, 1998.

\bibitem[Ans00]{a00}
Kurt~M Anstreicher.
\newblock The volumetric barrier for semidefinite programming.
\newblock {\em Mathematics of Operations Research}, 2000.

\bibitem[AW21]{aw21}
Josh Alman and Virginia~Vassilevska Williams.
\newblock A refined laser method and faster matrix multiplication.
\newblock In {\em Proceedings of the 2021 ACM-SIAM Symposium on Discrete
  Algorithms (SODA)}, pages 522--539. SIAM, 2021.

\bibitem[BDJM73]{bdm73}
Charles~George Broyden, John~E Dennis~Jr, and Jorge~J Mor{\'e}.
\newblock On the local and superlinear convergence of quasi-newton methods.
\newblock {\em IMA Journal of Applied Mathematics}, 12(3):223--245, 1973.

\bibitem[BGM22]{bgm22}
Jimmy Ba, Roger Grosse, and James Martens.
\newblock Distributed second-order optimization using kronecker-factored
  approximations.
\newblock In {\em International conference on learning representations}, 2022.

\bibitem[BSY23]{bsy23}
Song Bian, Zhao Song, and Junze Yin.
\newblock Federated empirical risk minimization via second-order method.
\newblock {\em arXiv preprint arXiv:2305.17482}, 2023.

\bibitem[CH86]{ch86}
Samprit Chatterjee and Ali~S Hadi.
\newblock Influential observations, high leverage points, and outliers in
  linear regression.
\newblock {\em Statistical science}, pages 379--393, 1986.

\bibitem[CLE{\etalchar{+}}19]{cle+19}
Nicholas Carlini, Chang Liu, {\'U}lfar Erlingsson, Jernej Kos, and Dawn Song.
\newblock The secret sharer: Evaluating and testing unintended memorization in
  neural networks.
\newblock In {\em 28th USENIX Security Symposium (USENIX Security 19)}, pages
  267--284, 2019.

\bibitem[CLV17]{clv17}
Daniele Calandriello, Alessandro Lazaric, and Michal Valko.
\newblock Distributed adaptive sampling for kernel matrix approximation.
\newblock In {\em Artificial Intelligence and Statistics}, pages 1421--1429.
  PMLR, 2017.

\bibitem[CY21]{cy21}
Yifan Chen and Yun Yang.
\newblock Fast statistical leverage score approximation in kernel ridge
  regression.
\newblock In {\em International Conference on Artificial Intelligence and
  Statistics}, pages 2935--2943. PMLR, 2021.

\bibitem[DKM06]{dkm06}
Petros Drineas, Ravi Kannan, and Michael~W Mahoney.
\newblock Fast monte carlo algorithms for matrices i: Approximating matrix
  multiplication.
\newblock {\em SIAM Journal on Computing}, 36(1):132--157, 2006.

\bibitem[DLS23]{dls23}
Yichuan Deng, Zhihang Li, and Zhao Song.
\newblock Attention scheme inspired softmax regression.
\newblock {\em arXiv preprint arXiv:2304.10411}, 2023.

\bibitem[DMIMW12]{dmmw12}
Petros Drineas, Malik Magdon-Ismail, Michael~W Mahoney, and David~P Woodruff.
\newblock Fast approximation of matrix coherence and statistical leverage.
\newblock {\em The Journal of Machine Learning Research}, 13(1):3475--3506,
  2012.

\bibitem[DR{\etalchar{+}}14]{dr14}
Cynthia Dwork, Aaron Roth, et~al.
\newblock The algorithmic foundations of differential privacy.
\newblock {\em Foundations and Trends{\textregistered} in Theoretical Computer
  Science}, 9(3--4):211--407, 2014.

\bibitem[DSW22]{dsw22}
Yichuan Deng, Zhao Song, and Omri Weinstein.
\newblock Discrepancy minimization in input-sparsity time.
\newblock {\em arXiv preprint arXiv:2210.12468}, 2022.

\bibitem[DWZ23]{dwz23}
Ran Duan, Hongxun Wu, and Renfei Zhou.
\newblock Faster matrix multiplication via asymmetric hashing.
\newblock In {\em FOCS}, 2023.

\bibitem[EMM20]{emm20}
Tam{\'a}s Erd{\'e}lyi, Cameron Musco, and Christopher Musco.
\newblock Fourier sparse leverage scores and approximate kernel learning.
\newblock {\em Advances in Neural Information Processing Systems}, 33:109--122,
  2020.

\bibitem[GM16]{gm16}
Roger Grosse and James Martens.
\newblock A kronecker-factored approximate fisher matrix for convolution
  layers.
\newblock In {\em International Conference on Machine Learning}, pages
  573--582. PMLR, 2016.

\bibitem[GSWY23]{gswy23}
Yeqi Gao, Zhao Song, Weixin Wang, and Junze Yin.
\newblock A fast optimization view: Reformulating single layer attention in llm
  based on tensor and svm trick, and solving it in matrix multiplication time.
\newblock {\em arXiv preprint arXiv:2309.07418}, 2023.

\bibitem[GSY23a]{gsy23_coin}
Yeqi Gao, Zhao Song, and Junze Yin.
\newblock Gradientcoin: A peer-to-peer decentralized large language models.
\newblock {\em arXiv preprint arXiv:2308.10502}, 2023.

\bibitem[GSY23b]{gsy23_hyper}
Yeqi Gao, Zhao Song, and Junze Yin.
\newblock An iterative algorithm for rescaled hyperbolic functions regression.
\newblock {\em arXiv preprint arXiv:2305.00660}, 2023.

\bibitem[GSYZ23]{gsyz23}
Yuzhou Gu, Zhao Song, Junze Yin, and Lichen Zhang.
\newblock Low rank matrix completion via robust alternating minimization in
  nearly linear time.
\newblock {\em arXiv preprint arXiv:2302.11068}, 2023.

\bibitem[HJS{\etalchar{+}}22]{hjs+22}
Baihe Huang, Shunhua Jiang, Zhao Song, Runzhou Tao, and Ruizhe Zhang.
\newblock Solving sdp faster: A robust ipm framework and efficient
  implementation.
\newblock In {\em 2022 IEEE 63rd Annual Symposium on Foundations of Computer
  Science (FOCS)}, pages 233--244. IEEE, 2022.

\bibitem[Hua11]{l11}
Zejun Huang.
\newblock On the spectral radius and the spectral norm of hadamard products of
  nonnegative matrices.
\newblock {\em Linear algebra and its applications}, 434(2):457--462, 2011.

\bibitem[JKL{\etalchar{+}}20]{jkl+20}
Haotian Jiang, Tarun Kathuria, Yin~Tat Lee, Swati Padmanabhan, and Zhao Song.
\newblock A faster interior point method for semidefinite programming.
\newblock In {\em 2020 IEEE 61st annual symposium on foundations of computer
  science (FOCS)}, pages 910--918. IEEE, 2020.

\bibitem[JRM{\etalchar{+}}99]{jrm+99}
Craig~A Jensen, Russell~D Reed, Robert~Jackson Marks, Mohamed~A El-Sharkawi,
  Jae-Byung Jung, Robert~T Miyamoto, Gregory~M Anderson, and Christian~J Eggen.
\newblock Inversion of feedforward neural networks: algorithms and
  applications.
\newblock {\em Proceedings of the IEEE}, 87(9):1536--1549, 1999.

\bibitem[LG14]{lg14}
Fran{\c{c}}ois Le~Gall.
\newblock Powers of tensors and fast matrix multiplication.
\newblock In {\em Proceedings of the 39th international symposium on symbolic
  and algebraic computation}, pages 296--303, 2014.

\bibitem[LG23]{lg23}
Fran{\c{c}}ois Le~Gall.
\newblock Faster rectangular matrix multiplication by combination loss
  analysis.
\newblock {\em arXiv preprint arXiv:2307.06535}, 2023.

\bibitem[LHC{\etalchar{+}}20]{lhc+20}
Fanghui Liu, Xiaolin Huang, Yudong Chen, Jie Yang, and Johan Suykens.
\newblock Random fourier features via fast surrogate leverage weighted
  sampling.
\newblock In {\em Proceedings of the AAAI Conference on Artificial
  Intelligence}, volume~34, pages 4844--4851, 2020.

\bibitem[LKN99]{lkn99}
Bao-Liang Lu, Hajime Kita, and Yoshikazu Nishikawa.
\newblock Inverting feedforward neural networks using linear and nonlinear
  programming.
\newblock {\em IEEE Transactions on Neural networks}, 10(6):1271--1290, 1999.

\bibitem[LLS{\etalchar{+}}24]{lls+24c}
Xiaoyu Li, Yingyu Liang, Zhenmei Shi, Zhao Song, and Junwei Yu.
\newblock Fast john ellipsoid computation with differential privacy
  optimization.
\newblock {\em arXiv preprint arXiv:2408.06395}, 2024.

\bibitem[LN89]{ln89}
Dong~C Liu and Jorge Nocedal.
\newblock On the limited memory bfgs method for large scale optimization.
\newblock {\em Mathematical programming}, 45(1):503--528, 1989.

\bibitem[LSSS24]{lsss24_dp_ntk}
Yingyu Liang, Zhizhou Sha, Zhenmei Shi, and Zhao Song.
\newblock Differential privacy mechanisms in neural tangent kernel regression.
\newblock {\em arXiv preprint arXiv:2407.13621}, 2024.

\bibitem[LSSZ24]{lssz24_dp_tree}
Yingyu Liang, Zhenmei Shi, Zhao Song, and Yufa Zhou.
\newblock Differential privacy of cross-attention with provable guarantee.
\newblock {\em arXiv preprint arXiv:2407.14717}, 2024.

\bibitem[LSW{\etalchar{+}}24]{lsw+24}
Zhihang Li, Zhao Song, Weixin Wang, Junze Yin, and Zheng Yu.
\newblock How to inverting the leverage score distribution?
\newblock {\em arXiv preprint arXiv:2404.13785}, 2024.

\bibitem[LSWY23]{lswy23}
Zhihang Li, Zhao Song, Zifan Wang, and Junze Yin.
\newblock Local convergence of approximate newton method for two layer
  nonlinear regression.
\newblock {\em arXiv preprint arXiv:2311.15390}, 2023.

\bibitem[LSZ23a]{lsz23}
Zhihang Li, Zhao Song, and Tianyi Zhou.
\newblock Solving regularized exp, cosh and sinh regression problems.
\newblock {\em arXiv preprint arXiv:2303.15725}, 2023.

\bibitem[LSZ{\etalchar{+}}23b]{lsz+23}
S.~Cliff Liu, Zhao Song, Hengjie Zhang, Lichen Zhang, and Tianyi Zhou.
\newblock Space-efficient interior point method, with applications to linear
  programming and maximum weight bipartite matching.
\newblock In {\em International Colloquium on Automata, Languages and
  Programming (ICALP)}, pages 88:1--88:14, 2023.

\bibitem[LTOS19]{ltos19}
Zhu Li, Jean-Francois Ton, Dino Oglic, and Dino Sejdinovic.
\newblock Towards a unified analysis of random fourier features.
\newblock In {\em International conference on machine learning}, pages
  3905--3914. PMLR, 2019.

\bibitem[M{\etalchar{+}}10]{m10}
James Martens et~al.
\newblock Deep learning via hessian-free optimization.
\newblock In {\em Icml}, volume~27, pages 735--742, 2010.

\bibitem[M{\etalchar{+}}11]{m11}
Michael~W Mahoney et~al.
\newblock Randomized algorithms for matrices and data.
\newblock {\em Foundations and Trends{\textregistered} in Machine Learning},
  3(2):123--224, 2011.

\bibitem[McC18]{m18}
Shannon McCurdy.
\newblock Ridge regression and provable deterministic ridge leverage score
  sampling.
\newblock {\em Advances in Neural Information Processing Systems}, 31, 2018.

\bibitem[MD09]{md09}
Michael~W Mahoney and Petros Drineas.
\newblock Cur matrix decompositions for improved data analysis.
\newblock {\em Proceedings of the National Academy of Sciences},
  106(3):697--702, 2009.

\bibitem[MG15]{mg15}
James Martens and Roger Grosse.
\newblock Optimizing neural networks with kronecker-factored approximate
  curvature.
\newblock In {\em International conference on machine learning}, pages
  2408--2417. PMLR, 2015.

\bibitem[MM17]{mm17}
Cameron Musco and Christopher Musco.
\newblock Recursive sampling for the nystrom method.
\newblock {\em Advances in neural information processing systems}, 30, 2017.

\bibitem[PKB14]{pkb14}
Dimitris Papailiopoulos, Anastasios Kyrillidis, and Christos Boutsidis.
\newblock Provable deterministic leverage score sampling.
\newblock In {\em Proceedings of the 20th ACM SIGKDD international conference
  on Knowledge discovery and data mining}, pages 997--1006, 2014.

\bibitem[PSM{\etalchar{+}}18]{papernot2018scalable}
Nicolas Papernot, Shuang Song, Ilya Mironov, Ananth Raghunathan, Kunal Talwar,
  and {\'U}lfar Erlingsson.
\newblock Scalable private learning with pate.
\newblock In {\em International conference on learning representations}, 2018.

\bibitem[SS02]{ss02}
Bernhard Sch{\"o}lkopf and Alexander~J Smola.
\newblock {\em Learning with kernels: support vector machines, regularization,
  optimization, and beyond}.
\newblock MIT press, 2002.

\bibitem[SWY23]{swy23}
Zhao Song, Weixin Wang, and Junze Yin.
\newblock A unified scheme of resnet and softmax.
\newblock {\em arXiv preprint arXiv:2309.13482}, 2023.

\bibitem[SYYZ22]{syyz22}
Zhao Song, Xin Yang, Yuanyuan Yang, and Tianyi Zhou.
\newblock Faster algorithm for structured john ellipsoid computation.
\newblock {\em arXiv preprint arXiv:2211.14407}, 2022.

\bibitem[SYYZ23a]{syyz23_dp}
Zhao Song, Xin Yang, Yuanyuan Yang, and Lichen Zhang.
\newblock Sketching meets differential privacy: fast algorithm for dynamic
  kronecker projection maintenance.
\newblock In {\em International Conference on Machine Learning (ICML)}, pages
  32418--32462. PMLR, 2023.

\bibitem[SYYZ23b]{syyz23}
Zhao Song, Mingquan Ye, Junze Yin, and Lichen Zhang.
\newblock Efficient alternating minimization with applications to weighted low
  rank approximation.
\newblock {\em arXiv preprint arXiv:2306.04169}, 2023.

\bibitem[SYZ23]{syz23}
Zhao Song, Junze Yin, and Ruizhe Zhang.
\newblock Revisiting quantum algorithms for linear regressions: Quadratic
  speedups without data-dependent parameters.
\newblock {\em arXiv preprint arXiv:2311.14823}, 2023.

\bibitem[SZZ21]{szz21}
Zhao Song, Lichen Zhang, and Ruizhe Zhang.
\newblock Training multi-layer over-parametrized neural network in subquadratic
  time.
\newblock {\em arXiv preprint arXiv:2112.07628}, 2021.

\bibitem[VP12]{vp12}
Oriol Vinyals and Daniel Povey.
\newblock Krylov subspace descent for deep learning.
\newblock In {\em Artificial intelligence and statistics}, pages 1261--1268.
  PMLR, 2012.

\bibitem[Wil12]{w12}
Virginia~Vassilevska Williams.
\newblock Multiplying matrices faster than coppersmith-winograd.
\newblock In {\em Proceedings of the forty-fourth annual ACM symposium on
  Theory of computing}, pages 887--898, 2012.

\bibitem[WXXZ23]{wxxz23}
Virginia~Vassilevska Williams, Yinzhan Xu, Zixuan Xu, and Renfei Zhou.
\newblock New bounds for matrix multiplication: from alpha to omega, 2023.

\bibitem[ZJP{\etalchar{+}}20]{zjp+20}
Yuheng Zhang, Ruoxi Jia, Hengzhi Pei, Wenxiao Wang, Bo~Li, and Dawn Song.
\newblock The secret revealer: Generative model-inversion attacks against deep
  neural networks.
\newblock In {\em Proceedings of the IEEE/CVF conference on computer vision and
  pattern recognition}, pages 253--261, 2020.

\bibitem[ZLH19]{zlh19}
Ligeng Zhu, Zhijian Liu, and Song Han.
\newblock Deep leakage from gradients.
\newblock {\em Advances in neural information processing systems}, 32, 2019.

\end{thebibliography}

\else

\fi




\end{document}